\pdfoutput=1

\documentclass[11pt]{article}

\usepackage[preprint]{acl}

\usepackage{times}
\usepackage{latexsym}

\usepackage[T1]{fontenc}

\usepackage[utf8]{inputenc}

\usepackage{microtype}

\usepackage{inconsolata}

\usepackage{graphicx}
\usepackage{multirow}
\usepackage{multicol}
\usepackage{float}

\usepackage{amssymb}
\definecolor{medorange}{HTML}{ffb255}
\definecolor{medgreen}{HTML}{98c127}
\definecolor{medpink}{HTML}{f45f74}

\usepackage{booktabs}
\usepackage{array}

\newcommand{\greenbullet}[0]{\textcolor{medgreen}{$\bullet$}\color{black}}
\newcommand{\orangebullet}[0]{\textcolor{medorange}{$\scalebox{0.8}{$\ \blacktriangle$}$
}\color{black}}
\newcommand{\redbar}[0]{\textcolor{medpink}{\textbf{$\times$}}\color{black}}

\title{Evaluating Step-by-step Reasoning Traces: A Survey}

\author{Jinu Lee \and Julia Hockenmaier \\
  University of Illinois Urbana-Champaign \\
  \texttt{\{jinulee2, juliahmr\}@illinois.edu}}

\begin{document}
\maketitle
\begin{abstract}
Step-by-step reasoning is widely used to enhance the reasoning ability of large language models (LLMs) in complex problems. 
Evaluating the quality of reasoning traces is crucial for understanding and improving LLM reasoning. However, existing evaluation practices are highly inconsistent, resulting in fragmented progress across evaluator design and benchmark development. To address this gap, this survey provides a comprehensive overview of step-by-step reasoning evaluation, proposing a taxonomy of evaluation criteria with four top-level categories (factuality, validity, coherence, and utility). Based on the taxonomy, we review different datasets, evaluator implementations, and recent findings, leading to promising directions for future research.
\end{abstract}

\section{Introduction}
\label{sec:introduction}

Large language models (LLMs) have demonstrated remarkable capabilities in reasoning on complex problems, such as logic, math, and science. At the core of this versatility lies \textbf{step-by-step reasoning} \citep{NEURIPS2022_9d560961, NEURIPS2022_8bb0d291}, where the LLM generates an intermediate reasoning trace before presenting the final answer.

\begin{figure}
    \centering
    \includegraphics[width=\linewidth]{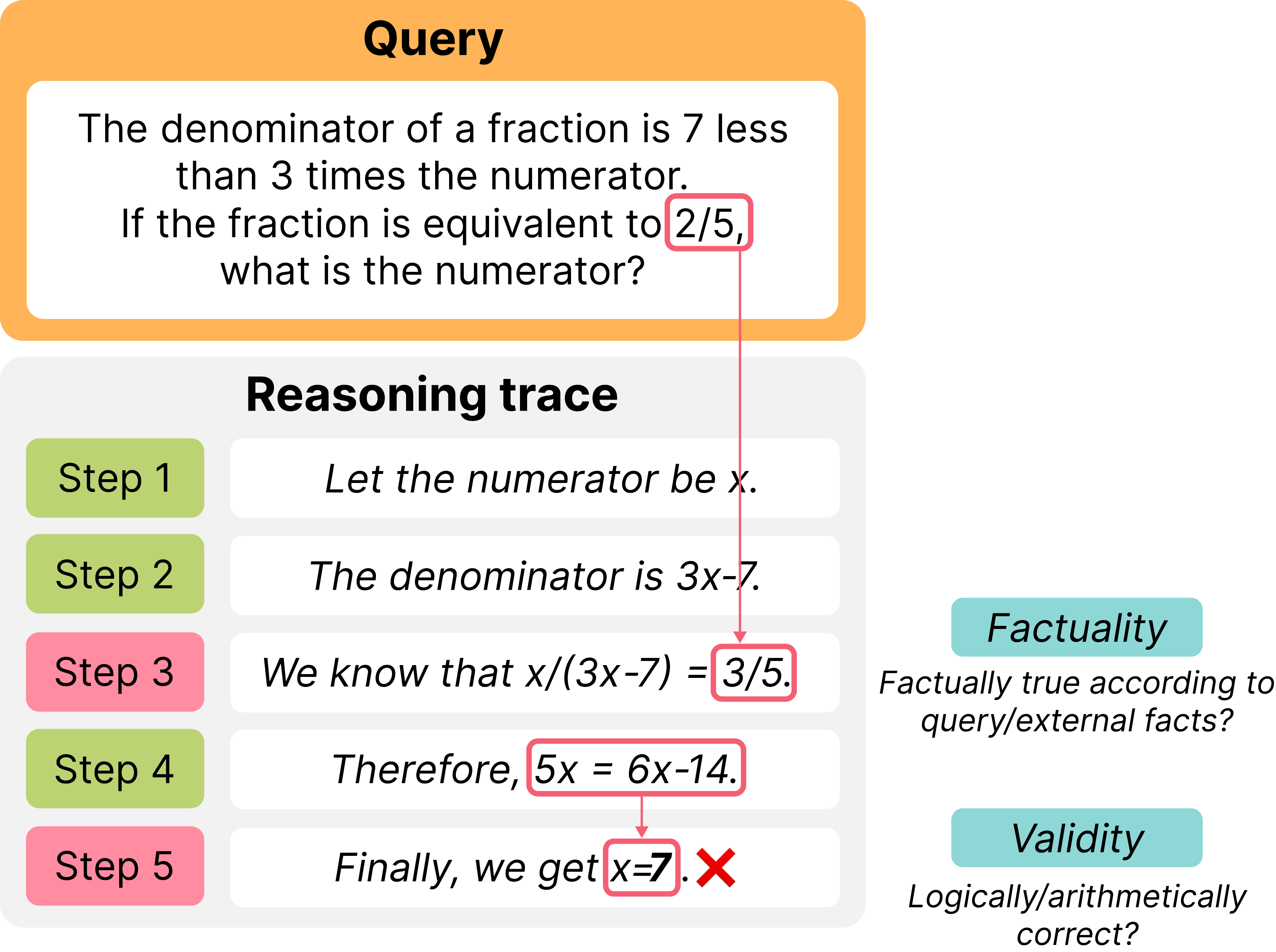}
    \caption{Illustrative example of reasoning trace evaluation.}
    \label{fig:intro}
\end{figure}

The reasoning ability of LLMs is often measured in terms of \textit{answer accuracy}, \textit{i.e.}, finding the correct answer for a complex reasoning problem  \citep{cobbe2021trainingverifierssolvemath, zhong2021arlsatinvestigatinganalyticalreasoning}. However, answer accuracy is generally insufficient for measuring LLMs' reasoning ability, as the correct answer does not imply the correctness of the preceding reasoning trace \citep{lanham2023measuringfaithfulnesschainofthoughtreasoning, mirzadeh2024gsmsymbolicunderstandinglimitationsmathematical, paul-etal-2024-making}. Furthermore, assessing the quality of the reasoning trace can directly lead to better reasoning ability of LLMs by verifier-guided search \citep{DBLP:conf/iclr/0002WSLCNCZ23, NEURIPS2023_271db992, hao2024llmreasonersnewevaluation} and reinforcement learning \citep{lu2024stepcontrolleddpoleveragingstepwise, cui2025processreinforcementimplicitrewards}.

Consequently, reasoning trace evaluation is an active research topic, with numerous new evaluators and datasets continuously being proposed. However, this rapid growth has led to a proliferation of evaluators and datasets without establishing a consensus on the \textbf{criteria} (\textit{what to evaluate}). In this survey, we aim to provide a systematic review of existing step-by-step reasoning evaluation criteria, which will serve as a foundation for implementing evaluators. 

Implementing an evaluator also introduces several practical decision choices. Different architectures involve trade-offs between computational cost and expected performance, while non-architectural factors like training data and data format also play a significant role. This survey seeks to categorize and compare various evaluator implementations, highlighting key trade-offs and revealing additional dimensions that merit consideration.

The key contributions of this survey are:
\begin{itemize}
    \item Defining a clear, universal taxonomy of step-by-step evaluation \textbf{criteria} (\S\ref{sec:criteria}).
    \item Surveying existing \textbf{datasets} and \textbf{evaluators} for step-by-step reasoning evaluation based on their implementations, across diverse reasoning tasks and criteria (\S\ref{sec:resources}-\S\ref{sec:evaluators}).
    \item Identifying recent findings and promising directions for trace evaluation (\S\ref{sec:analysis}-\S\ref{sec:future-directions}).
\end{itemize}
\section{Background}
\label{sec:background}

\subsection{Step-by-step reasoning}

\label{sec:task-description}

\textbf{Step-by-step reasoning} is where LLMs generate a series of intermediate natural language steps ("thoughts") before outputting the final answer \citep{NEURIPS2022_9d560961}. Each instance consists of two parts: a \textbf{query} and a \textbf{reasoning trace}, and the \textbf{final answer} as a part of the reasoning trace. Upon seeing the query (user input), the LLM autoregressively generates its solution as a reasoning trace. Finally, a trace should include a \textbf{final answer} for the query, which can be compared to the ground truth.
See Appendix \ref{sec:appendix-task} for details on different reasoning tasks.

\subsection{Evaluation}

Reasoning trace \textbf{evaluators} assess the quality of the reasoning trace and assign a score, reflecting whether it is good or not based on the criterion. Evaluators can be intrinsic metrics like uncertainty to models specialized for reasoning trace evaluation; see Section \ref{sec:evaluators} for different types of evaluators.

\subsection{Meta-evaluation}
\label{sec:meta-eval}

\begin{figure}[tb]
    \centering
    \includegraphics[width=0.95\linewidth]{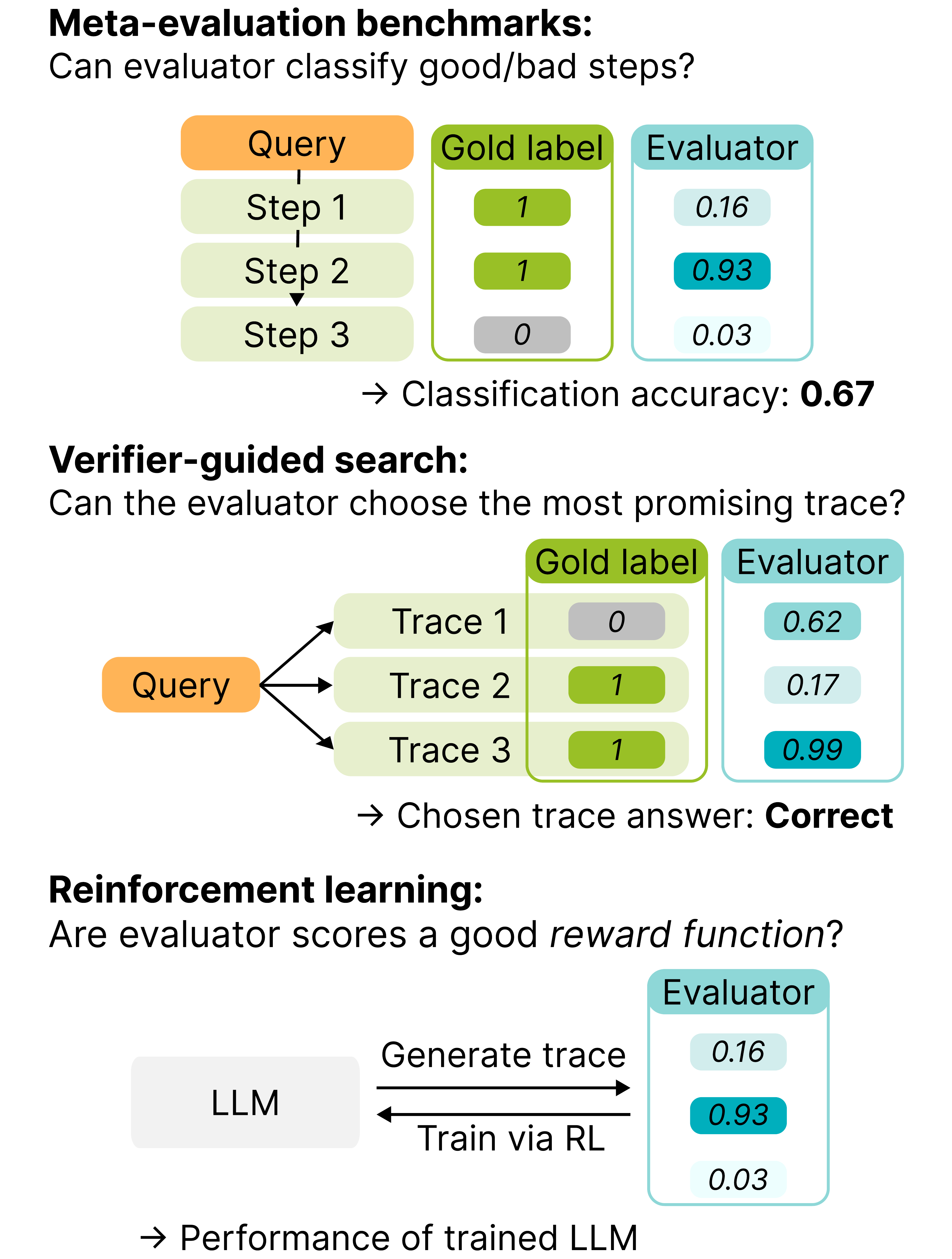}
    \caption{Illustration of three popular meta-evaluation methods: meta-evaluation benchmarks, verifier-guided search, and reinforcement learning.}
    \label{fig:elements}
\end{figure}

How can we \textit{evaluate these evaluators (meta-evaluation)}? Two common directions apply: (1) using meta-evaluation benchmarks with step-wise labels, or (2) measuring the improvement in the downstream task performance (Figure \ref{fig:elements}).

\subsubsection{Meta-evaluation Benchmarks}

\textbf{Meta-evaluation benchmarks} contains labels indicating a step's quality based on the predefined criteria. In this setting, the evaluator's performance is measured by the classification accuracy of these labels. These benchmarks offer a fine-grained view of which criteria the evaluator can handle well and which cannot \citep{song2025prmbenchfinegrainedchallengingbenchmark}. However, constructing these data often requires costly human annotation \citep{DBLP:conf/iclr/LightmanKBEBLLS24, zheng2024processbenchidentifyingprocesserrors} and the gains in meta-evaluation benchmark might not generalize to downstream performance \citep{zhang2025lessonsdevelopingprocessreward}.
Further details can be found in Appendix \ref{sec:appendix-resources}.

\subsubsection{Downstream performance improvement}
\label{sec:downstream}

As the fundamental goal of evaluators is to \textit{improve} the reasoning ability of LLMs, the evaluator's quality can also be measured by the improvement in downstream reasoning tasks.

\textbf{Verifier-guided search} uses evaluator scores to find the most promising trace after exploring different paths. Popular methods include \textbf{Best-of-N decoding} (independently sampling N traces and selecting one) \citep{DBLP:conf/iclr/LightmanKBEBLLS24, zhang2024generativeverifiersrewardmodeling} and \textbf{tree search} (sampling multiple candidate steps and choosing the most promising path) \citep{NEURIPS2023_271db992, guan2024searchverifyfeedbackgeneration, zhu2024deductivebeamsearchdecoding}. The performance is often compared to majority voting without evaluators (Self-consistency; \citet{DBLP:conf/iclr/0002WSLCNCZ23}), where a bigger gap indicates a better evaluator performance.

\textbf{Reinforcement learning} (RL) uses evaluator scores as a \textit{reward} to further train an LLM \citep{uesato2022solvingmathwordproblems, pan2023letsreinforcestepstep, NEURIPS2024_76ec4dc3}. If the evaluator provides useful training rewards, the trained model will reach higher final answer accuracy.
Moreover, as evaluators that are vulnerable to spurious features like length lead to \textit{reward hacking}, successful RL also indicates the evaluator's robustness \citep{NEURIPS2024_76ec4dc3}.

\begin{figure*}[t]
    \centering
    \includegraphics[width=\linewidth]{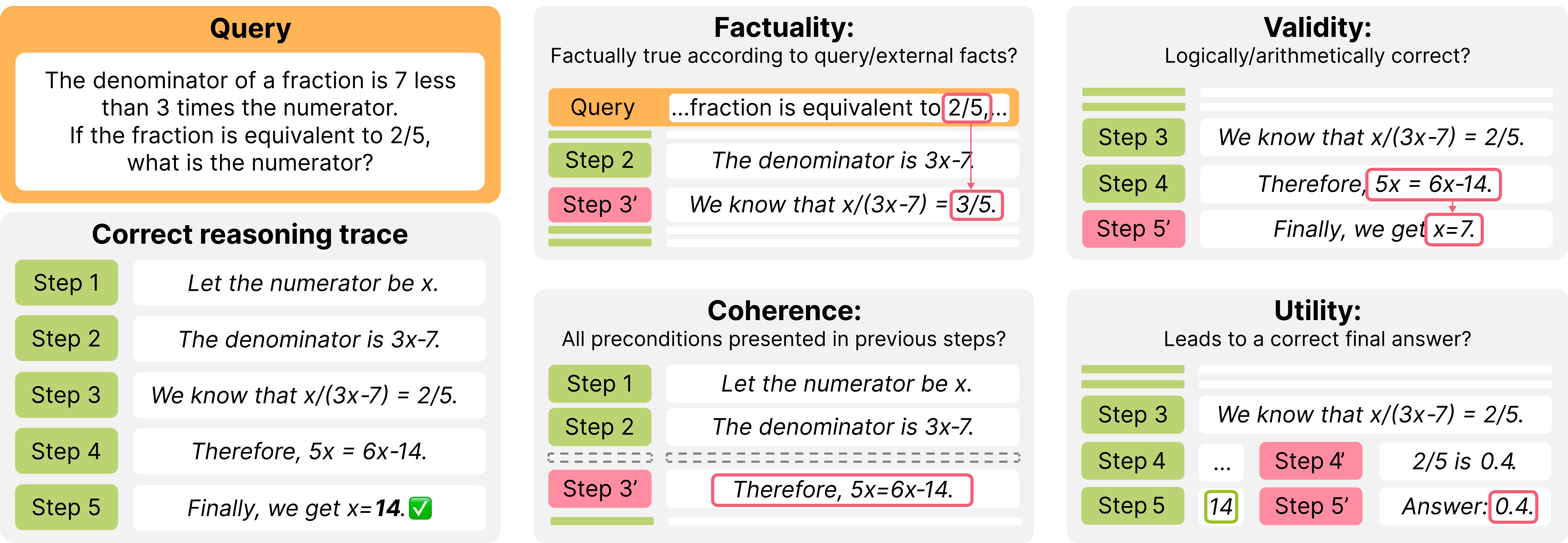}
    \caption{Illustration of the proposed categories of step-by-step reasoning evaluation criteria, \textit{i.e.} factuality, validity, coherence, and utility. 
    The left shows an example of a query and a reasoning trace. The other four blocks demonstrate examples that fail to satisfy the respective metric. Red filled rectangles indicate the error's location, and the outlined boxes and arrows show the cause of the error. The trace example is originally from \citet{DBLP:conf/iclr/LightmanKBEBLLS24}.}
    \label{fig:criteria}
\end{figure*}

\section{Evaluation Criteria}
\label{sec:criteria}

Previous studies have proposed various criteria for evaluating step-by-step reasoning \citep{DBLP:conf/iclr/GolovnevaCPCZFC23, DBLP:conf/iclr/LightmanKBEBLLS24, wang-etal-2024-math, jacovi-etal-2024-chain}, but these works failed to propose a complete taxonomy that covers diverse reasoning tasks (\textit{e.g.}, literature in factual reasoning and math reasoning have focused on different criteria). In this section, we propose a unified taxonomy of reasoning trace evaluation criteria that spans different reasoning tasks and evaluators. We categorize them into four key dimensions: \textbf{Factuality}, \textbf{Validity}, \textbf{Coherence}, and \textbf{Utility} (Figure \ref{fig:criteria})\footnote{These criteria are independent but not mutually exclusive (a step can fail to satisfy multiple criteria).}.

\subsection{Factuality}

\textbf{Factuality} evaluates if the factual information can be grounded in reliable sources.

The narrower notion of factuality is \textbf{groundedness}, where the generated trace should be true according to the \textit{query} \citep{NEURIPS2020_6b493230, gao2024retrievalaugmentedgenerationlargelanguage}. For instance, if the retrieved document explicitly mentions that \textit{Einstein was born in 1879}, the step mentioning that he was born in \textit{1789} is ungrounded. In less factual tasks like math, groundedness also indicates using correct numbers and constraints given in the query.

However, the reasoning process might require factual knowledge not directly mentioned in the query. This type of factuality can be referred to as \textbf{parametric knowledge}. While steps containing trivia-style facts can be readily verified by retrieval-based fact checkers \citep{thorne-etal-2018-fever}, verifying subtle, commonsensical knowledge remains an open challenge \citep{toroghi-etal-2024-right}.

\subsection{Validity}

\textbf{Validity} evaluates if a reasoning step contains no logical errors.

The validity of a reasoning step can be defined in terms of \textit{entailment} \citep{bowman-etal-2015-large}, which is widely accepted in factual/commonsense-based reasoning \citep{prasad-etal-2023-receval, wu-etal-2024-synchronous}. Under this definition, a step is considered valid if it can be directly entailed from previous steps \citep{tafjord-etal-2021-proofwriter, dalvi-etal-2021-explaining, PrOntoQA} or at least does not contradict them \citep{DBLP:conf/iclr/GolovnevaCPCZFC23, prasad-etal-2023-receval, zhu2024deductivebeamsearchdecoding}.

In tasks like math or logic, the more common form of validity is \textit{correctness}, \textit{e.g.} performing accurate calculations in arithmetic reasoning \citep{DBLP:conf/iclr/LightmanKBEBLLS24, jacovi-etal-2024-chain, zheng2024processbenchidentifyingprocesserrors} or inferring the correct logical conclusion based on the provided premises \citep{wu2024cofcastepwisecounterfactualmultihop, song2025prmbenchfinegrainedchallengingbenchmark}.



\begin{table*}[ht]
    \centering
    \small
    \begin{tabular}{p{4.9cm} >{\centering\arraybackslash}p{0.6cm} >{\centering\arraybackslash}p{0.6cm} p{3.8cm} >{\centering\arraybackslash}p{1cm} >{\centering\arraybackslash}p{1cm} >{\centering\arraybackslash}p{0.8cm}}
        \toprule
        \multicolumn{1}{c}{\textbf{Dataset}} & \textbf{Train} & \textbf{Eval} & \textbf{Domain} & \textbf{Criteria} & \textbf{\# Trace} & \textbf{Human} \\
        \midrule
        ROSCOE \citep{golovneva2023pathfinderguidedsearchmultistep} & & \greenbullet & Math, Common & \texttt{FVU} & 1.0k & \greenbullet \\
        RAGTruth$^\dagger$ \citep{niu2024ragtruthhallucinationcorpusdeveloping} & \greenbullet & \greenbullet & Fact & \texttt{F} & 5.9k & \greenbullet \\
        HaluEval$^\dagger$\citep{li2023haluevallargescalehallucinationevaluation} & \greenbullet & \greenbullet & Fact & \texttt{F} & 10k & \orangebullet \\
        Math-Shepherd \citep{wang-etal-2024-math} & \greenbullet & & Math & \texttt{U} & 440k & \redbar \\
        PRM800k \citep{DBLP:conf/iclr/LightmanKBEBLLS24} & \greenbullet & \greenbullet & Math & \texttt{V}& 75k & \greenbullet \\
        REVEAL \citep{jacovi-etal-2024-chain} & & \greenbullet & Common & \texttt{FVC} & 3.4k & \greenbullet \\
        MATH-Minos \citep{gao2024llmcriticshelpcatch} & \greenbullet & & Math & \texttt{V} & 440k & \redbar \\
        SCDPO \citep{lu2024stepcontrolleddpoleveragingstepwise} & \greenbullet & & Math & \texttt{U} & 30k & \redbar \\
        MR-GSM8k \citep{zeng2024mrgsm8kmetareasoningbenchmarklarge} & & \greenbullet & Math & \texttt{V} & 3.0k & \greenbullet \\
        BIG-Bench-Mistake \citep{tyen-etal-2024-llms} & & \greenbullet & Symbolic & \texttt{VCU}& 2.2k & \greenbullet \\
        CriticBench \citep{lin2024criticbenchbenchmarkingllmscritiquecorrect} & & \greenbullet & Math, Common, Symbolic & \texttt{VU}& 3.8k & \redbar \\
        ProcessBench \citep{zheng2024processbenchidentifyingprocesserrors} & & \greenbullet & Math & \texttt{V} & 3.4k & \greenbullet \\
        MR-Ben \citep{NEURIPS2024_d81cb1f4}  & & \greenbullet & Science, Deductive, Coding & \texttt{V} & 6.0k & \greenbullet \\
        MR-MATH \citep{xia2025evaluatingmathematicalreasoningaccuracy} & & \greenbullet & Math & \texttt{VU} & 0.1k & \greenbullet \\
        PRMBench \citep{song2025prmbenchfinegrainedchallengingbenchmark} & & \greenbullet & Math & \texttt{VCU} & 6.2k & \orangebullet \\
        PRM-Clinic \citep{wang2025processsupervisedrewardmodelsverifying} &  & \greenbullet & Expert(Clinic) & \texttt{FVC} & 9.7k & \redbar \\
        VersaPRM \citep{zeng2025versaprmmultidomainprocessreward} & \greenbullet & & Expert & \texttt{FV} & 84.1k & \redbar \\
        BiGGenBench$^\dagger$ \citep{kim2024biggenbenchprincipledbenchmark} & & \greenbullet & Math, Logic & Custom & 0.1k & \redbar \\
        \bottomrule
    \end{tabular}
    \caption{List of evaluator training data and meta-evaluation benchmarks. $^\dagger$ symbol indicates that the datasets include other tasks, such as summarization, instruction following, \textit{etc}, where the \textbf{\# Trace} column only counts the reasoning subset. \textbf{Train/Eval} columns denote if the dataset is used for training or meta-evaluation. \textbf{Domain} indicates what tasks are used to sample the reasoning trace. \textbf{Criteria} column shows the criteria used to annotate the data classified according to Section \ref{sec:criteria}, where \texttt{FVCU} stands for factuality, validity, coherence, and utility, respectively. BiGGenBench \citep{kim2024biggenbenchprincipledbenchmark} applies hand-written, query-specific evaluation criteria (\textbf{Custom}). \textbf{Human} column indicates human annotation, where \greenbullet\orangebullet\redbar\ denotes full human annotation, automatic annotation/perturbation with human verification, and full LLM-based annotation, respectively.}
    \label{tab:resources}
\end{table*}


\subsection{Coherence}
\label{sec:coherence}

\textbf{Coherence} measures if a reasoning step's \textit{preconditions are satisfied} by the previous steps \citep{wang-etal-2023-towards, lee2025symbasymbolicbackwardchaining}. For instance, if a trace includes the reasoning step \textit{"Next, we add 42 to 16."} but the origin of the value 42 was never explained in the previous steps; this step is considered incoherent. An intuitive way to obtain an incoherent trace is randomly shuffling a coherent trace \citep{wang-etal-2023-towards, nguyen-etal-2024-direct}, as the premise of some steps will not appear anywhere in the previous steps (\textit{incoherent}) even though it can be eventually deduced from the query (\textit{valid}).

Note that coherence judgment is inherently subjective and pragmatic compared to other criteria \citep{jacovi-etal-2024-chain}. For instance, seemingly trivial steps like \textit{"A part of something is present in that something"} in WorldTree V2 \citep{xie-etal-2020-worldtree} are annotated as necessary in \citet{dalvi-etal-2021-explaining} but not necessary in \citet{Ott_2023}.



\subsection{Utility}
\label{sec:utility}

\textbf{Utility} measures whether a reasoning step contributes to getting the correct final answer.

The narrower interpretation of utility is \textit{progress}, or whether the step is correctly following the ground truth solution \citep{PrOntoQA, nguyen-etal-2024-direct}. For instance, in Game of 24 (making the number 24 using 4 natural numbers and basic arithmetic operations) \citep{NEURIPS2023_271db992}, a solution can be defined as a sequence of operations (\textit{e.g.} $5+7=12$ $\rightarrow$ $12-6=6$ $\rightarrow$ $6\times4=24$). In this task, the utility of a step (making $5+7=12$ from $5$ and $7$) can be directly assessed by checking if it is a part of a correct solution.

The more general version of utility is \textit{value function} (estimated reward). \citep{chen-etal-2023-rev, wang-etal-2024-math, setlur2024rewardingprogressscalingautomated}. Value function is often measured using Monte Carlo Tree Search (MCTS), where the step's value is determined by the average/maximum reward of sampled continuations. Evaluating utility as a value function offers high scalability as it only requires the gold answer for computing the reward, without any human annotation or ground-truth solutions \citep{wang-etal-2024-math, lai2024stepdpostepwisepreferenceoptimization, cui2025processreinforcementimplicitrewards}.

\section{Meta-evaluation Datasets}
\label{sec:resources}

Datasets that annotate LLM-generated reasoning traces serve as key resources for training evaluators and conducting meta-evaluations between evaluators. A summary of existing datasets is provided in Table \ref{tab:resources}.

Among these, one of the most influential is PRM800k \citep{DBLP:conf/iclr/LightmanKBEBLLS24}. PRM800k consists of crowdsourced tertiary validity labels (positive, negative, neutral) assigned step by step, framing reasoning trace error detection as a sequence classification problem. Its design has inspired several successors \citep{zeng2024mrgsm8kmetareasoningbenchmarklarge, xia2025evaluatingmathematicalreasoningaccuracy}, setting the paradigm for subsequent reasoning trace evaluation resources.

To address different needs, several extensions have been developed. Since human annotations are costly and difficult to scale, many works have explored automatic labeling—either by estimating utility through Monte Carlo Tree Search (MCTS) \citep{wang-etal-2024-math, luo2024improvemathematicalreasoninglanguage, setlur2024rewardingprogressscalingautomated} or by generating perturbed traces with LLMs \citep{lu2024stepcontrolleddpoleveragingstepwise, song2025prmbenchfinegrainedchallengingbenchmark}. More recent datasets further broaden the scope by enabling multi-criteria meta-evaluation \citep{jacovi-etal-2024-chain, tyen-etal-2024-llms, song2025prmbenchfinegrainedchallengingbenchmark} and expanding coverage beyond mathematics into diverse domains \citep{NEURIPS2024_d81cb1f4, zeng2025versaprmmultidomainprocessreward}.

Additional details are provided in Appendix \ref{sec:appendix-resources}.

\section{Evaluator types}
\label{sec:evaluators}

\begin{table}[tb]
    \centering
    \footnotesize
    \begin{tabular}{lcccc}
        \toprule
        Metric impl. & \texttt{F} & \texttt{V} & \texttt{C} & \texttt{U} \\
        \midrule
        Rule-based  & \orangebullet & \orangebullet  & \orangebullet  & \orangebullet  \\
        Uncertainty  &  \greenbullet &   &  \orangebullet &  \\
        $\mathcal{V}$-information & & \orangebullet & \orangebullet & \greenbullet \\
        LLM-as-value-function &  & &  & \greenbullet \\
        Cross-encoder & \greenbullet & \greenbullet & \orangebullet & \orangebullet \\
        Sequence classifiers & \orangebullet & \greenbullet &  & \greenbullet \\
        Critic models & \greenbullet & \greenbullet & \orangebullet & \greenbullet \\
        Generative verifiers &  & \orangebullet &  & \\
        \bottomrule
    \end{tabular}
    \caption{Mapping between each metric implementation type to the category commonly used, where the acronym \texttt{FVCU} corresponds to factuality, validity, coherence, and utility, respectively. For each combination of metric and implementation, \greenbullet\ denotes that there are at least 3 published works, and \orangebullet\ denotes that there are 1 or 2. The full table can be found in Table \ref{tab:metrics}.}
    \label{tab:metrics-to-criteria}
\end{table}
\newcommand{\sota}{$^{\textrm{\#1}}$}

\begin{table*}[htbp]
\centering \footnotesize
\begin{tabular}{>{\raggedright\arraybackslash}p{0.19\textwidth} >{\raggedright\arraybackslash}p{0.33\textwidth} >{\centering\arraybackslash}p{0.17\textwidth} > {\centering\arraybackslash}p{0.07\textwidth} p{0.15\textwidth}}
\toprule
\textbf{Type} & \textbf{Name} & \textbf{Domains} & \textbf{Criteria} & \textbf{Note} \\ 
\midrule
\multirow{2}{*}{\textbf{Rule-based}} 
    & DiVeRSe$^E$ \citep{li-etal-2023-making} & Arith & \texttt{VU} & Computation graph \\
    & Direct Evaluation \citep{nguyen-etal-2024-direct} & Factual & \texttt{FVCU} & Knowledge graph \\
\midrule

\multirow{5}{*}{\textbf{Uncertainty}}
    & UoT \citep{hu2024uncertainty} & Common, Expert & \texttt{U} &  \\
    & Entropy-based decoding \citep{qiu2024entropybaseddecodingretrievalaugmentedlarge} & Factual & \texttt{F} & \\
    & Semantic entropy probes \citep{farquhar2024detecting, kossen2024semanticentropyprobesrobust} & Factual, Common & \texttt{F} & \\
    & SynCheck$^E$ \citep{wu-etal-2024-synchronous} & Factual & \texttt{F} & \\
    & UnCert-CoT \citep{zhu2025uncertaintyguidedchainofthoughtcodegeneration} & Code & \texttt{V} & \\
\midrule

\multirow{3}{*}{\textbf{$\mathcal{V}$-information}} 
    & REV \citep{chen-etal-2023-rev} & Common & \texttt{U} & \\
    & ReCEval$^E$ \citep{prasad-etal-2023-receval} & Common, Arith & \texttt{CVU} & \\
    & EPVI \citep{wang-etal-2024-analyzing} & Arith, Common & \texttt{U} & \\
\midrule

\multirow{9}{*}{\textbf{LLM-as-value-function}} 
    & GenRM \citep{mahan2024generativerewardmodels} & Math, Logic, Code & \texttt{U} & \\
    & V-STaR \citep{hosseini2024vstartrainingverifiersselftaught} & Arith, Code & \texttt{U} & \\
    & MCTS-DPO \citep{xie2024montecarlotreesearch} & Math, Common, Science & \texttt{U} & \\
    & Step-DPO \citep{lai2024stepdpostepwisepreferenceoptimization} & Math & \texttt{U} & \\
    & Tree-PLV \citep{he-etal-2024-advancing} & Math, Common & \texttt{U} & \\
    & Step-Controlled DPO \citep{lu2024stepcontrolleddpoleveragingstepwise} & Math & \texttt{U} & \\
    & IRPO \citep{pang2024iterativereasoningpreferenceoptimization} & Math, Common & \texttt{U} & \\
    & PRIME \citep{cui2025processreinforcementimplicitrewards} & Math & \texttt{U} & \\
\midrule

\multirow{6}{*}{\textbf{Cross-encoders}} 
    & ROSCOE-LI \citep{DBLP:conf/iclr/GolovnevaCPCZFC23} & Common, Arith & \texttt{FVC} & Off-the-shelf \\
    & ReCEval$^E$ \citep{prasad-etal-2023-receval} & Common, Arith & \texttt{CVU} & Off-the-shelf \\
    & DiVeRSe$^E$ \citep{li-etal-2023-making} & Arith  & \texttt{VU} & Off-the-shelf \\
    & DBS \citep{zhu2024deductivebeamsearchdecoding} & Common, Arith, Symbolic & \texttt{FVCU} & Synthetic data \\
    & SynCheck$^E$ \citep{wu-etal-2024-synchronous} & Factual & \texttt{F} & Off-the-shelf \\
\midrule

\multirow{10}{*}{\textbf{Sequence Classifiers}} 
    & GSM8k-verifier \citep{cobbe2021trainingverifierssolvemath} & Arith & \texttt{U} & Outcome \\
    & PRM800K \citep{DBLP:conf/iclr/LightmanKBEBLLS24} & Math & \texttt{V} & Process \\
    & MATH-Minos \citep{gao2024llmcriticshelpcatch} & Math & \texttt{V} & Outcome/process\\
    & Math-Shepherd \citep{wang-etal-2024-math} & Math & \texttt{U} & Process \\
    & Eurus-PRM \citep{yuan2024implicitprm} & Math & \texttt{U} & Process \\
    & PAV \citep{setlur2024rewardingprogressscalingautomated} & Math & \texttt{U} & Process \\
    & ReasonEval \citep{xia2025evaluatingmathematicalreasoningaccuracy} & Math & \texttt{V} & Process \\
    & Qwen-PRM \citep{zhang2025lessonsdevelopingprocessreward} & Math, Science & \texttt{VU} & Process \\
    & VersaPRM \citep{zeng2025versaprmmultidomainprocessreward} & Expert & \texttt{FV} & Process \\
\midrule

\multirow{16}{*}{\textbf{Critic models}} 
    & Verify-CoT \citep{NEURIPS2023_72393bd4} & Math, Symbolic & \texttt{V} & Partial context \\
    & Tree-of-thoughts \citep{NEURIPS2023_271db992} & Arith, Common & \texttt{U} & No fine-tune \\
    & RAGTruth \citep{niu2024ragtruthhallucinationcorpusdeveloping} & Common & \texttt{F} & \\
    & CPO \citep{zhang2024chainpreferenceoptimizationimproving} & Factual, Arith & \texttt{U} & \\
    & F\textsuperscript{2}-Verification \citep{wang-etal-2024-boosting-language} & Common, Symbolic, Arith & \texttt{FV} & No fine-tune \\
    & OCEAN \citep{wu2024oceanofflinechainofthoughtevaluation} & Factual, Common & \texttt{F} & \\
    & Critic-CoT \citep{zheng2024criticcotboostingreasoningabilities} & Math, Common, Science, Logic & \texttt{U} & \\
    & AutoRace \citep{hao2024llmreasonersnewevaluation} & Math, Common, Logic & Custom & No fine-tune \\
    & R-PRM \citep{she2025rprmreasoningdrivenprocessreward} & Math & \texttt{V} & \\
    & PARC \citep{mukherjee2025premiseaugmentedreasoningchainsimprove} & Math & \texttt{V} & No fine-tune, \ \ \ Partial context \\
    & Reasoning evaluators \citep{kim2025scalingevaluationtimecomputereasoning} & Math, Science, Code & \texttt{V} & No fine-tune \\
    & ThinkPRM \citep{khalifa2025processrewardmodelsthink} & Math, Science, Code & \texttt{V} & \\
\midrule

\multirow{2}{*}{\textbf{Generative verifiers}} 
    & CLoud \citep{ankner2024critiqueoutloudrewardmodels} & Math, Logic, Code & \texttt{V} & \\
    & Generative verifier \citep{zhang2024generativeverifiersrewardmodeling} & Math, Symbolic & \texttt{V} & \\

\bottomrule
\end{tabular}
\caption{Evaluators for step-by-step reasoning, grouped by implementation type ($^E$ denotes that the method is an \textbf{e}nsemble of different methods). The \textbf{Domain} column specifies the domains used for meta-evaluating each evaluator. "Arith" refers to arithmetic reasoning tasks, "Common" denotes commonsense question answering, and "Expert" corresponds to specialized expert domains like biomedical, legal, and financial reasoning (Appendix \ref{sec:appendix-task}). When evaluators are tested on both arithmetic and general math tasks, only "Math" is listed. The acronym \texttt{FVCU} in the \textbf{Criteria} column represents factuality, validity, coherence, and utility, respectively. For AutoRace \citep{hao2024llmreasonersnewevaluation}, LLMs are instructed to list the criteria based on incorrect traces (Custom).}
\label{tab:metrics}
\end{table*}


The goal of reasoning trace evaluators is to assess reasoning traces by assigning scores. However, choosing the right evaluator for the target criteria and task is non-trivial. For instance, there is no guarantee that evaluators designed for factuality and multi-hop question answering will seamlessly work on math reasoning problems.

In this survey, we provide a comprehensive overview of diverse reasoning trace evaluators,  \citep{luo2024hallucination, wei2025surveyfeedbackbasedmultistepreasoning}. We summarize eight popular evaluator types based on the criteria they evaluate (summarized in Table \ref{tab:metrics-to-criteria}), along with other practical strengths and weaknesses.

\subsection{Rule-based matching}

For tasks where the ground truth solution can be expressed as a \textit{graph of entities}, a step corresponds to a directed edge between two entities, as in knowledge graphs for factual reasoning \citep{nguyen-etal-2024-direct} or computation graphs for arithmetic problems \citep{li-etal-2023-making}. In this setting, factuality reduces to identifying the correct relation between entities, coherence to the correct ordering of steps, and utility to the existence of the step in the gold reasoning chain \citep{nguyen-etal-2024-direct, PrOntoQA}. However, this approach does not generalize for tasks without clear symbolic representations, \textit{e.g.}, commonsense reasoning or complex math reasoning beyond arithmetic word problems.

\subsection{Intrinsic metrics}

\paragraph{Uncertainty} Uncertainty of the model can be used as an intrinsic proxy for the generated content's quality \citep{xiao-wang-2021-hallucination, zhang-etal-2023-enhancing-uncertainty}. \citet{qiu2024entropybaseddecodingretrievalaugmentedlarge} use \textit{token probability entropy}, defined as $\Sigma_{t\in V} p(t)\textrm{log}(p(t))$ where $p$ is the probability distribution of all tokens in vocabulary $V$. \citet{farquhar2024detecting} and \citet{kossen2024semanticentropyprobesrobust} extend the approach by clustering semantically similar tokens and calculating the entropy for each cluster.
While uncertainty-based evaluators are primarily used for factuality \citep{wu-etal-2024-synchronous, farquhar2024detecting}, they have also been applied for evaluating validity \citep{zhu2025uncertaintyguidedchainofthoughtcodegeneration} or utility \citep{hu2024uncertainty}, indicating that uncertainty can be a criteria-agnostic proxy of the quality of steps.

\paragraph{$\mathcal{V}$-information} \citet{chen-etal-2023-rev, prasad-etal-2023-receval} adopt $\mathcal{V}$-information (VI) \citep{hewitt-etal-2021-conditional} from information theory. Informally, VI measures if a model family $\mathcal{V}$ can generate the correct goal string $g$ with higher probability when the target string $t$ is given to the model. Formally, $\mathcal{V}I(t\rightarrow g) = \log p_\mathcal{V}(g|t) - \log p_\mathcal{V}(g|\phi)$ when $\phi$ is an empty string. When $g$ is the final answer and $t$ is the trace, VI becomes the difference between the answer token's probability between Chain-of-thought reasoning and zero-shot reasoning, which indicates how much information the trace provides to predicting the final answer (utility) \citep{chen-etal-2023-rev}. When $g$ is a step and $t$ is the list of previous steps, high VI means that a step is likely to follow from the context, which roughly corresponds to coherence \citep{prasad-etal-2023-receval}.

\paragraph{LLM-as-value-function} RL can train LLMs to align rewards to token probabilities (relative to the \textit{base probability} obtained from the initial model), with training objectives like DPO \citep{NEURIPS2023_a85b405e} and GRPO \citep{shao2024deepseekmathpushinglimitsmathematical}. For instance, when the reward is determined by the final answer correctness, the token probabilities directly correspond to utility \citep{mahan2024generativerewardmodels, lai2024stepdpostepwisepreferenceoptimization, xie2024montecarlotreesearch, pang2024iterativereasoningpreferenceoptimization}. Unlike sequence classifiers that lose their trace generation ability after fine-tuning, these models retain (and improve) the ability to generate traces. However, this method requires numerous good and bad reasoning traces for training. Consequently, most existing LLM-as-value-function evaluators focus on utility, as it is easier to scale up the data by simply checking if the final answer is correct.

\subsection{External evaluators}

\paragraph{Cross-encoders} Cross-encoders simultaneously encode multiple sentences using a single, small network often with millions of parameters \citep{devlin-etal-2019-bert, liu2019robertarobustlyoptimizedbert}. They have been widely applied to solve tasks such as natural language inference \citep{bowman-etal-2015-large} and fact verification \citep{thorne-etal-2018-fever}, where one has to determine if the \textit{hypothesis} can be inferred from the given \textit{premise}. Cross-encoders trained on off-the-shelf tasks \citep{DBLP:conf/iclr/GolovnevaCPCZFC23, zha-etal-2023-alignscore, prasad-etal-2023-receval} or LLM-perturbed data \citep{zhu2024deductivebeamsearchdecoding} can be used to evaluate a reasoning step based on the query (factuality) or previous steps (validity). However, their limited language understanding ability and shorter context length restrict their performance in more complex tasks.


\paragraph{Sequence classifiers (Reward Models)}\footnote{While \textit{reward model} generally refers to any model that predicts the desirability of an action in reinforcement learning, the term '(process/outcome) reward model' in the context of reasoning trace evaluation often refers to the sequence classifier architecture.} Sequence classifiers are language models with a lightweight classification head attached to the final hidden state, trained to predict a numeric score in a supervised manner \citep{DBLP:conf/iclr/LightmanKBEBLLS24, wang-etal-2024-math, setlur2024rewardingprogressscalingautomated}. Sequence classifiers can be further divided into (1) process (step-level) evaluator \textit{vs.} outcome (trace-level) evaluator based on the granularity of each step \citep{DBLP:conf/iclr/LightmanKBEBLLS24}, and (2) \textit{validity evaluator} \textit{vs.} \textit{utility evaluator} based on the source of the training data. These models achieve significant performance and efficiency \citep{cobbe2021trainingverifierssolvemath, zhang2025lessonsdevelopingprocessreward}, but they often require costly stepwise labels for training \citep{DBLP:conf/iclr/LightmanKBEBLLS24}. Furthermore, they cannot generate rationales for a high or a low score, having limited explainability and leading to spurious errors \citep{ankner2024critiqueoutloudrewardmodels, she2025rprmreasoningdrivenprocessreward}.

\paragraph{Critic models (LLM-as-a-judge)} Critic models are LLMs that are trained or prompted to evaluate the reasoning traces \citep{NEURIPS2023_91f18a12, DBLP:conf/iclr/KimS0JLLYSKTS24, zheng2024processbenchidentifyingprocesserrors, lin2024criticbenchbenchmarkingllmscritiquecorrect}. This approach views trace evaluation as one of many reasoning tasks, where common techniques like Chain-of-thought prompting \citep{huang2024largelanguagemodelsselfcorrect} and reinforcement learning with verifiable rewards \citep{chen2025rmr1rewardmodelingreasoning} can apply. Numerous works show that LLMs are versatile critics; they can effectively evaluate factuality, validity, coherence, and utility in diverse reasoning tasks with or without fine-tuning \citep{NEURIPS2023_271db992, jacovi-etal-2024-chain, wu-etal-2024-mitigating, niu2024ragtruthhallucinationcorpusdeveloping}. While conceptually simple and compatible with closed-source models, generating the rationales requires significant execution time and computation compared to other evaluator types.  


\paragraph{Generative Verifiers} This paradigm lies in the middle ground of sequence classifiers and critic models. These models first autoregressively generate the evaluation rationale as critic models do. When the generation terminates, like sequence classifiers, a small, fine-tuned head predicts the scores conditioned on both the original reasoning trace and evaluation rationales generated by itself \citep{ankner2024critiqueoutloudrewardmodels, zhang2024generativeverifiersrewardmodeling}.

\section{Further improving evaluators}
\label{sec:analysis}

This section discusses some of the recent empirical findings on improving evaluators beyond choosing different types, \textit{e.g.}, training data, input format, and scaling compute.


\paragraph{Validity and utility are complementary} Validity measures if the step is logically correct, while utility measures if the step makes progress towards the correct answer. Initially, utility-based process reward models were proposed as an \textit{alternative} for validity, since constructing validity data often requires a costly annotation process \citep{wang-etal-2024-math}. Under the hood, there lies an implicit assumption that useful steps are mostly valid.

\begin{figure}
    \centering
    \includegraphics[width=\linewidth]{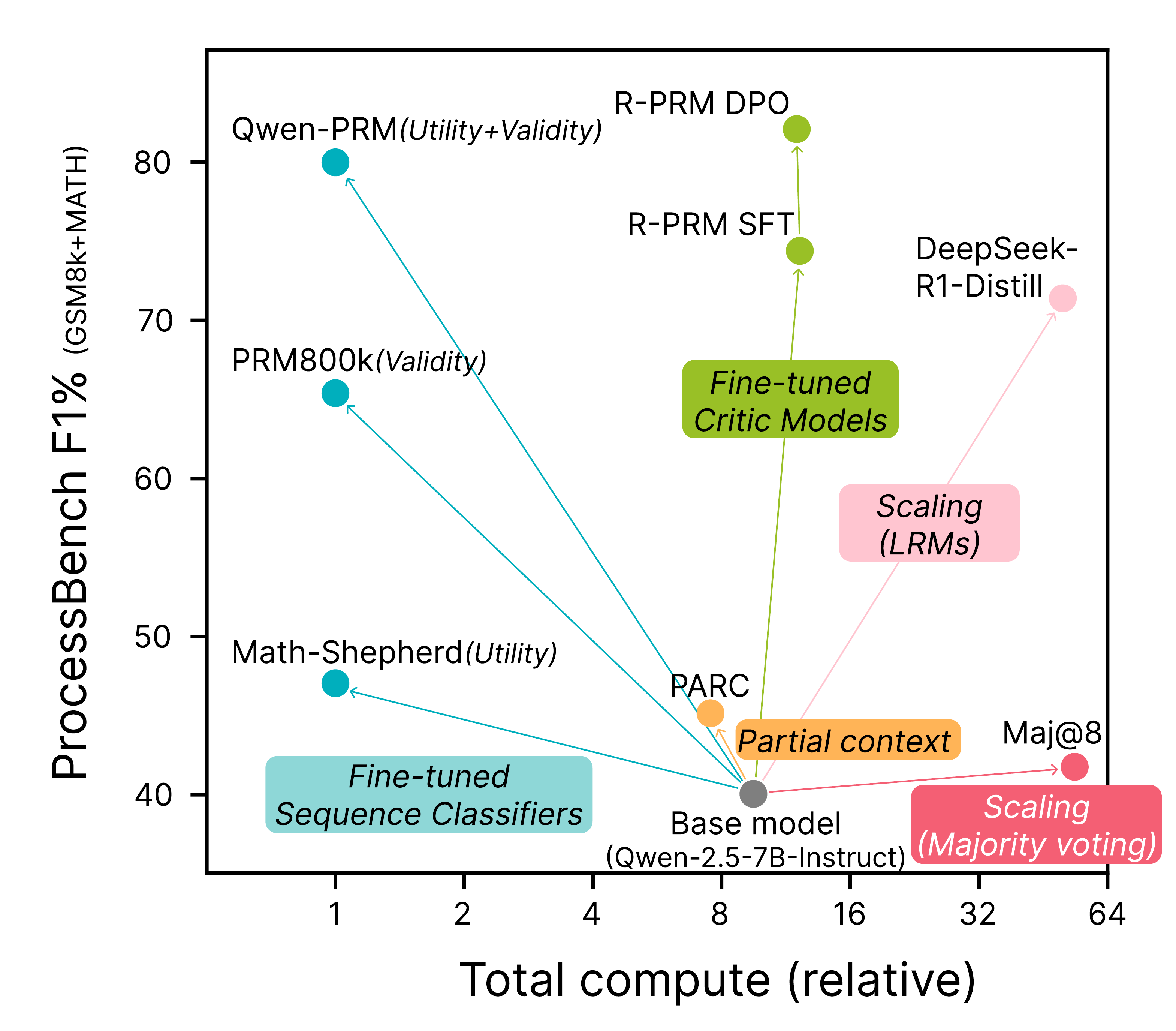}
    \caption{Plot of different evaluators introduced in Section \ref{sec:analysis}, plotted by ProcessBench performance \citep{zheng2024processbenchidentifyingprocesserrors} (GSM8k, MATH subsets averaged) versus total compute for evaluating a trace. While these evaluators share the same base model (Qwen-2.5-7B), they improve the base model's trace evaluation capability in different ways. Details can be found in Appendix \ref{sec:appendix-analysis}.}
    \label{fig:compute-to-processbench}
\end{figure}

However, recent works show that the two criteria are rather \textit{complementary}, from training sequence classifiers to using critic models. \citet{zhang2025lessonsdevelopingprocessreward} trains a sequence classifier by only considering steps that are both valid (judged by critic models) and useful (by MCTS-based rollouts) steps as positive, substantially improving performance over baselines that only consider validity or utility (Figure \ref{fig:compute-to-processbench}, \textit{Sequence Classifiers}). \citet{sun2024easytohardgeneralizationscalablealignment, kim2025scalingevaluationtimecomputereasoning} has also shown that averaging validity and utility scores from critic models results in better performance than individual scores in Best-of-N decoding.

The misalignment between validity and utility is mainly caused by steps that are logically wrong but reach the correct answer \citep{zheng2024processbenchidentifyingprocesserrors, wang2025examiningfalsepositivesinference, kim2025scalingevaluationtimecomputereasoning}. These invalid but useful steps, also known as \textit{unfaithful reasoning} of LLMs \citep{lyu-etal-2023-faithful, schnitzler2024morehopqa}, might lead to overestimation of reasoning ability \citep{lyu-etal-2023-faithful, petrov2025proofbluffevaluatingllms}.

\paragraph{Partial context allows efficient and accurate evaluation} Validity and coherence evaluate a step based on its previous steps. The most intuitive way is to use the \textit{full context} (all preceding steps). However, this approach is not feasible when the trace exceeds the context length of the evaluators, \textit{e.g.}, large reasoning models' traces are often too long to apply critic models \citep{kim2025scalingevaluationtimecomputereasoning}.

An alternative solution is to use a \textit{partial context}, where only relevant parts of the query and preceding steps are selected and passed to the evaluator \citep{NEURIPS2023_72393bd4, mukherjee2025premiseaugmentedreasoningchainsimprove}. These works first construct a directed \textit{entailment graph}, and evaluate the step only based on the identified premises. This allows evaluators to use shorter context, which is both computationally efficient \citep{NEURIPS2023_72393bd4} and even more accurate as distractors are removed from the context \citep{mukherjee2025premiseaugmentedreasoningchainsimprove} (Figure \ref{fig:compute-to-processbench}, \textit{Partial context}). Moreover, the graph structure also distinguishes direct errors (premises are valid but the reasoning is invalid) and accumulated errors (premises are invalid but reasoning is valid) \citep{mukherjee2025premiseaugmentedreasoningchainsimprove}. 


\paragraph{Test-time scaling improves evaluator performance} Test-time scaling is a general paradigm where investing more test-time compute leads to improved performance. Test-time compute can be scaled in diverse directions, such as sampling the output multiple times \citep{DBLP:conf/iclr/0002WSLCNCZ23, NEURIPS2023_271db992} or generating more tokens during a single inference \citep{snell2024scalingllmtesttimecompute, qwenlmQwQReflect, deepseekai2025deepseekr1incentivizingreasoningcapability}.

This paradigm can be extended to critic models that reason on reasoning traces, especially in meta-evaluation benchmarks. When applying majority voting of independently sampled $K$ evaluator scores in generative models (critic model, generative verifiers), the accuracy in predicting incorrect steps increases linearly with the scale of $\log K$ \citep{singhi2025solveverifycomputeoptimalproblem, kim2025scalingevaluationtimecomputereasoning, she2025rprmreasoningdrivenprocessreward} (Figure \ref{fig:compute-to-processbench}, \textit{Majority voting}). Furthermore, using large reasoning models (LRMs) with stronger reasoning capability by generating longer traces \citep{zheng2024processbenchidentifyingprocesserrors, kim2025scalingevaluationtimecomputereasoning, khalifa2025processrewardmodelsthink} leads to significant improvement in error detection (Figure \ref{fig:compute-to-processbench}, \textit{Scaling (LRMs)}).

In verifier-guided search settings, one can either scale exploration or scale evaluation. For instance, in Best-of-N decoding, one can increase the number of responses or use critic models that produce longer outputs. What is the optimal strategy with a constrained computing budget? For relatively weaker evaluators, simple majority voting \citep{DBLP:conf/iclr/0002WSLCNCZ23} often outperforms verifier-guided search \citep{zhang2025lessonsdevelopingprocessreward, singhi2025solveverifycomputeoptimalproblem}. However, using stronger evaluators, \textit{e.g.}, sequence classifiers with better training data \citep{zhang2025lessonsdevelopingprocessreward} or critic models with stronger reasoning capabilities \citep{khalifa2025processrewardmodelsthink, kim2025scalingevaluationtimecomputereasoning} for Best-of-N can effectively outperform majority voting using the same computation budget.
\section{Future directions}
\label{sec:future-directions}


\paragraph{Evaluating real-world reasoning traces with external knowledge} Existing datasets for reasoning trace evaluation are mostly restricted to simple factual reasoning (\textit{e.g.} factual multi-hop question answering) or self-contained problems (\textit{e.g.} math problems). However, many realistic reasoning tasks such as repository-level coding \citep{zhang-etal-2023-repocoder}, medicine \citep{savage2024diagnostic}, and law \citep{holzenberger-van-durme-2021-factoring, kimyeeun-etal-2024-developing} require external up-to-date knowledge retrieval-augmented generation \citep{NEURIPS2020_6b493230}. Developing evaluators and meta-evaluation benchmarks for these tasks will significantly enhance the applicability of reasoning trace evaluation in more realistic scenarios. 

\paragraph{Evaluating long, complex reasoning traces} Following OpenAI \texttt{o1} \citep{openai2024openaio1card}, numerous large reasoning models (LRMs) that generate long, complex traces involving self-verification and backtracking were introduced \citep{deepseekai2025deepseekr1incentivizingreasoningcapability, muennighoff2025s1simpletesttimescaling, gandhi2025cognitivebehaviorsenableselfimproving}. However, existing evaluators are not suitable for these complex traces. For instance, assigning a single scalar score (\textit{e.g.}, sequence classifiers) will make invalid steps corrected afterwards (\textit{Wait, this reasoning is not correct.}) and ones not corrected indistinguishable. Since LRM reasoning traces can contain critical errors \citep{petrov2025proofbluffevaluatingllms, anthropic2025reasoningmodels}, the effort to develop evaluation resources for such traces will lead to a better understanding of LRMs' behaviors and further improvement in their performance and credibility.

\paragraph{Advanced methods for finding premises.} NLI-based validity and coherence evaluation significantly benefit from determining the previous steps that the current step uses as a premise \citep{mukherjee2025premiseaugmentedreasoningchainsimprove}. However, finding such steps is not a trivial task. ROSCOE \citep{DBLP:conf/iclr/GolovnevaCPCZFC23} uses the minimum NLI score of all (previous step, current step) combinations, which ignores cases where a step has multiple premises. Recent works \citep{NEURIPS2023_72393bd4, tyen-etal-2024-llms, mukherjee2025premiseaugmentedreasoningchainsimprove} make the reasoner LLM annotate the premises of the given step. Plausible but underexplored approaches include applying \textit{uncertainty-based methods} \citep{chen-etal-2023-rev, wu-etal-2024-synchronous} or training a parser that annotates the logical dependencies between steps as graphs \citep{lee2025reasoningflowsemanticstructurecomplex}.

\paragraph{Symbol-grounded evaluation of reasoning traces} Reasoning tasks often have a symbolic ground truth solution. For instance, deductive reasoning tasks can be represented with formal logic, and arithmetic problems can be expressed as a series of equations or symbolic theorems. These solutions provide precise, formal ways to define evaluators, including validity and utility (progress). However, not much work has been done to exploit the parallel between reasoning traces and the underlying symbolic solution. While several rule-based approaches parse reasoning traces for evaluation in relatively simpler reasoning tasks \citep{PrOntoQA, nguyen-etal-2024-direct, li-etal-2023-making}, no attempts have been made to extend this paradigm to evaluate reasoning traces for more complex and realistic tasks like first-order logic reasoning \citep{han-etal-2024-folio, han-etal-2024-p} and formal math reasoning that use interactive theorem provers (\textit{e.g.}, Lean, Isabelle) \citep{yang2023leandojo, gao2024heraldnaturallanguageannotated}.


\paragraph{Rubric-based evaluation for complex and expert-level tasks.} Existing evaluators often apply \textit{identical} evaluations for all reasoning trace, \textit{e.g.}, using the same LLM-as-a-judge prompt for all inputs. However, as the reasoning tasks require more domain knowledge and expertise, there is an increasing need for highly specific \textbf{rubrics} for evaluating reasoning traces \citep{kim2024biggenbenchprincipledbenchmark}. For instance, one can calculate the sum of an arithmetic sequence by adding all terms one by one or finding a general term; the problem-specific rubrics explicitly prefer the latter. However, manual rubric generation is costly and less scalable, which motivates automatic extraction/generation of high-quality reasoning trace rubrics. AutoRace \citep{hao2024llmreasonersnewevaluation} aims to generate rubrics automatically based on incorrect responses, while RaR \citep{gunjal2025rubricsrewardsreinforcementlearning} extracts checklist-style rubrics from ground-truth biomedical documents. Still, automatically obtaining expert-level, high-quality rubrics for more diverse reasoning tasks remains an open question.
\section{Conclusion}
\label{sec:conclusion}

This survey aims to organize existing criteria and methods for step-by-step reasoning evaluation, which is crucial for understanding and improving LLM's reasoning capabilities. 
We provide a unified taxonomy for evaluation criteria, a comprehensive review of existing evaluators and their implementation, and examine recent directions on how to improve these evaluators.

Still, diverse challenges remain in evaluating step-by-step reasoning traces. As new reasoning tasks and methods emerge, existing evaluators often become obsolete for evaluating complex reasoning traces from new tasks and models. As LLMs are now involved in challenging and high-stakes reasoning tasks in the real world, understanding the nature of their errors and precisely evaluating the reasoning trace will remain important.


\section{Limitation}

\paragraph{References} This survey includes an extensive list of recent publications (mostly between 2022 and 2025) on reasoning trace evaluation, sourced from *ACL, EMNLP, NeurIPS, and arXiv preprints, \textit{etc}. While there might be missing references due to the sheer volume of works produced in this field, we will continue to update missing references and newly released impactful works that contribute to the field.

\paragraph{Survey on diverse empirical results} While Figure \ref{fig:compute-to-processbench} contains a controlled comparison between different approaches like training sequence classifier with different data, using partial context, or applying test-time scaling techniques, the comparison is limited to ProcessBench results for two reasons: (1) While most paper report reasoning performance improvement results (Section \ref{sec:meta-eval}), these results are often not directly comparable because they make use of different base model, which strongly affect the overall performance. (2) Other meta-evaluation benchmarks than ProcessBench \citep{jacovi-etal-2024-chain, zeng2024mrgsm8kmetareasoningbenchmarklarge, song2025prmbenchfinegrainedchallengingbenchmark} have not been applied to diverse evaluator implementations at the time of writing. 

\section{Acknowledgements}

We thank Sagnik Mukherjee for his valuable help in revising the paper and sharing data for PARC in Section 6, which greatly improved the completeness of this work.


\bibliography{custom}

\begin{thebibliography}{175}
\providecommand{\natexlab}[1]{#1}

\bibitem[{Akbar et~al.(2024)Akbar, Hossain, Wood, Chin, Salinas, Alvarez, and Cornejo}]{akbar-etal-2024-hallumeasure}
Shayan~Ali Akbar, Md~Mosharaf Hossain, Tess Wood, Si-Chi Chin, Erica~M Salinas, Victor Alvarez, and Erwin Cornejo. 2024.
\newblock \href {https://doi.org/10.18653/v1/2024.emnlp-main.837} {{H}allu{M}easure: Fine-grained hallucination measurement using chain-of-thought reasoning}.
\newblock In \emph{Proceedings of the 2024 Conference on Empirical Methods in Natural Language Processing}, pages 15020--15037, Miami, Florida, USA. Association for Computational Linguistics.

\bibitem[{Amini et~al.(2019)Amini, Gabriel, Lin, Koncel-Kedziorski, Choi, and Hajishirzi}]{amini-etal-2019-mathqa}
Aida Amini, Saadia Gabriel, Shanchuan Lin, Rik Koncel-Kedziorski, Yejin Choi, and Hannaneh Hajishirzi. 2019.
\newblock \href {https://doi.org/10.18653/v1/N19-1245} {{M}ath{QA}: Towards interpretable math word problem solving with operation-based formalisms}.
\newblock In \emph{Proceedings of the 2019 Conference of the North {A}merican Chapter of the Association for Computational Linguistics: Human Language Technologies, Volume 1 (Long and Short Papers)}, pages 2357--2367, Minneapolis, Minnesota. Association for Computational Linguistics.

\bibitem[{Anil et~al.(2023)Anil, Dai, Firat, Johnson, Lepikhin, Passos, Shakeri, Taropa, Bailey, Chen, Chu, Clark, Shafey, Huang, Meier-Hellstern, Mishra, Moreira, Omernick, Robinson, Ruder, Tay, Xiao, Xu, Zhang, Abrego, Ahn, Austin, Barham, Botha, Bradbury, Brahma, Brooks, Catasta, Cheng, Cherry, Choquette-Choo, Chowdhery, Crepy, Dave, Dehghani, Dev, Devlin, Díaz, Du, Dyer, Feinberg, Feng, Fienber, Freitag, Garcia, Gehrmann, Gonzalez, Gur-Ari, Hand, Hashemi, Hou, Howland, Hu, Hui, Hurwitz, Isard, Ittycheriah, Jagielski, Jia, Kenealy, Krikun, Kudugunta, Lan, Lee, Lee, Li, Li, Li, Li, Li, Lim, Lin, Liu, Liu, Maggioni, Mahendru, Maynez, Misra, Moussalem, Nado, Nham, Ni, Nystrom, Parrish, Pellat, Polacek, Polozov, Pope, Qiao, Reif, Richter, Riley, Ros, Roy, Saeta, Samuel, Shelby, Slone, Smilkov, So, Sohn, Tokumine, Valter, Vasudevan, Vodrahalli, Wang, Wang, Wang, Wang, Wieting, Wu, Xu, Xu, Xue, Yin, Yu, Zhang, Zheng, Zheng, Zhou, Zhou, Petrov, and Wu}]{anil2023palm2technicalreport}
Rohan Anil, Andrew~M. Dai, Orhan Firat, Melvin Johnson, Dmitry Lepikhin, Alexandre Passos, Siamak Shakeri, Emanuel Taropa, Paige Bailey, Zhifeng Chen, Eric Chu, Jonathan~H. Clark, Laurent~El Shafey, Yanping Huang, Kathy Meier-Hellstern, Gaurav Mishra, Erica Moreira, Mark Omernick, Kevin Robinson, Sebastian Ruder, Yi~Tay, Kefan Xiao, Yuanzhong Xu, Yujing Zhang, Gustavo~Hernandez Abrego, Junwhan Ahn, Jacob Austin, Paul Barham, Jan Botha, James Bradbury, Siddhartha Brahma, Kevin Brooks, Michele Catasta, Yong Cheng, Colin Cherry, Christopher~A. Choquette-Choo, Aakanksha Chowdhery, Clément Crepy, Shachi Dave, Mostafa Dehghani, Sunipa Dev, Jacob Devlin, Mark Díaz, Nan Du, Ethan Dyer, Vlad Feinberg, Fangxiaoyu Feng, Vlad Fienber, Markus Freitag, Xavier Garcia, Sebastian Gehrmann, Lucas Gonzalez, Guy Gur-Ari, Steven Hand, Hadi Hashemi, Le~Hou, Joshua Howland, Andrea Hu, Jeffrey Hui, Jeremy Hurwitz, Michael Isard, Abe Ittycheriah, Matthew Jagielski, Wenhao Jia, Kathleen Kenealy, Maxim Krikun, Sneha Kudugunta, Chang
  Lan, Katherine Lee, Benjamin Lee, Eric Li, Music Li, Wei Li, YaGuang Li, Jian Li, Hyeontaek Lim, Hanzhao Lin, Zhongtao Liu, Frederick Liu, Marcello Maggioni, Aroma Mahendru, Joshua Maynez, Vedant Misra, Maysam Moussalem, Zachary Nado, John Nham, Eric Ni, Andrew Nystrom, Alicia Parrish, Marie Pellat, Martin Polacek, Alex Polozov, Reiner Pope, Siyuan Qiao, Emily Reif, Bryan Richter, Parker Riley, Alex~Castro Ros, Aurko Roy, Brennan Saeta, Rajkumar Samuel, Renee Shelby, Ambrose Slone, Daniel Smilkov, David~R. So, Daniel Sohn, Simon Tokumine, Dasha Valter, Vijay Vasudevan, Kiran Vodrahalli, Xuezhi Wang, Pidong Wang, Zirui Wang, Tao Wang, John Wieting, Yuhuai Wu, Kelvin Xu, Yunhan Xu, Linting Xue, Pengcheng Yin, Jiahui Yu, Qiao Zhang, Steven Zheng, Ce~Zheng, Weikang Zhou, Denny Zhou, Slav Petrov, and Yonghui Wu. 2023.
\newblock \href {https://arxiv.org/abs/2305.10403} {Palm 2 technical report}.
\newblock \emph{Preprint}, arXiv:2305.10403.

\bibitem[{Ankner et~al.(2024)Ankner, Paul, Cui, Chang, and Ammanabrolu}]{ankner2024critiqueoutloudrewardmodels}
Zachary Ankner, Mansheej Paul, Brandon Cui, Jonathan~D. Chang, and Prithviraj Ammanabrolu. 2024.
\newblock \href {https://arxiv.org/abs/2408.11791} {Critique-out-loud reward models}.
\newblock \emph{Preprint}, arXiv:2408.11791.

\bibitem[{Banerjee et~al.(2024)Banerjee, Agarwal, and Singla}]{banerjee2024llmshallucinateneedlive}
Sourav Banerjee, Ayushi Agarwal, and Saloni Singla. 2024.
\newblock \href {https://arxiv.org/abs/2409.05746} {Llms will always hallucinate, and we need to live with this}.
\newblock \emph{Preprint}, arXiv:2409.05746.

\bibitem[{Bisk et~al.(2019)Bisk, Zellers, Bras, Gao, and Choi}]{bisk2019piqareasoningphysicalcommonsense}
Yonatan Bisk, Rowan Zellers, Ronan~Le Bras, Jianfeng Gao, and Yejin Choi. 2019.
\newblock \href {https://arxiv.org/abs/1911.11641} {Piqa: Reasoning about physical commonsense in natural language}.
\newblock \emph{Preprint}, arXiv:1911.11641.

\bibitem[{Bowman et~al.(2015)Bowman, Angeli, Potts, and Manning}]{bowman-etal-2015-large}
Samuel~R. Bowman, Gabor Angeli, Christopher Potts, and Christopher~D. Manning. 2015.
\newblock \href {https://doi.org/10.18653/v1/D15-1075} {A large annotated corpus for learning natural language inference}.
\newblock In \emph{Proceedings of the 2015 Conference on Empirical Methods in Natural Language Processing}, pages 632--642, Lisbon, Portugal. Association for Computational Linguistics.

\bibitem[{Brown et~al.(2020)Brown, Mann, Ryder, Subbiah, Kaplan, Dhariwal, Neelakantan, Shyam, Sastry, Askell, Agarwal, Herbert-Voss, Krueger, Henighan, Child, Ramesh, Ziegler, Wu, Winter, Hesse, Chen, Sigler, Litwin, Gray, Chess, Clark, Berner, McCandlish, Radford, Sutskever, and Amodei}]{brown2020languagemodelsfewshotlearners}
Tom~B. Brown, Benjamin Mann, Nick Ryder, Melanie Subbiah, Jared Kaplan, Prafulla Dhariwal, Arvind Neelakantan, Pranav Shyam, Girish Sastry, Amanda Askell, Sandhini Agarwal, Ariel Herbert-Voss, Gretchen Krueger, Tom Henighan, Rewon Child, Aditya Ramesh, Daniel~M. Ziegler, Jeffrey Wu, Clemens Winter, Christopher Hesse, Mark Chen, Eric Sigler, Mateusz Litwin, Scott Gray, Benjamin Chess, Jack Clark, Christopher Berner, Sam McCandlish, Alec Radford, Ilya Sutskever, and Dario Amodei. 2020.
\newblock \href {https://arxiv.org/abs/2005.14165} {Language models are few-shot learners}.
\newblock \emph{Preprint}, arXiv:2005.14165.

\bibitem[{Camburu et~al.(2018)Camburu, Rockt\"{a}schel, Lukasiewicz, and Blunsom}]{NEURIPS2018_4c7a167b}
Oana-Maria Camburu, Tim Rockt\"{a}schel, Thomas Lukasiewicz, and Phil Blunsom. 2018.
\newblock \href {https://proceedings.neurips.cc/paper_files/paper/2018/file/4c7a167bb329bd92580a99ce422d6fa6-Paper.pdf} {e-snli: Natural language inference with natural language explanations}.
\newblock In \emph{Advances in Neural Information Processing Systems}, volume~31. Curran Associates, Inc.

\bibitem[{Chan et~al.(2024)Chan, Chowdhury, Jaffe, Aung, Sherburn, Mays, Starace, Liu, Maksin, Patwardhan, Weng, and Madry}]{DBLP:journals/corr/abs-2410-07095}
Jun~Shern Chan, Neil Chowdhury, Oliver Jaffe, James Aung, Dane Sherburn, Evan Mays, Giulio Starace, Kevin Liu, Leon Maksin, Tejal Patwardhan, Lilian Weng, and Aleksander Madry. 2024.
\newblock \href {https://doi.org/10.48550/ARXIV.2410.07095} {Mle-bench: Evaluating machine learning agents on machine learning engineering}.
\newblock \emph{CoRR}, abs/2410.07095.

\bibitem[{Chen et~al.(2023)Chen, Brahman, Ren, Ji, Choi, and Swayamdipta}]{chen-etal-2023-rev}
Hanjie Chen, Faeze Brahman, Xiang Ren, Yangfeng Ji, Yejin Choi, and Swabha Swayamdipta. 2023.
\newblock \href {https://doi.org/10.18653/v1/2023.acl-long.112} {{REV}: Information-theoretic evaluation of free-text rationales}.
\newblock In \emph{Proceedings of the 61st Annual Meeting of the Association for Computational Linguistics (Volume 1: Long Papers)}, pages 2007--2030, Toronto, Canada. Association for Computational Linguistics.

\bibitem[{Chen et~al.(2021)Chen, Tworek, Jun, Yuan, de~Oliveira~Pinto, Kaplan, Edwards, Burda, Joseph, Brockman, Ray, Puri, Krueger, Petrov, Khlaaf, Sastry, Mishkin, Chan, Gray, Ryder, Pavlov, Power, Kaiser, Bavarian, Winter, Tillet, Such, Cummings, Plappert, Chantzis, Barnes, Herbert-Voss, Guss, Nichol, Paino, Tezak, Tang, Babuschkin, Balaji, Jain, Saunders, Hesse, Carr, Leike, Achiam, Misra, Morikawa, Radford, Knight, Brundage, Murati, Mayer, Welinder, McGrew, Amodei, McCandlish, Sutskever, and Zaremba}]{chen2021evaluatinglargelanguagemodels}
Mark Chen, Jerry Tworek, Heewoo Jun, Qiming Yuan, Henrique~Ponde de~Oliveira~Pinto, Jared Kaplan, Harri Edwards, Yuri Burda, Nicholas Joseph, Greg Brockman, Alex Ray, Raul Puri, Gretchen Krueger, Michael Petrov, Heidy Khlaaf, Girish Sastry, Pamela Mishkin, Brooke Chan, Scott Gray, Nick Ryder, Mikhail Pavlov, Alethea Power, Lukasz Kaiser, Mohammad Bavarian, Clemens Winter, Philippe Tillet, Felipe~Petroski Such, Dave Cummings, Matthias Plappert, Fotios Chantzis, Elizabeth Barnes, Ariel Herbert-Voss, William~Hebgen Guss, Alex Nichol, Alex Paino, Nikolas Tezak, Jie Tang, Igor Babuschkin, Suchir Balaji, Shantanu Jain, William Saunders, Christopher Hesse, Andrew~N. Carr, Jan Leike, Josh Achiam, Vedant Misra, Evan Morikawa, Alec Radford, Matthew Knight, Miles Brundage, Mira Murati, Katie Mayer, Peter Welinder, Bob McGrew, Dario Amodei, Sam McCandlish, Ilya Sutskever, and Wojciech Zaremba. 2021.
\newblock \href {https://arxiv.org/abs/2107.03374} {Evaluating large language models trained on code}.
\newblock \emph{Preprint}, arXiv:2107.03374.

\bibitem[{Chen et~al.(2025{\natexlab{a}})Chen, Li, Wang, Jin, Qian, Wang, Wang, Zhang, Zhang, Zhang, Tong, and Ji}]{chen2025rmr1rewardmodelingreasoning}
Xiusi Chen, Gaotang Li, Ziqi Wang, Bowen Jin, Cheng Qian, Yu~Wang, Hongru Wang, Yu~Zhang, Denghui Zhang, Tong Zhang, Hanghang Tong, and Heng Ji. 2025{\natexlab{a}}.
\newblock \href {https://arxiv.org/abs/2505.02387} {Rm-r1: Reward modeling as reasoning}.
\newblock \emph{Preprint}, arXiv:2505.02387.

\bibitem[{Chen et~al.(2025{\natexlab{b}})Chen, Benton, Radhakrishnan, Uesato, Denison, Schulman, Somani, Hase, Wagner, Roger, Mikulik, Bowman, Leike, Kaplan, and Perez}]{anthropic2025reasoningmodels}
Yanda Chen, Joe Benton, Ansh Radhakrishnan, Jonathan Uesato, Carson Denison, John Schulman, Arushi Somani, Peter Hase, Misha Wagner, Fabien Roger, Vlad Mikulik, Sam Bowman, Jan Leike, Jared Kaplan, and Ethan Perez. 2025{\natexlab{b}}.
\newblock \href {https://assets.anthropic.com/m/71876fabef0f0ed4/original/reasoning_models_paper.pdf} {Reasoning models don’t always say what they think}.

\bibitem[{Chen et~al.(2022)Chen, Chen, Smiley, Shah, Borova, Langdon, Moussa, Beane, Huang, Routledge, and Wang}]{chen2022finqadatasetnumericalreasoning}
Zhiyu Chen, Wenhu Chen, Charese Smiley, Sameena Shah, Iana Borova, Dylan Langdon, Reema Moussa, Matt Beane, Ting-Hao Huang, Bryan Routledge, and William~Yang Wang. 2022.
\newblock \href {https://arxiv.org/abs/2109.00122} {Finqa: A dataset of numerical reasoning over financial data}.
\newblock \emph{Preprint}, arXiv:2109.00122.

\bibitem[{Chiang and Lee(2024)}]{chiang-lee-2024-reasoning}
Cheng-Han Chiang and Hung-yi Lee. 2024.
\newblock \href {https://aclanthology.org/2024.eacl-short.15/} {Over-reasoning and redundant calculation of large language models}.
\newblock In \emph{Proceedings of the 18th Conference of the European Chapter of the Association for Computational Linguistics (Volume 2: Short Papers)}, pages 161--169, St. Julian{'}s, Malta. Association for Computational Linguistics.

\bibitem[{Clark et~al.(2018)Clark, Cowhey, Etzioni, Khot, Sabharwal, Schoenick, and Tafjord}]{clark2018thinksolvedquestionanswering}
Peter Clark, Isaac Cowhey, Oren Etzioni, Tushar Khot, Ashish Sabharwal, Carissa Schoenick, and Oyvind Tafjord. 2018.
\newblock \href {https://arxiv.org/abs/1803.05457} {Think you have solved question answering? try arc, the ai2 reasoning challenge}.
\newblock \emph{Preprint}, arXiv:1803.05457.

\bibitem[{Cobbe et~al.(2021)Cobbe, Kosaraju, Bavarian, Chen, Jun, Kaiser, Plappert, Tworek, Hilton, Nakano, Hesse, and Schulman}]{cobbe2021trainingverifierssolvemath}
Karl Cobbe, Vineet Kosaraju, Mohammad Bavarian, Mark Chen, Heewoo Jun, Lukasz Kaiser, Matthias Plappert, Jerry Tworek, Jacob Hilton, Reiichiro Nakano, Christopher Hesse, and John Schulman. 2021.
\newblock \href {https://arxiv.org/abs/2110.14168} {Training verifiers to solve math word problems}.
\newblock \emph{Preprint}, arXiv:2110.14168.

\bibitem[{Creswell and Shanahan(2022)}]{creswell2022faithfulreasoningusinglarge}
Antonia Creswell and Murray Shanahan. 2022.
\newblock \href {https://arxiv.org/abs/2208.14271} {Faithful reasoning using large language models}.
\newblock \emph{Preprint}, arXiv:2208.14271.

\bibitem[{Cui et~al.(2025)Cui, Yuan, Wang, Wang, Li, He, Fan, Yu, Xu, Chen, Yuan, Chen, Zhang, Lv, Wang, Yao, Han, Peng, Cheng, Liu, Sun, Zhou, and Ding}]{cui2025processreinforcementimplicitrewards}
Ganqu Cui, Lifan Yuan, Zefan Wang, Hanbin Wang, Wendi Li, Bingxiang He, Yuchen Fan, Tianyu Yu, Qixin Xu, Weize Chen, Jiarui Yuan, Huayu Chen, Kaiyan Zhang, Xingtai Lv, Shuo Wang, Yuan Yao, Xu~Han, Hao Peng, Yu~Cheng, Zhiyuan Liu, Maosong Sun, Bowen Zhou, and Ning Ding. 2025.
\newblock \href {https://arxiv.org/abs/2502.01456} {Process reinforcement through implicit rewards}.
\newblock \emph{Preprint}, arXiv:2502.01456.

\bibitem[{Dai et~al.(2025)Dai, Wu, Zheng, Wei, Shi, Jin, Liu, Dun, Huang, and Yan}]{dai2025processsupervisionguidedpolicyoptimization}
Ning Dai, Zheng Wu, Renjie Zheng, Ziyun Wei, Wenlei Shi, Xing Jin, Guanlin Liu, Chen Dun, Liang Huang, and Lin Yan. 2025.
\newblock \href {https://arxiv.org/abs/2410.17621} {Process supervision-guided policy optimization for code generation}.
\newblock \emph{Preprint}, arXiv:2410.17621.

\bibitem[{Dalvi et~al.(2021)Dalvi, Jansen, Tafjord, Xie, Smith, Pipatanangkura, and Clark}]{dalvi-etal-2021-explaining}
Bhavana Dalvi, Peter Jansen, Oyvind Tafjord, Zhengnan Xie, Hannah Smith, Leighanna Pipatanangkura, and Peter Clark. 2021.
\newblock \href {https://doi.org/10.18653/v1/2021.emnlp-main.585} {Explaining answers with entailment trees}.
\newblock In \emph{Proceedings of the 2021 Conference on Empirical Methods in Natural Language Processing}, pages 7358--7370, Online and Punta Cana, Dominican Republic. Association for Computational Linguistics.

\bibitem[{DeepSeek-AI(2025)}]{deepseekai2025deepseekr1incentivizingreasoningcapability}
DeepSeek-AI. 2025.
\newblock \href {https://arxiv.org/abs/2501.12948} {Deepseek-r1: Incentivizing reasoning capability in llms via reinforcement learning}.
\newblock \emph{Preprint}, arXiv:2501.12948.

\bibitem[{Devlin et~al.(2019)Devlin, Chang, Lee, and Toutanova}]{devlin-etal-2019-bert}
Jacob Devlin, Ming-Wei Chang, Kenton Lee, and Kristina Toutanova. 2019.
\newblock \href {https://doi.org/10.18653/v1/N19-1423} {{BERT}: Pre-training of deep bidirectional transformers for language understanding}.
\newblock In \emph{Proceedings of the 2019 Conference of the North {A}merican Chapter of the Association for Computational Linguistics: Human Language Technologies, Volume 1 (Long and Short Papers)}, pages 4171--4186, Minneapolis, Minnesota. Association for Computational Linguistics.

\bibitem[{Fan et~al.(2024)Fan, Hua, Li, Ling, and Zhang}]{fan2024nphardevaldynamicbenchmarkreasoning}
Lizhou Fan, Wenyue Hua, Lingyao Li, Haoyang Ling, and Yongfeng Zhang. 2024.
\newblock \href {https://arxiv.org/abs/2312.14890} {Nphardeval: Dynamic benchmark on reasoning ability of large language models via complexity classes}.
\newblock \emph{Preprint}, arXiv:2312.14890.

\bibitem[{Farquhar et~al.(2024)Farquhar, Kossen, Kuhn, and Gal}]{farquhar2024detecting}
Sebastian Farquhar, Jannik Kossen, Lorenz Kuhn, and Yarin Gal. 2024.
\newblock Detecting hallucinations in large language models using semantic entropy.
\newblock \emph{Nature}, 630(8017):625--630.

\bibitem[{Gandhi et~al.(2025)Gandhi, Chakravarthy, Singh, Lile, and Goodman}]{gandhi2025cognitivebehaviorsenableselfimproving}
Kanishk Gandhi, Ayush Chakravarthy, Anikait Singh, Nathan Lile, and Noah~D. Goodman. 2025.
\newblock \href {https://arxiv.org/abs/2503.01307} {Cognitive behaviors that enable self-improving reasoners, or, four habits of highly effective stars}.
\newblock \emph{Preprint}, arXiv:2503.01307.

\bibitem[{Gao et~al.(2024{\natexlab{a}})Gao, Cai, Xu, Wang, Zheng, Lin, Lu, Liu, Zhou, Xiao, Hu, Liu, and Chang}]{gao2024llmcriticshelpcatch}
Bofei Gao, Zefan Cai, Runxin Xu, Peiyi Wang, Ce~Zheng, Runji Lin, Keming Lu, Dayiheng Liu, Chang Zhou, Wen Xiao, Junjie Hu, Tianyu Liu, and Baobao Chang. 2024{\natexlab{a}}.
\newblock \href {https://arxiv.org/abs/2406.14024} {Llm critics help catch bugs in mathematics: Towards a better mathematical verifier with natural language feedback}.
\newblock \emph{Preprint}, arXiv:2406.14024.

\bibitem[{Gao et~al.(2024{\natexlab{b}})Gao, Song, Yang, Cai, Miao, Dong, Li, Ma, Chen, Xu, Tang, Wang, Zan, Quan, Zhang, Sha, Zhang, Ren, Liu, and Chang}]{gao2024omnimathuniversalolympiadlevel}
Bofei Gao, Feifan Song, Zhe Yang, Zefan Cai, Yibo Miao, Qingxiu Dong, Lei Li, Chenghao Ma, Liang Chen, Runxin Xu, Zhengyang Tang, Benyou Wang, Daoguang Zan, Shanghaoran Quan, Ge~Zhang, Lei Sha, Yichang Zhang, Xuancheng Ren, Tianyu Liu, and Baobao Chang. 2024{\natexlab{b}}.
\newblock \href {https://arxiv.org/abs/2410.07985} {Omni-math: A universal olympiad level mathematic benchmark for large language models}.
\newblock \emph{Preprint}, arXiv:2410.07985.

\bibitem[{Gao et~al.(2024{\natexlab{c}})Gao, Wang, Jiang, Gao, Qin, Xu, and Dong}]{gao2024heraldnaturallanguageannotated}
Guoxiong Gao, Yutong Wang, Jiedong Jiang, Qi~Gao, Zihan Qin, Tianyi Xu, and Bin Dong. 2024{\natexlab{c}}.
\newblock \href {https://arxiv.org/abs/2410.10878} {Herald: A natural language annotated lean 4 dataset}.
\newblock \emph{Preprint}, arXiv:2410.10878.

\bibitem[{Gao et~al.(2024{\natexlab{d}})Gao, Xiong, Gao, Jia, Pan, Bi, Dai, Sun, Wang, and Wang}]{gao2024retrievalaugmentedgenerationlargelanguage}
Yunfan Gao, Yun Xiong, Xinyu Gao, Kangxiang Jia, Jinliu Pan, Yuxi Bi, Yi~Dai, Jiawei Sun, Meng Wang, and Haofen Wang. 2024{\natexlab{d}}.
\newblock \href {https://arxiv.org/abs/2312.10997} {Retrieval-augmented generation for large language models: A survey}.
\newblock \emph{Preprint}, arXiv:2312.10997.

\bibitem[{Geva et~al.(2021)Geva, Khashabi, Segal, Khot, Roth, and Berant}]{geva-etal-2021-aristotle}
Mor Geva, Daniel Khashabi, Elad Segal, Tushar Khot, Dan Roth, and Jonathan Berant. 2021.
\newblock \href {https://doi.org/10.1162/tacl_a_00370} {Did aristotle use a laptop? a question answering benchmark with implicit reasoning strategies}.
\newblock \emph{Transactions of the Association for Computational Linguistics}, 9:346--361.

\bibitem[{Glazer et~al.(2024)Glazer, Erdil, Besiroglu, Chicharro, Chen, Gunning, Olsson, Denain, Ho, de~Oliveira~Santos, Järviniemi, Barnett, Sandler, Vrzala, Sevilla, Ren, Pratt, Levine, Barkley, Stewart, Grechuk, Grechuk, Enugandla, and Wildon}]{glazer2024frontiermathbenchmarkevaluatingadvanced}
Elliot Glazer, Ege Erdil, Tamay Besiroglu, Diego Chicharro, Evan Chen, Alex Gunning, Caroline~Falkman Olsson, Jean-Stanislas Denain, Anson Ho, Emily de~Oliveira~Santos, Olli Järviniemi, Matthew Barnett, Robert Sandler, Matej Vrzala, Jaime Sevilla, Qiuyu Ren, Elizabeth Pratt, Lionel Levine, Grant Barkley, Natalie Stewart, Bogdan Grechuk, Tetiana Grechuk, Shreepranav~Varma Enugandla, and Mark Wildon. 2024.
\newblock \href {https://arxiv.org/abs/2411.04872} {Frontiermath: A benchmark for evaluating advanced mathematical reasoning in ai}.
\newblock \emph{Preprint}, arXiv:2411.04872.

\bibitem[{Golovneva et~al.(2023{\natexlab{a}})Golovneva, Chen, Poff, Corredor, Zettlemoyer, Fazel{-}Zarandi, and Celikyilmaz}]{DBLP:conf/iclr/GolovnevaCPCZFC23}
Olga Golovneva, Moya Chen, Spencer Poff, Martin Corredor, Luke Zettlemoyer, Maryam Fazel{-}Zarandi, and Asli Celikyilmaz. 2023{\natexlab{a}}.
\newblock \href {https://openreview.net/forum?id=xYlJRpzZtsY} {{ROSCOE:} {A} suite of metrics for scoring step-by-step reasoning}.
\newblock In \emph{The Eleventh International Conference on Learning Representations, {ICLR} 2023, Kigali, Rwanda, May 1-5, 2023}. OpenReview.net.

\bibitem[{Golovneva et~al.(2023{\natexlab{b}})Golovneva, O'Brien, Pasunuru, Wang, Zettlemoyer, Fazel-Zarandi, and Celikyilmaz}]{golovneva2023pathfinderguidedsearchmultistep}
Olga Golovneva, Sean O'Brien, Ramakanth Pasunuru, Tianlu Wang, Luke Zettlemoyer, Maryam Fazel-Zarandi, and Asli Celikyilmaz. 2023{\natexlab{b}}.
\newblock \href {https://arxiv.org/abs/2312.05180} {Pathfinder: Guided search over multi-step reasoning paths}.
\newblock \emph{Preprint}, arXiv:2312.05180.

\bibitem[{Guan et~al.(2024)Guan, Liu, Lu, Cao, He, Han, Sun, Lou, Yu, Lu, and Lin}]{guan2024searchverifyfeedbackgeneration}
Xinyan Guan, Yanjiang Liu, Xinyu Lu, Boxi Cao, Ben He, Xianpei Han, Le~Sun, Jie Lou, Bowen Yu, Yaojie Lu, and Hongyu Lin. 2024.
\newblock \href {https://arxiv.org/abs/2411.11504} {Search, verify and feedback: Towards next generation post-training paradigm of foundation models via verifier engineering}.
\newblock \emph{Preprint}, arXiv:2411.11504.

\bibitem[{Guha et~al.(2023)Guha, Nyarko, Ho, Ré, Chilton, Narayana, Chohlas-Wood, Peters, Waldon, Rockmore, Zambrano, Talisman, Hoque, Surani, Fagan, Sarfaty, Dickinson, Porat, Hegland, Wu, Nudell, Niklaus, Nay, Choi, Tobia, Hagan, Ma, Livermore, Rasumov-Rahe, Holzenberger, Kolt, Henderson, Rehaag, Goel, Gao, Williams, Gandhi, Zur, Iyer, and Li}]{guha2023legalbenchcollaborativelybuiltbenchmark}
Neel Guha, Julian Nyarko, Daniel~E. Ho, Christopher Ré, Adam Chilton, Aditya Narayana, Alex Chohlas-Wood, Austin Peters, Brandon Waldon, Daniel~N. Rockmore, Diego Zambrano, Dmitry Talisman, Enam Hoque, Faiz Surani, Frank Fagan, Galit Sarfaty, Gregory~M. Dickinson, Haggai Porat, Jason Hegland, Jessica Wu, Joe Nudell, Joel Niklaus, John Nay, Jonathan~H. Choi, Kevin Tobia, Margaret Hagan, Megan Ma, Michael Livermore, Nikon Rasumov-Rahe, Nils Holzenberger, Noam Kolt, Peter Henderson, Sean Rehaag, Sharad Goel, Shang Gao, Spencer Williams, Sunny Gandhi, Tom Zur, Varun Iyer, and Zehua Li. 2023.
\newblock \href {https://arxiv.org/abs/2308.11462} {Legalbench: A collaboratively built benchmark for measuring legal reasoning in large language models}.
\newblock \emph{Preprint}, arXiv:2308.11462.

\bibitem[{Gunjal et~al.(2025)Gunjal, Wang, Lau, Nath, Liu, and Hendryx}]{gunjal2025rubricsrewardsreinforcementlearning}
Anisha Gunjal, Anthony Wang, Elaine Lau, Vaskar Nath, Bing Liu, and Sean Hendryx. 2025.
\newblock \href {https://arxiv.org/abs/2507.17746} {Rubrics as rewards: Reinforcement learning beyond verifiable domains}.
\newblock \emph{Preprint}, arXiv:2507.17746.

\bibitem[{Han et~al.(2024{\natexlab{a}})Han, Schoelkopf, Zhao, Qi, Riddell, Zhou, Coady, Peng, Qiao, Benson, Sun, Wardle-Solano, Szab{\'o}, Zubova, Burtell, Fan, Liu, Wong, Sailor, Ni, Nan, Kasai, Yu, Zhang, Fabbri, Kryscinski, Yavuz, Liu, Lin, Joty, Zhou, Xiong, Ying, Cohan, and Radev}]{han-etal-2024-folio}
Simeng Han, Hailey Schoelkopf, Yilun Zhao, Zhenting Qi, Martin Riddell, Wenfei Zhou, James Coady, David Peng, Yujie Qiao, Luke Benson, Lucy Sun, Alexander Wardle-Solano, Hannah Szab{\'o}, Ekaterina Zubova, Matthew Burtell, Jonathan Fan, Yixin Liu, Brian Wong, Malcolm Sailor, Ansong Ni, Linyong Nan, Jungo Kasai, Tao Yu, Rui Zhang, Alexander Fabbri, Wojciech~Maciej Kryscinski, Semih Yavuz, Ye~Liu, Xi~Victoria Lin, Shafiq Joty, Yingbo Zhou, Caiming Xiong, Rex Ying, Arman Cohan, and Dragomir Radev. 2024{\natexlab{a}}.
\newblock \href {https://doi.org/10.18653/v1/2024.emnlp-main.1229} {{FOLIO}: Natural language reasoning with first-order logic}.
\newblock In \emph{Proceedings of the 2024 Conference on Empirical Methods in Natural Language Processing}, pages 22017--22031, Miami, Florida, USA. Association for Computational Linguistics.

\bibitem[{Han et~al.(2024{\natexlab{b}})Han, Yu, Shen, Qi, Riddell, Zhou, Qiao, Zhao, Yavuz, Liu, Joty, Zhou, Xiong, Radev, Ying, and Cohan}]{han-etal-2024-p}
Simeng Han, Aaron Yu, Rui Shen, Zhenting Qi, Martin Riddell, Wenfei Zhou, Yujie Qiao, Yilun Zhao, Semih Yavuz, Ye~Liu, Shafiq Joty, Yingbo Zhou, Caiming Xiong, Dragomir Radev, Rex Ying, and Arman Cohan. 2024{\natexlab{b}}.
\newblock \href {https://doi.org/10.18653/v1/2024.findings-emnlp.966} {{P}-{FOLIO}: Evaluating and improving logical reasoning with abundant human-written reasoning chains}.
\newblock In \emph{Findings of the Association for Computational Linguistics: EMNLP 2024}, pages 16553--16565, Miami, Florida, USA. Association for Computational Linguistics.

\bibitem[{Hao et~al.(2024)Hao, Gu, Luo, Liu, Shao, Wang, Xie, Ma, Samavedhi, Gao, Wang, and Hu}]{hao2024llmreasonersnewevaluation}
Shibo Hao, Yi~Gu, Haotian Luo, Tianyang Liu, Xiyan Shao, Xinyuan Wang, Shuhua Xie, Haodi Ma, Adithya Samavedhi, Qiyue Gao, Zhen Wang, and Zhiting Hu. 2024.
\newblock \href {https://arxiv.org/abs/2404.05221} {Llm reasoners: New evaluation, library, and analysis of step-by-step reasoning with large language models}.
\newblock \emph{Preprint}, arXiv:2404.05221.

\bibitem[{He et~al.(2024{\natexlab{a}})He, Luo, Bai, Hu, Thai, Shen, Hu, Han, Huang, Zhang, Liu, Qi, Liu, and Sun}]{he-etal-2024-olympiadbench}
Chaoqun He, Renjie Luo, Yuzhuo Bai, Shengding Hu, Zhen Thai, Junhao Shen, Jinyi Hu, Xu~Han, Yujie Huang, Yuxiang Zhang, Jie Liu, Lei Qi, Zhiyuan Liu, and Maosong Sun. 2024{\natexlab{a}}.
\newblock \href {https://doi.org/10.18653/v1/2024.acl-long.211} {{O}lympiad{B}ench: A challenging benchmark for promoting {AGI} with olympiad-level bilingual multimodal scientific problems}.
\newblock In \emph{Proceedings of the 62nd Annual Meeting of the Association for Computational Linguistics (Volume 1: Long Papers)}, pages 3828--3850, Bangkok, Thailand. Association for Computational Linguistics.

\bibitem[{He et~al.(2024{\natexlab{b}})He, Shen, Zhang, Tan, and Lu}]{he-etal-2024-advancing}
Mingqian He, Yongliang Shen, Wenqi Zhang, Zeqi Tan, and Weiming Lu. 2024{\natexlab{b}}.
\newblock \href {https://doi.org/10.18653/v1/2024.emnlp-main.125} {Advancing process verification for large language models via tree-based preference learning}.
\newblock In \emph{Proceedings of the 2024 Conference on Empirical Methods in Natural Language Processing}, pages 2086--2099, Miami, Florida, USA. Association for Computational Linguistics.

\bibitem[{Hendrycks et~al.(2021)Hendrycks, Burns, Basart, Zou, Mazeika, Song, and Steinhardt}]{hendrycks2021measuringmassivemultitasklanguage}
Dan Hendrycks, Collin Burns, Steven Basart, Andy Zou, Mantas Mazeika, Dawn Song, and Jacob Steinhardt. 2021.
\newblock \href {https://arxiv.org/abs/2009.03300} {Measuring massive multitask language understanding}.
\newblock \emph{Preprint}, arXiv:2009.03300.

\bibitem[{Hewitt et~al.(2021)Hewitt, Ethayarajh, Liang, and Manning}]{hewitt-etal-2021-conditional}
John Hewitt, Kawin Ethayarajh, Percy Liang, and Christopher Manning. 2021.
\newblock \href {https://doi.org/10.18653/v1/2021.emnlp-main.122} {Conditional probing: measuring usable information beyond a baseline}.
\newblock In \emph{Proceedings of the 2021 Conference on Empirical Methods in Natural Language Processing}, pages 1626--1639, Online and Punta Cana, Dominican Republic. Association for Computational Linguistics.

\bibitem[{Holzenberger et~al.(2020)Holzenberger, Blair-Stanek, and Durme}]{holzenberger2020datasetstatutoryreasoningtax}
Nils Holzenberger, Andrew Blair-Stanek, and Benjamin~Van Durme. 2020.
\newblock \href {https://arxiv.org/abs/2005.05257} {A dataset for statutory reasoning in tax law entailment and question answering}.
\newblock \emph{Preprint}, arXiv:2005.05257.

\bibitem[{Holzenberger and Van~Durme(2021)}]{holzenberger-van-durme-2021-factoring}
Nils Holzenberger and Benjamin Van~Durme. 2021.
\newblock \href {https://doi.org/10.18653/v1/2021.acl-long.213} {Factoring statutory reasoning as language understanding challenges}.
\newblock In \emph{Proceedings of the 59th Annual Meeting of the Association for Computational Linguistics and the 11th International Joint Conference on Natural Language Processing (Volume 1: Long Papers)}, pages 2742--2758, Online. Association for Computational Linguistics.

\bibitem[{Hosseini et~al.(2024)Hosseini, Yuan, Malkin, Courville, Sordoni, and Agarwal}]{hosseini2024vstartrainingverifiersselftaught}
Arian Hosseini, Xingdi Yuan, Nikolay Malkin, Aaron Courville, Alessandro Sordoni, and Rishabh Agarwal. 2024.
\newblock \href {https://arxiv.org/abs/2402.06457} {V-star: Training verifiers for self-taught reasoners}.
\newblock \emph{Preprint}, arXiv:2402.06457.

\bibitem[{Hu et~al.(2024)Hu, Liu, Feng, Zhao, Ng, Luu, He, Koh, and Hooi}]{hu2024uncertainty}
Zhiyuan Hu, Chumin Liu, Xidong Feng, Yilun Zhao, See-Kiong Ng, Anh~Tuan Luu, Junxian He, Pang Wei~W Koh, and Bryan Hooi. 2024.
\newblock Uncertainty of thoughts: Uncertainty-aware planning enhances information seeking in llms.
\newblock \emph{Advances in Neural Information Processing Systems}, 37:24181--24215.

\bibitem[{Huang et~al.(2024{\natexlab{a}})Huang, Chen, Mishra, Zheng, Yu, Song, and Zhou}]{huang2024largelanguagemodelsselfcorrect}
Jie Huang, Xinyun Chen, Swaroop Mishra, Huaixiu~Steven Zheng, Adams~Wei Yu, Xinying Song, and Denny Zhou. 2024{\natexlab{a}}.
\newblock \href {https://arxiv.org/abs/2310.01798} {Large language models cannot self-correct reasoning yet}.
\newblock \emph{Preprint}, arXiv:2310.01798.

\bibitem[{Huang et~al.(2024{\natexlab{b}})Huang, Yu, Ma, Zhong, Feng, Wang, Chen, Peng, Feng, Qin et~al.}]{huang2024survey}
Lei Huang, Weijiang Yu, Weitao Ma, Weihong Zhong, Zhangyin Feng, Haotian Wang, Qianglong Chen, Weihua Peng, Xiaocheng Feng, Bing Qin, et~al. 2024{\natexlab{b}}.
\newblock A survey on hallucination in large language models: Principles, taxonomy, challenges, and open questions.
\newblock \emph{ACM Transactions on Information Systems}.

\bibitem[{Jacovi et~al.(2024)Jacovi, Bitton, Bohnet, Herzig, Honovich, Tseng, Collins, Aharoni, and Geva}]{jacovi-etal-2024-chain}
Alon Jacovi, Yonatan Bitton, Bernd Bohnet, Jonathan Herzig, Or~Honovich, Michael Tseng, Michael Collins, Roee Aharoni, and Mor Geva. 2024.
\newblock \href {https://doi.org/10.18653/v1/2024.acl-long.254} {A chain-of-thought is as strong as its weakest link: A benchmark for verifiers of reasoning chains}.
\newblock In \emph{Proceedings of the 62nd Annual Meeting of the Association for Computational Linguistics (Volume 1: Long Papers)}, pages 4615--4634, Bangkok, Thailand. Association for Computational Linguistics.

\bibitem[{Ji et~al.(2023)Ji, Lee, Frieske, Yu, Su, Xu, Ishii, Bang, Madotto, and Fung}]{ji2023survey}
Ziwei Ji, Nayeon Lee, Rita Frieske, Tiezheng Yu, Dan Su, Yan Xu, Etsuko Ishii, Ye~Jin Bang, Andrea Madotto, and Pascale Fung. 2023.
\newblock Survey of hallucination in natural language generation.
\newblock \emph{ACM Computing Surveys}, 55(12):1--38.

\bibitem[{Jimenez et~al.(2024)Jimenez, Yang, Wettig, Yao, Pei, Press, and Narasimhan}]{jimenez2024swebenchlanguagemodelsresolve}
Carlos~E. Jimenez, John Yang, Alexander Wettig, Shunyu Yao, Kexin Pei, Ofir Press, and Karthik Narasimhan. 2024.
\newblock \href {https://arxiv.org/abs/2310.06770} {Swe-bench: Can language models resolve real-world github issues?}
\newblock \emph{Preprint}, arXiv:2310.06770.

\bibitem[{Kang et~al.(2024)Kang, Zhao, Hsu, and Lee}]{kang2024empiricalcomplexityreasoningplanning}
Liwei Kang, Zirui Zhao, David Hsu, and Wee~Sun Lee. 2024.
\newblock \href {https://arxiv.org/abs/2404.11041} {On the empirical complexity of reasoning and planning in llms}.
\newblock \emph{Preprint}, arXiv:2404.11041.

\bibitem[{Khalifa et~al.(2025)Khalifa, Agarwal, Logeswaran, Kim, Peng, Lee, Lee, and Wang}]{khalifa2025processrewardmodelsthink}
Muhammad Khalifa, Rishabh Agarwal, Lajanugen Logeswaran, Jaekyeom Kim, Hao Peng, Moontae Lee, Honglak Lee, and Lu~Wang. 2025.
\newblock \href {https://arxiv.org/abs/2504.16828} {Process reward models that think}.
\newblock \emph{Preprint}, arXiv:2504.16828.

\bibitem[{Kim et~al.(2024{\natexlab{a}})Kim, Shin, Choi, Jang, Longpre, Lee, Yun, Shin, Kim, Thorne, and Seo}]{DBLP:conf/iclr/KimS0JLLYSKTS24}
Seungone Kim, Jamin Shin, Yejin Choi, Joel Jang, Shayne Longpre, Hwaran Lee, Sangdoo Yun, Seongjin Shin, Sungdong Kim, James Thorne, and Minjoon Seo. 2024{\natexlab{a}}.
\newblock \href {https://openreview.net/forum?id=8euJaTveKw} {Prometheus: Inducing fine-grained evaluation capability in language models}.
\newblock In \emph{The Twelfth International Conference on Learning Representations, {ICLR} 2024, Vienna, Austria, May 7-11, 2024}. OpenReview.net.

\bibitem[{Kim et~al.(2025{\natexlab{a}})Kim, Suk, Cho, Longpre, Kim, Yoon, Son, Cho, Shafayat, Baek, Park, Hwang, Jo, Cho, Shin, Lee, Oh, Lee, Ho, Joo, Ko, Lee, Chae, Shin, Jang, Ye, Lin, Welleck, Neubig, Lee, Lee, and Seo}]{kim2024biggenbenchprincipledbenchmark}
Seungone Kim, Juyoung Suk, Ji~Yong Cho, Shayne Longpre, Chaeeun Kim, Dongkeun Yoon, Guijin Son, Yejin Cho, Sheikh Shafayat, Jinheon Baek, Sue~Hyun Park, Hyeonbin Hwang, Jinkyung Jo, Hyowon Cho, Haebin Shin, Seongyun Lee, Hanseok Oh, Noah Lee, Namgyu Ho, Se~June Joo, Miyoung Ko, Yoonjoo Lee, Hyungjoo Chae, Jamin Shin, Joel Jang, Seonghyeon Ye, Bill~Yuchen Lin, Sean Welleck, Graham Neubig, Moontae Lee, Kyungjae Lee, and Minjoon Seo. 2025{\natexlab{a}}.
\newblock \href {https://aclanthology.org/2025.naacl-long.303/} {The {B}i{GG}en bench: A principled benchmark for fine-grained evaluation of language models with language models}.
\newblock In \emph{Proceedings of the 2025 Conference of the Nations of the Americas Chapter of the Association for Computational Linguistics: Human Language Technologies (Volume 1: Long Papers)}, pages 5877--5919, Albuquerque, New Mexico. Association for Computational Linguistics.

\bibitem[{Kim et~al.(2025{\natexlab{b}})Kim, Wu, Lee, Yue, Lee, Moon, Gashteovski, Lawrence, Hockenmaier, Neubig, and Welleck}]{kim2025scalingevaluationtimecomputereasoning}
Seungone Kim, Ian Wu, Jinu Lee, Xiang Yue, Seongyun Lee, Mingyeong Moon, Kiril Gashteovski, Carolin Lawrence, Julia Hockenmaier, Graham Neubig, and Sean Welleck. 2025{\natexlab{b}}.
\newblock \href {https://arxiv.org/abs/2503.19877} {Scaling evaluation-time compute with reasoning models as process evaluators}.
\newblock \emph{Preprint}, arXiv:2503.19877.

\bibitem[{Kim et~al.(2024{\natexlab{b}})Kim, Choi, Choi, Choi, Park, and Hwang}]{kimyeeun-etal-2024-developing}
Yeeun Kim, Youngrok Choi, Eunkyung Choi, JinHwan Choi, Hai~Jin Park, and Wonseok Hwang. 2024{\natexlab{b}}.
\newblock \href {https://doi.org/10.18653/v1/2024.findings-emnlp.319} {Developing a pragmatic benchmark for assessing {K}orean legal language understanding in large language models}.
\newblock In \emph{Findings of the Association for Computational Linguistics: EMNLP 2024}, pages 5573--5595, Miami, Florida, USA. Association for Computational Linguistics.

\bibitem[{Kojima et~al.(2022)Kojima, Gu, Reid, Matsuo, and Iwasawa}]{NEURIPS2022_8bb0d291}
Takeshi Kojima, Shixiang~(Shane) Gu, Machel Reid, Yutaka Matsuo, and Yusuke Iwasawa. 2022.
\newblock \href {https://proceedings.neurips.cc/paper_files/paper/2022/file/8bb0d291acd4acf06ef112099c16f326-Paper-Conference.pdf} {Large language models are zero-shot reasoners}.
\newblock In \emph{Advances in Neural Information Processing Systems}, volume~35, pages 22199--22213. Curran Associates, Inc.

\bibitem[{Koncel-Kedziorski et~al.(2016)Koncel-Kedziorski, Roy, Amini, Kushman, and Hajishirzi}]{koncel-kedziorski-etal-2016-mawps}
Rik Koncel-Kedziorski, Subhro Roy, Aida Amini, Nate Kushman, and Hannaneh Hajishirzi. 2016.
\newblock \href {https://doi.org/10.18653/v1/N16-1136} {{MAWPS}: A math word problem repository}.
\newblock In \emph{Proceedings of the 2016 Conference of the North {A}merican Chapter of the Association for Computational Linguistics: Human Language Technologies}, pages 1152--1157, San Diego, California. Association for Computational Linguistics.

\bibitem[{Kossen et~al.(2024)Kossen, Han, Razzak, Schut, Malik, and Gal}]{kossen2024semanticentropyprobesrobust}
Jannik Kossen, Jiatong Han, Muhammed Razzak, Lisa Schut, Shreshth Malik, and Yarin Gal. 2024.
\newblock \href {https://arxiv.org/abs/2406.15927} {Semantic entropy probes: Robust and cheap hallucination detection in llms}.
\newblock \emph{Preprint}, arXiv:2406.15927.

\bibitem[{Kurtic et~al.(2024)Kurtic, Moeini, and Alistarh}]{kurtic-etal-2024-mathador}
Eldar Kurtic, Amir Moeini, and Dan Alistarh. 2024.
\newblock \href {https://doi.org/10.18653/v1/2024.emnlp-main.946} {Mathador-{LM}: A dynamic benchmark for mathematical reasoning on large language models}.
\newblock In \emph{Proceedings of the 2024 Conference on Empirical Methods in Natural Language Processing}, pages 17020--17027, Miami, Florida, USA. Association for Computational Linguistics.

\bibitem[{Kwiatkowski et~al.(2019)Kwiatkowski, Palomaki, Redfield, Collins, Parikh, Alberti, Epstein, Polosukhin, Devlin, Lee, Toutanova, Jones, Kelcey, Chang, Dai, Uszkoreit, Le, and Petrov}]{kwiatkowski-etal-2019-natural}
Tom Kwiatkowski, Jennimaria Palomaki, Olivia Redfield, Michael Collins, Ankur Parikh, Chris Alberti, Danielle Epstein, Illia Polosukhin, Jacob Devlin, Kenton Lee, Kristina Toutanova, Llion Jones, Matthew Kelcey, Ming-Wei Chang, Andrew~M. Dai, Jakob Uszkoreit, Quoc Le, and Slav Petrov. 2019.
\newblock \href {https://doi.org/10.1162/tacl_a_00276} {Natural questions: A benchmark for question answering research}.
\newblock \emph{Transactions of the Association for Computational Linguistics}, 7:452--466.

\bibitem[{Lai et~al.(2024)Lai, Tian, Chen, Yang, Peng, and Jia}]{lai2024stepdpostepwisepreferenceoptimization}
Xin Lai, Zhuotao Tian, Yukang Chen, Senqiao Yang, Xiangru Peng, and Jiaya Jia. 2024.
\newblock \href {https://arxiv.org/abs/2406.18629} {Step-dpo: Step-wise preference optimization for long-chain reasoning of llms}.
\newblock \emph{Preprint}, arXiv:2406.18629.

\bibitem[{Lambert et~al.(2025)Lambert, Pyatkin, Morrison, Miranda, Lin, Chandu, Dziri, Kumar, Zick, Choi, Smith, and Hajishirzi}]{lambert-etal-2025-rewardbench}
Nathan Lambert, Valentina Pyatkin, Jacob Morrison, Lester James~Validad Miranda, Bill~Yuchen Lin, Khyathi Chandu, Nouha Dziri, Sachin Kumar, Tom Zick, Yejin Choi, Noah~A. Smith, and Hannaneh Hajishirzi. 2025.
\newblock \href {https://aclanthology.org/2025.findings-naacl.96/} {{R}eward{B}ench: Evaluating reward models for language modeling}.
\newblock In \emph{Findings of the Association for Computational Linguistics: NAACL 2025}, pages 1755--1797, Albuquerque, New Mexico. Association for Computational Linguistics.

\bibitem[{Lanham et~al.(2023)Lanham, Chen, Radhakrishnan, Steiner, Denison, Hernandez, Li, Durmus, Hubinger, Kernion, Lukošiūtė, Nguyen, Cheng, Joseph, Schiefer, Rausch, Larson, McCandlish, Kundu, Kadavath, Yang, Henighan, Maxwell, Telleen-Lawton, Hume, Hatfield-Dodds, Kaplan, Brauner, Bowman, and Perez}]{lanham2023measuringfaithfulnesschainofthoughtreasoning}
Tamera Lanham, Anna Chen, Ansh Radhakrishnan, Benoit Steiner, Carson Denison, Danny Hernandez, Dustin Li, Esin Durmus, Evan Hubinger, Jackson Kernion, Kamilė Lukošiūtė, Karina Nguyen, Newton Cheng, Nicholas Joseph, Nicholas Schiefer, Oliver Rausch, Robin Larson, Sam McCandlish, Sandipan Kundu, Saurav Kadavath, Shannon Yang, Thomas Henighan, Timothy Maxwell, Timothy Telleen-Lawton, Tristan Hume, Zac Hatfield-Dodds, Jared Kaplan, Jan Brauner, Samuel~R. Bowman, and Ethan Perez. 2023.
\newblock \href {https://arxiv.org/abs/2307.13702} {Measuring faithfulness in chain-of-thought reasoning}.
\newblock \emph{Preprint}, arXiv:2307.13702.

\bibitem[{Lee and Hwang(2025)}]{lee2025symbasymbolicbackwardchaining}
Jinu Lee and Wonseok Hwang. 2025.
\newblock \href {https://arxiv.org/abs/2402.12806} {Symba: Symbolic backward chaining for structured natural language reasoning}.
\newblock \emph{Preprint}, arXiv:2402.12806.

\bibitem[{Lee et~al.(2025{\natexlab{a}})Lee, Liu, Ma, Han, Wang, Ji, and Hockenmaier}]{lee2025entailmentpreservingfirstorderlogicrepresentations}
Jinu Lee, Qi~Liu, Runzhi Ma, Vincent Han, Ziqi Wang, Heng Ji, and Julia Hockenmaier. 2025{\natexlab{a}}.
\newblock \href {https://arxiv.org/abs/2502.16757} {Entailment-preserving first-order logic representations in natural language entailment}.
\newblock \emph{Preprint}, arXiv:2502.16757.

\bibitem[{Lee et~al.(2025{\natexlab{b}})Lee, Mukherjee, Hakkani-Tur, and Hockenmaier}]{lee2025reasoningflowsemanticstructurecomplex}
Jinu Lee, Sagnik Mukherjee, Dilek Hakkani-Tur, and Julia Hockenmaier. 2025{\natexlab{b}}.
\newblock \href {https://arxiv.org/abs/2506.02532} {Reasoningflow: Semantic structure of complex reasoning traces}.
\newblock \emph{Preprint}, arXiv:2506.02532.

\bibitem[{Lewis et~al.(2020)Lewis, Perez, Piktus, Petroni, Karpukhin, Goyal, K\"{u}ttler, Lewis, Yih, Rockt\"{a}schel, Riedel, and Kiela}]{NEURIPS2020_6b493230}
Patrick Lewis, Ethan Perez, Aleksandra Piktus, Fabio Petroni, Vladimir Karpukhin, Naman Goyal, Heinrich K\"{u}ttler, Mike Lewis, Wen-tau Yih, Tim Rockt\"{a}schel, Sebastian Riedel, and Douwe Kiela. 2020.
\newblock \href {https://proceedings.neurips.cc/paper_files/paper/2020/file/6b493230205f780e1bc26945df7481e5-Paper.pdf} {Retrieval-augmented generation for knowledge-intensive nlp tasks}.
\newblock In \emph{Advances in Neural Information Processing Systems}, volume~33, pages 9459--9474. Curran Associates, Inc.

\bibitem[{Li et~al.(2023{\natexlab{a}})Li, Cheng, Zhao, Nie, and Wen}]{li2023haluevallargescalehallucinationevaluation}
Junyi Li, Xiaoxue Cheng, Wayne~Xin Zhao, Jian-Yun Nie, and Ji-Rong Wen. 2023{\natexlab{a}}.
\newblock \href {https://arxiv.org/abs/2305.11747} {Halueval: A large-scale hallucination evaluation benchmark for large language models}.
\newblock \emph{Preprint}, arXiv:2305.11747.

\bibitem[{Li et~al.(2024{\natexlab{a}})Li, Wang, Tran, Xia, and Du}]{NEURIPS2024_e560a0b2}
Ruosen Li, Zimu Wang, Son Tran, Lei Xia, and Xinya Du. 2024{\natexlab{a}}.
\newblock \href {https://proceedings.neurips.cc/paper_files/paper/2024/file/e560a0b22e4432003d0dba63ff8dc457-Paper-Datasets_and_Benchmarks_Track.pdf} {Meqa: A benchmark for multi-hop event-centric question answering with explanations}.
\newblock In \emph{Advances in Neural Information Processing Systems}, volume~37, pages 126835--126862. Curran Associates, Inc.

\bibitem[{Li et~al.(2024{\natexlab{b}})Li, Li, Shi, Xu, Du, Tan, and Huang}]{li-etal-2024-alphafin}
Xiang Li, Zhenyu Li, Chen Shi, Yong Xu, Qing Du, Mingkui Tan, and Jun Huang. 2024{\natexlab{b}}.
\newblock \href {https://aclanthology.org/2024.lrec-main.69/} {{A}lpha{F}in: Benchmarking financial analysis with retrieval-augmented stock-chain framework}.
\newblock In \emph{Proceedings of the 2024 Joint International Conference on Computational Linguistics, Language Resources and Evaluation (LREC-COLING 2024)}, pages 773--783, Torino, Italia. ELRA and ICCL.

\bibitem[{Li et~al.(2023{\natexlab{b}})Li, Lin, Zhang, Fu, Chen, Lou, and Chen}]{li-etal-2023-making}
Yifei Li, Zeqi Lin, Shizhuo Zhang, Qiang Fu, Bei Chen, Jian-Guang Lou, and Weizhu Chen. 2023{\natexlab{b}}.
\newblock \href {https://doi.org/10.18653/v1/2023.acl-long.291} {Making language models better reasoners with step-aware verifier}.
\newblock In \emph{Proceedings of the 61st Annual Meeting of the Association for Computational Linguistics (Volume 1: Long Papers)}, pages 5315--5333, Toronto, Canada. Association for Computational Linguistics.

\bibitem[{Li et~al.(2022)Li, Choi, Chung, Kushman, Schrittwieser, Leblond, Eccles, Keeling, Gimeno, Dal~Lago et~al.}]{li2022competition}
Yujia Li, David Choi, Junyoung Chung, Nate Kushman, Julian Schrittwieser, R{\'e}mi Leblond, Tom Eccles, James Keeling, Felix Gimeno, Agustin Dal~Lago, et~al. 2022.
\newblock Competition-level code generation with alphacode.
\newblock \emph{Science}, 378(6624):1092--1097.

\bibitem[{Lightman et~al.(2024)Lightman, Kosaraju, Burda, Edwards, Baker, Lee, Leike, Schulman, Sutskever, and Cobbe}]{DBLP:conf/iclr/LightmanKBEBLLS24}
Hunter Lightman, Vineet Kosaraju, Yuri Burda, Harrison Edwards, Bowen Baker, Teddy Lee, Jan Leike, John Schulman, Ilya Sutskever, and Karl Cobbe. 2024.
\newblock \href {https://openreview.net/forum?id=v8L0pN6EOi} {Let's verify step by step}.
\newblock In \emph{The Twelfth International Conference on Learning Representations, {ICLR} 2024, Vienna, Austria, May 7-11, 2024}. OpenReview.net.

\bibitem[{Lin et~al.(2024)Lin, Gou, Liang, Luo, Liu, and Yang}]{lin2024criticbenchbenchmarkingllmscritiquecorrect}
Zicheng Lin, Zhibin Gou, Tian Liang, Ruilin Luo, Haowei Liu, and Yujiu Yang. 2024.
\newblock \href {https://arxiv.org/abs/2402.14809} {Criticbench: Benchmarking llms for critique-correct reasoning}.
\newblock \emph{Preprint}, arXiv:2402.14809.

\bibitem[{Ling et~al.(2023)Ling, Fang, Li, Huang, Lee, Memisevic, and Su}]{NEURIPS2023_72393bd4}
Zhan Ling, Yunhao Fang, Xuanlin Li, Zhiao Huang, Mingu Lee, Roland Memisevic, and Hao Su. 2023.
\newblock \href {https://proceedings.neurips.cc/paper_files/paper/2023/file/72393bd47a35f5b3bee4c609e7bba733-Paper-Conference.pdf} {Deductive verification of chain-of-thought reasoning}.
\newblock In \emph{Advances in Neural Information Processing Systems}, volume~36, pages 36407--36433. Curran Associates, Inc.

\bibitem[{Liu et~al.(2019)Liu, Ott, Goyal, Du, Joshi, Chen, Levy, Lewis, Zettlemoyer, and Stoyanov}]{liu2019robertarobustlyoptimizedbert}
Yinhan Liu, Myle Ott, Naman Goyal, Jingfei Du, Mandar Joshi, Danqi Chen, Omer Levy, Mike Lewis, Luke Zettlemoyer, and Veselin Stoyanov. 2019.
\newblock \href {https://arxiv.org/abs/1907.11692} {Roberta: A robustly optimized bert pretraining approach}.
\newblock \emph{Preprint}, arXiv:1907.11692.

\bibitem[{Lu et~al.(2025)Lu, Tan, Xu, Yao, Qu, Chu, Xu, and Qi}]{lu2025scp116khighqualityproblemsolutiondataset}
Dakuan Lu, Xiaoyu Tan, Rui Xu, Tianchu Yao, Chao Qu, Wei Chu, Yinghui Xu, and Yuan Qi. 2025.
\newblock \href {https://arxiv.org/abs/2501.15587} {Scp-116k: A high-quality problem-solution dataset and a generalized pipeline for automated extraction in the higher education science domain}.
\newblock \emph{Preprint}, arXiv:2501.15587.

\bibitem[{Lu et~al.(2024)Lu, Zhou, Wang, Ren, Shi, Pan, Zhan, and Li}]{lu2024stepcontrolleddpoleveragingstepwise}
Zimu Lu, Aojun Zhou, Ke~Wang, Houxing Ren, Weikang Shi, Junting Pan, Mingjie Zhan, and Hongsheng Li. 2024.
\newblock \href {https://arxiv.org/abs/2407.00782} {Step-controlled dpo: Leveraging stepwise error for enhanced mathematical reasoning}.
\newblock \emph{Preprint}, arXiv:2407.00782.

\bibitem[{Luo et~al.(2024{\natexlab{a}})Luo, Li, Wu, Jenkin, Liu, and Dudek}]{luo2024hallucination}
Junliang Luo, Tianyu Li, Di~Wu, Michael Jenkin, Steve Liu, and Gregory Dudek. 2024{\natexlab{a}}.
\newblock Hallucination detection and hallucination mitigation: An investigation.
\newblock \emph{arXiv preprint arXiv:2401.08358}.

\bibitem[{Luo et~al.(2024{\natexlab{b}})Luo, Liu, Liu, Phatale, Guo, Lara, Li, Shu, Zhu, Meng, Sun, and Rastogi}]{luo2024improvemathematicalreasoninglanguage}
Liangchen Luo, Yinxiao Liu, Rosanne Liu, Samrat Phatale, Meiqi Guo, Harsh Lara, Yunxuan Li, Lei Shu, Yun Zhu, Lei Meng, Jiao Sun, and Abhinav Rastogi. 2024{\natexlab{b}}.
\newblock \href {https://arxiv.org/abs/2406.06592} {Improve mathematical reasoning in language models by automated process supervision}.
\newblock \emph{Preprint}, arXiv:2406.06592.

\bibitem[{Lyu et~al.(2023)Lyu, Havaldar, Stein, Zhang, Rao, Wong, Apidianaki, and Callison-Burch}]{lyu-etal-2023-faithful}
Qing Lyu, Shreya Havaldar, Adam Stein, Li~Zhang, Delip Rao, Eric Wong, Marianna Apidianaki, and Chris Callison-Burch. 2023.
\newblock \href {https://doi.org/10.18653/v1/2023.ijcnlp-main.20} {Faithful chain-of-thought reasoning}.
\newblock In \emph{Proceedings of the 13th International Joint Conference on Natural Language Processing and the 3rd Conference of the Asia-Pacific Chapter of the Association for Computational Linguistics (Volume 1: Long Papers)}, pages 305--329, Nusa Dua, Bali. Association for Computational Linguistics.

\bibitem[{Mahan et~al.(2024)Mahan, Phung, Rafailov, Blagden, Lile, Castricato, Fränken, Finn, and Albalak}]{mahan2024generativerewardmodels}
Dakota Mahan, Duy~Van Phung, Rafael Rafailov, Chase Blagden, Nathan Lile, Louis Castricato, Jan-Philipp Fränken, Chelsea Finn, and Alon Albalak. 2024.
\newblock \href {https://arxiv.org/abs/2410.12832} {Generative reward models}.
\newblock \emph{Preprint}, arXiv:2410.12832.

\bibitem[{Maynez et~al.(2020)Maynez, Narayan, Bohnet, and McDonald}]{maynez2020faithfulnessfactualityabstractivesummarization}
Joshua Maynez, Shashi Narayan, Bernd Bohnet, and Ryan McDonald. 2020.
\newblock \href {https://arxiv.org/abs/2005.00661} {On faithfulness and factuality in abstractive summarization}.
\newblock \emph{Preprint}, arXiv:2005.00661.

\bibitem[{Miao et~al.(2020)Miao, Liang, and Su}]{miao-etal-2020-diverse}
Shen-yun Miao, Chao-Chun Liang, and Keh-Yih Su. 2020.
\newblock \href {https://doi.org/10.18653/v1/2020.acl-main.92} {A diverse corpus for evaluating and developing {E}nglish math word problem solvers}.
\newblock In \emph{Proceedings of the 58th Annual Meeting of the Association for Computational Linguistics}, pages 975--984, Online. Association for Computational Linguistics.

\bibitem[{Mihaylov et~al.(2018)Mihaylov, Clark, Khot, and Sabharwal}]{mihaylov2018suitarmorconductelectricity}
Todor Mihaylov, Peter Clark, Tushar Khot, and Ashish Sabharwal. 2018.
\newblock \href {https://arxiv.org/abs/1809.02789} {Can a suit of armor conduct electricity? a new dataset for open book question answering}.
\newblock \emph{Preprint}, arXiv:1809.02789.

\bibitem[{Mirzadeh et~al.(2024)Mirzadeh, Alizadeh, Shahrokhi, Tuzel, Bengio, and Farajtabar}]{mirzadeh2024gsmsymbolicunderstandinglimitationsmathematical}
Iman Mirzadeh, Keivan Alizadeh, Hooman Shahrokhi, Oncel Tuzel, Samy Bengio, and Mehrdad Farajtabar. 2024.
\newblock \href {https://arxiv.org/abs/2410.05229} {Gsm-symbolic: Understanding the limitations of mathematical reasoning in large language models}.
\newblock \emph{Preprint}, arXiv:2410.05229.

\bibitem[{Muennighoff et~al.(2025)Muennighoff, Yang, Shi, Li, Fei-Fei, Hajishirzi, Zettlemoyer, Liang, Candès, and Hashimoto}]{muennighoff2025s1simpletesttimescaling}
Niklas Muennighoff, Zitong Yang, Weijia Shi, Xiang~Lisa Li, Li~Fei-Fei, Hannaneh Hajishirzi, Luke Zettlemoyer, Percy Liang, Emmanuel Candès, and Tatsunori Hashimoto. 2025.
\newblock \href {https://arxiv.org/abs/2501.19393} {s1: Simple test-time scaling}.
\newblock \emph{Preprint}, arXiv:2501.19393.

\bibitem[{Mukherjee et~al.(2025)Mukherjee, Chinta, Kim, Sharma, and Hakkani-Tür}]{mukherjee2025premiseaugmentedreasoningchainsimprove}
Sagnik Mukherjee, Abhinav Chinta, Takyoung Kim, Tarun~Anoop Sharma, and Dilek Hakkani-Tür. 2025.
\newblock \href {https://arxiv.org/abs/2502.02362} {Premise-augmented reasoning chains improve error identification in math reasoning with llms}.
\newblock \emph{Preprint}, arXiv:2502.02362.

\bibitem[{Nguyen et~al.(2024)Nguyen, Luo, Shiri, Phung, Li, Vu, and Haffari}]{nguyen-etal-2024-direct}
Thi Nguyen, Linhao Luo, Fatemeh Shiri, Dinh Phung, Yuan-Fang Li, Thuy-Trang Vu, and Gholamreza Haffari. 2024.
\newblock \href {https://doi.org/10.18653/v1/2024.findings-acl.168} {Direct evaluation of chain-of-thought in multi-hop reasoning with knowledge graphs}.
\newblock In \emph{Findings of the Association for Computational Linguistics: ACL 2024}, pages 2862--2883, Bangkok, Thailand. Association for Computational Linguistics.

\bibitem[{Niu et~al.(2024)Niu, Wu, Zhu, Xu, Shum, Zhong, Song, and Zhang}]{niu2024ragtruthhallucinationcorpusdeveloping}
Cheng Niu, Yuanhao Wu, Juno Zhu, Siliang Xu, Kashun Shum, Randy Zhong, Juntong Song, and Tong Zhang. 2024.
\newblock \href {https://arxiv.org/abs/2401.00396} {Ragtruth: A hallucination corpus for developing trustworthy retrieval-augmented language models}.
\newblock \emph{Preprint}, arXiv:2401.00396.

\bibitem[{Olausson et~al.(2023)Olausson, Gu, Lipkin, Zhang, Solar-Lezama, Tenenbaum, and Levy}]{olausson-etal-2023-linc}
Theo Olausson, Alex Gu, Ben Lipkin, Cedegao Zhang, Armando Solar-Lezama, Joshua Tenenbaum, and Roger Levy. 2023.
\newblock \href {https://doi.org/10.18653/v1/2023.emnlp-main.313} {{LINC}: A neurosymbolic approach for logical reasoning by combining language models with first-order logic provers}.
\newblock In \emph{Proceedings of the 2023 Conference on Empirical Methods in Natural Language Processing}, pages 5153--5176, Singapore. Association for Computational Linguistics.

\bibitem[{OpenAI(2024{\natexlab{a}})}]{openai2024gpt4technicalreport}
OpenAI. 2024{\natexlab{a}}.
\newblock \href {https://arxiv.org/abs/2303.08774} {Gpt-4 technical report}.
\newblock \emph{Preprint}, arXiv:2303.08774.

\bibitem[{OpenAI(2024{\natexlab{b}})}]{openai2024openaio1card}
OpenAI. 2024{\natexlab{b}}.
\newblock \href {https://arxiv.org/abs/2412.16720} {Openai o1 system card}.
\newblock \emph{Preprint}, arXiv:2412.16720.

\bibitem[{Ott et~al.(2023)Ott, Hebenstreit, Liévin, Hother, Moradi, Mayrhauser, Praas, Winther, and Samwald}]{Ott_2023}
Simon Ott, Konstantin Hebenstreit, Valentin Liévin, Christoffer~Egeberg Hother, Milad Moradi, Maximilian Mayrhauser, Robert Praas, Ole Winther, and Matthias Samwald. 2023.
\newblock \href {https://doi.org/10.1038/s41597-023-02433-3} {Thoughtsource: A central hub for large language model reasoning data}.
\newblock \emph{Scientific Data}, 10(1).

\bibitem[{Pan et~al.(2023{\natexlab{a}})Pan, Albalak, Wang, and Wang}]{pan-etal-2023-logic}
Liangming Pan, Alon Albalak, Xinyi Wang, and William Wang. 2023{\natexlab{a}}.
\newblock \href {https://doi.org/10.18653/v1/2023.findings-emnlp.248} {Logic-{LM}: Empowering large language models with symbolic solvers for faithful logical reasoning}.
\newblock In \emph{Findings of the Association for Computational Linguistics: EMNLP 2023}, pages 3806--3824, Singapore. Association for Computational Linguistics.

\bibitem[{Pan et~al.(2023{\natexlab{b}})Pan, Lialin, Muckatira, and Rumshisky}]{pan2023letsreinforcestepstep}
Sarah Pan, Vladislav Lialin, Sherin Muckatira, and Anna Rumshisky. 2023{\natexlab{b}}.
\newblock \href {https://arxiv.org/abs/2311.05821} {Let's reinforce step by step}.
\newblock \emph{Preprint}, arXiv:2311.05821.

\bibitem[{Pang et~al.(2024)Pang, Yuan, Cho, He, Sukhbaatar, and Weston}]{pang2024iterativereasoningpreferenceoptimization}
Richard~Yuanzhe Pang, Weizhe Yuan, Kyunghyun Cho, He~He, Sainbayar Sukhbaatar, and Jason Weston. 2024.
\newblock \href {https://arxiv.org/abs/2404.19733} {Iterative reasoning preference optimization}.
\newblock \emph{Preprint}, arXiv:2404.19733.

\bibitem[{Paul et~al.(2024)Paul, West, Bosselut, and Faltings}]{paul-etal-2024-making}
Debjit Paul, Robert West, Antoine Bosselut, and Boi Faltings. 2024.
\newblock \href {https://doi.org/10.18653/v1/2024.findings-emnlp.882} {Making reasoning matter: Measuring and improving faithfulness of chain-of-thought reasoning}.
\newblock In \emph{Findings of the Association for Computational Linguistics: EMNLP 2024}, pages 15012--15032, Miami, Florida, USA. Association for Computational Linguistics.

\bibitem[{Petrov et~al.(2025)Petrov, Dekoninck, Baltadzhiev, Drencheva, Minchev, Balunović, Jovanović, and Vechev}]{petrov2025proofbluffevaluatingllms}
Ivo Petrov, Jasper Dekoninck, Lyuben Baltadzhiev, Maria Drencheva, Kristian Minchev, Mislav Balunović, Nikola Jovanović, and Martin Vechev. 2025.
\newblock \href {https://arxiv.org/abs/2503.21934} {Proof or bluff? evaluating llms on 2025 usa math olympiad}.
\newblock \emph{Preprint}, arXiv:2503.21934.

\bibitem[{Prasad et~al.(2023)Prasad, Saha, Zhou, and Bansal}]{prasad-etal-2023-receval}
Archiki Prasad, Swarnadeep Saha, Xiang Zhou, and Mohit Bansal. 2023.
\newblock \href {https://doi.org/10.18653/v1/2023.emnlp-main.622} {{R}e{CE}val: Evaluating reasoning chains via correctness and informativeness}.
\newblock In \emph{Proceedings of the 2023 Conference on Empirical Methods in Natural Language Processing}, pages 10066--10086, Singapore. Association for Computational Linguistics.

\bibitem[{Qiu et~al.(2024)Qiu, Ou, Wu, Li, Liu, and King}]{qiu2024entropybaseddecodingretrievalaugmentedlarge}
Zexuan Qiu, Zijing Ou, Bin Wu, Jingjing Li, Aiwei Liu, and Irwin King. 2024.
\newblock \href {https://arxiv.org/abs/2406.17519} {Entropy-based decoding for retrieval-augmented large language models}.
\newblock \emph{Preprint}, arXiv:2406.17519.

\bibitem[{Qwen-Team(2024)}]{qwenlmQwQReflect}
Qwen-Team. 2024.
\newblock {Q}w{Q}: {R}eflect {D}eeply on the {B}oundaries of the {U}nknown --- qwenlm.github.io.
\newblock \url{https://qwenlm.github.io/blog/qwq-32b-preview/}.
\newblock [Accessed 13-02-2025].

\bibitem[{Rafailov et~al.(2023)Rafailov, Sharma, Mitchell, Manning, Ermon, and Finn}]{NEURIPS2023_a85b405e}
Rafael Rafailov, Archit Sharma, Eric Mitchell, Christopher~D Manning, Stefano Ermon, and Chelsea Finn. 2023.
\newblock \href {https://proceedings.neurips.cc/paper_files/paper/2023/file/a85b405ed65c6477a4fe8302b5e06ce7-Paper-Conference.pdf} {Direct preference optimization: Your language model is secretly a reward model}.
\newblock In \emph{Advances in Neural Information Processing Systems}, volume~36, pages 53728--53741. Curran Associates, Inc.

\bibitem[{Rein et~al.(2024)Rein, Hou, Stickland, Petty, Pang, Dirani, Michael, and Bowman}]{rein2024gpqa}
David Rein, Betty~Li Hou, Asa~Cooper Stickland, Jackson Petty, Richard~Yuanzhe Pang, Julien Dirani, Julian Michael, and Samuel~R. Bowman. 2024.
\newblock \href {https://openreview.net/forum?id=Ti67584b98} {{GPQA}: A graduate-level google-proof q\&a benchmark}.
\newblock In \emph{First Conference on Language Modeling}.

\bibitem[{Saparov and He(2023)}]{PrOntoQA}
Abulhair Saparov and He~He. 2023.
\newblock \href {https://openreview.net/forum?id=qFVVBzXxR2V} {Language models are greedy reasoners: A systematic formal analysis of chain-of-thought}.
\newblock In \emph{The Eleventh International Conference on Learning Representations}.

\bibitem[{Savage et~al.(2024)Savage, Nayak, Gallo, Rangan, and Chen}]{savage2024diagnostic}
Thomas Savage, Ashwin Nayak, Robert Gallo, Ekanath Rangan, and Jonathan~H Chen. 2024.
\newblock Diagnostic reasoning prompts reveal the potential for large language model interpretability in medicine.
\newblock \emph{NPJ Digital Medicine}, 7(1):20.

\bibitem[{Schnitzler et~al.(2024)Schnitzler, Ho, Huang, Boudin, Sugawara, and Aizawa}]{schnitzler2024morehopqa}
Julian Schnitzler, Xanh Ho, Jiahao Huang, Florian Boudin, Saku Sugawara, and Akiko Aizawa. 2024.
\newblock \href {https://arxiv.org/abs/2406.13397} {Morehopqa: More than multi-hop reasoning}.
\newblock \emph{Preprint}, arXiv:2406.13397.

\bibitem[{Setlur et~al.(2024)Setlur, Nagpal, Fisch, Geng, Eisenstein, Agarwal, Agarwal, Berant, and Kumar}]{setlur2024rewardingprogressscalingautomated}
Amrith Setlur, Chirag Nagpal, Adam Fisch, Xinyang Geng, Jacob Eisenstein, Rishabh Agarwal, Alekh Agarwal, Jonathan Berant, and Aviral Kumar. 2024.
\newblock \href {https://arxiv.org/abs/2410.08146} {Rewarding progress: Scaling automated process verifiers for llm reasoning}.
\newblock \emph{Preprint}, arXiv:2410.08146.

\bibitem[{Shao et~al.(2024)Shao, Wang, Zhu, Xu, Song, Bi, Zhang, Zhang, Li, Wu, and Guo}]{shao2024deepseekmathpushinglimitsmathematical}
Zhihong Shao, Peiyi Wang, Qihao Zhu, Runxin Xu, Junxiao Song, Xiao Bi, Haowei Zhang, Mingchuan Zhang, Y.~K. Li, Y.~Wu, and Daya Guo. 2024.
\newblock \href {https://arxiv.org/abs/2402.03300} {Deepseekmath: Pushing the limits of mathematical reasoning in open language models}.
\newblock \emph{Preprint}, arXiv:2402.03300.

\bibitem[{She et~al.(2025)She, Liu, Liu, Chen, Huang, and Huang}]{she2025rprmreasoningdrivenprocessreward}
Shuaijie She, Junxiao Liu, Yifeng Liu, Jiajun Chen, Xin Huang, and Shujian Huang. 2025.
\newblock \href {https://arxiv.org/abs/2503.21295} {R-prm: Reasoning-driven process reward modeling}.
\newblock \emph{Preprint}, arXiv:2503.21295.

\bibitem[{Singhi et~al.(2025)Singhi, Bansal, Hosseini, Grover, Chang, Rohrbach, and Rohrbach}]{singhi2025solveverifycomputeoptimalproblem}
Nishad Singhi, Hritik Bansal, Arian Hosseini, Aditya Grover, Kai-Wei Chang, Marcus Rohrbach, and Anna Rohrbach. 2025.
\newblock \href {https://arxiv.org/abs/2504.01005} {When to solve, when to verify: Compute-optimal problem solving and generative verification for llm reasoning}.
\newblock \emph{Preprint}, arXiv:2504.01005.

\bibitem[{Sinha et~al.(2019)Sinha, Sodhani, Dong, Pineau, and Hamilton}]{sinha-etal-2019-clutrr}
Koustuv Sinha, Shagun Sodhani, Jin Dong, Joelle Pineau, and William~L. Hamilton. 2019.
\newblock \href {https://doi.org/10.18653/v1/D19-1458} {{CLUTRR}: A diagnostic benchmark for inductive reasoning from text}.
\newblock In \emph{Proceedings of the 2019 Conference on Empirical Methods in Natural Language Processing and the 9th International Joint Conference on Natural Language Processing (EMNLP-IJCNLP)}, pages 4506--4515, Hong Kong, China. Association for Computational Linguistics.

\bibitem[{Snell et~al.(2024)Snell, Lee, Xu, and Kumar}]{snell2024scalingllmtesttimecompute}
Charlie Snell, Jaehoon Lee, Kelvin Xu, and Aviral Kumar. 2024.
\newblock \href {https://arxiv.org/abs/2408.03314} {Scaling llm test-time compute optimally can be more effective than scaling model parameters}.
\newblock \emph{Preprint}, arXiv:2408.03314.

\bibitem[{Song et~al.(2025)Song, Su, Qu, Zhou, and Cheng}]{song2025prmbenchfinegrainedchallengingbenchmark}
Mingyang Song, Zhaochen Su, Xiaoye Qu, Jiawei Zhou, and Yu~Cheng. 2025.
\newblock \href {https://arxiv.org/abs/2501.03124} {Prmbench: A fine-grained and challenging benchmark for process-level reward models}.
\newblock \emph{Preprint}, arXiv:2501.03124.

\bibitem[{Sprague et~al.(2024)Sprague, Yin, Rodriguez, Jiang, Wadhwa, Singhal, Zhao, Ye, Mahowald, and Durrett}]{sprague2024cotcotchainofthoughthelps}
Zayne Sprague, Fangcong Yin, Juan~Diego Rodriguez, Dongwei Jiang, Manya Wadhwa, Prasann Singhal, Xinyu Zhao, Xi~Ye, Kyle Mahowald, and Greg Durrett. 2024.
\newblock \href {https://arxiv.org/abs/2409.12183} {To cot or not to cot? chain-of-thought helps mainly on math and symbolic reasoning}.
\newblock \emph{Preprint}, arXiv:2409.12183.

\bibitem[{Sun et~al.(2024)Sun, Yu, Shen, Liu, Yang, Welleck, and Gan}]{sun2024easytohardgeneralizationscalablealignment}
Zhiqing Sun, Longhui Yu, Yikang Shen, Weiyang Liu, Yiming Yang, Sean Welleck, and Chuang Gan. 2024.
\newblock \href {https://arxiv.org/abs/2403.09472} {Easy-to-hard generalization: Scalable alignment beyond human supervision}.
\newblock \emph{Preprint}, arXiv:2403.09472.

\bibitem[{{\v{S}}uster and Daelemans(2018)}]{suster-daelemans-2018-clicr}
Simon {\v{S}}uster and Walter Daelemans. 2018.
\newblock \href {https://doi.org/10.18653/v1/N18-1140} {{C}li{CR}: a dataset of clinical case reports for machine reading comprehension}.
\newblock In \emph{Proceedings of the 2018 Conference of the North {A}merican Chapter of the Association for Computational Linguistics: Human Language Technologies, Volume 1 (Long Papers)}, pages 1551--1563, New Orleans, Louisiana. Association for Computational Linguistics.

\bibitem[{Suzgun et~al.(2022)Suzgun, Scales, Schärli, Gehrmann, Tay, Chung, Chowdhery, Le, Chi, Zhou, and Wei}]{suzgun2022challengingbigbenchtaskschainofthought}
Mirac Suzgun, Nathan Scales, Nathanael Schärli, Sebastian Gehrmann, Yi~Tay, Hyung~Won Chung, Aakanksha Chowdhery, Quoc~V. Le, Ed~H. Chi, Denny Zhou, and Jason Wei. 2022.
\newblock \href {https://arxiv.org/abs/2210.09261} {Challenging big-bench tasks and whether chain-of-thought can solve them}.
\newblock \emph{Preprint}, arXiv:2210.09261.

\bibitem[{Tafjord et~al.(2021)Tafjord, Dalvi, and Clark}]{tafjord-etal-2021-proofwriter}
Oyvind Tafjord, Bhavana Dalvi, and Peter Clark. 2021.
\newblock \href {https://doi.org/10.18653/v1/2021.findings-acl.317} {{P}roof{W}riter: Generating implications, proofs, and abductive statements over natural language}.
\newblock In \emph{Findings of the Association for Computational Linguistics: ACL-IJCNLP 2021}, pages 3621--3634, Online. Association for Computational Linguistics.

\bibitem[{Talmor and Berant(2018)}]{talmor-berant-2018-web}
Alon Talmor and Jonathan Berant. 2018.
\newblock \href {https://doi.org/10.18653/v1/N18-1059} {The web as a knowledge-base for answering complex questions}.
\newblock In \emph{Proceedings of the 2018 Conference of the North {A}merican Chapter of the Association for Computational Linguistics: Human Language Technologies, Volume 1 (Long Papers)}, pages 641--651, New Orleans, Louisiana. Association for Computational Linguistics.

\bibitem[{Talmor et~al.(2019)Talmor, Herzig, Lourie, and Berant}]{talmor-etal-2019-commonsenseqa}
Alon Talmor, Jonathan Herzig, Nicholas Lourie, and Jonathan Berant. 2019.
\newblock \href {https://doi.org/10.18653/v1/N19-1421} {{C}ommonsense{QA}: A question answering challenge targeting commonsense knowledge}.
\newblock In \emph{Proceedings of the 2019 Conference of the North {A}merican Chapter of the Association for Computational Linguistics: Human Language Technologies, Volume 1 (Long and Short Papers)}, pages 4149--4158, Minneapolis, Minnesota. Association for Computational Linguistics.

\bibitem[{Thorne et~al.(2018)Thorne, Vlachos, Christodoulopoulos, and Mittal}]{thorne-etal-2018-fever}
James Thorne, Andreas Vlachos, Christos Christodoulopoulos, and Arpit Mittal. 2018.
\newblock \href {https://doi.org/10.18653/v1/N18-1074} {{FEVER}: a large-scale dataset for fact extraction and {VER}ification}.
\newblock In \emph{Proceedings of the 2018 Conference of the North {A}merican Chapter of the Association for Computational Linguistics: Human Language Technologies, Volume 1 (Long Papers)}, pages 809--819, New Orleans, Louisiana. Association for Computational Linguistics.

\bibitem[{Tian et~al.(2021)Tian, Li, Chen, Xiao, He, and Jin}]{tian-etal-2021-diagnosing}
Jidong Tian, Yitian Li, Wenqing Chen, Liqiang Xiao, Hao He, and Yaohui Jin. 2021.
\newblock \href {https://doi.org/10.18653/v1/2021.emnlp-main.303} {Diagnosing the first-order logical reasoning ability through {L}ogic{NLI}}.
\newblock In \emph{Proceedings of the 2021 Conference on Empirical Methods in Natural Language Processing}, pages 3738--3747, Online and Punta Cana, Dominican Republic. Association for Computational Linguistics.

\bibitem[{Toroghi et~al.(2024)Toroghi, Guo, Abdollah~Pour, and Sanner}]{toroghi-etal-2024-right}
Armin Toroghi, Willis Guo, Mohammad~Mahdi Abdollah~Pour, and Scott Sanner. 2024.
\newblock \href {https://doi.org/10.18653/v1/2024.emnlp-main.378} {Right for right reasons: Large language models for verifiable commonsense knowledge graph question answering}.
\newblock In \emph{Proceedings of the 2024 Conference on Empirical Methods in Natural Language Processing}, pages 6601--6633, Miami, Florida, USA. Association for Computational Linguistics.

\bibitem[{Trivedi et~al.(2022)Trivedi, Balasubramanian, Khot, and Sabharwal}]{trivedi-etal-2022-musique}
Harsh Trivedi, Niranjan Balasubramanian, Tushar Khot, and Ashish Sabharwal. 2022.
\newblock \href {https://doi.org/10.1162/tacl_a_00475} {{M}u{S}i{Q}ue: Multihop questions via single-hop question composition}.
\newblock \emph{Transactions of the Association for Computational Linguistics}, 10:539--554.

\bibitem[{Tyagi et~al.(2024)Tyagi, Parmar, Kulkarni, Rrv, Patel, Nakamura, Mitra, and Baral}]{tyagi-etal-2024-step}
Nemika Tyagi, Mihir Parmar, Mohith Kulkarni, Aswin Rrv, Nisarg Patel, Mutsumi Nakamura, Arindam Mitra, and Chitta Baral. 2024.
\newblock \href {https://doi.org/10.18653/v1/2024.emnlp-main.1111} {Step-by-step reasoning to solve grid puzzles: Where do {LLM}s falter?}
\newblock In \emph{Proceedings of the 2024 Conference on Empirical Methods in Natural Language Processing}, pages 19898--19915, Miami, Florida, USA. Association for Computational Linguistics.

\bibitem[{Tyen et~al.(2024)Tyen, Mansoor, Carbune, Chen, and Mak}]{tyen-etal-2024-llms}
Gladys Tyen, Hassan Mansoor, Victor Carbune, Peter Chen, and Tony Mak. 2024.
\newblock \href {https://doi.org/10.18653/v1/2024.findings-acl.826} {{LLM}s cannot find reasoning errors, but can correct them given the error location}.
\newblock In \emph{Findings of the Association for Computational Linguistics: ACL 2024}, pages 13894--13908, Bangkok, Thailand. Association for Computational Linguistics.

\bibitem[{Uesato et~al.(2022)Uesato, Kushman, Kumar, Song, Siegel, Wang, Creswell, Irving, and Higgins}]{uesato2022solvingmathwordproblems}
Jonathan Uesato, Nate Kushman, Ramana Kumar, Francis Song, Noah Siegel, Lisa Wang, Antonia Creswell, Geoffrey Irving, and Irina Higgins. 2022.
\newblock \href {https://arxiv.org/abs/2211.14275} {Solving math word problems with process- and outcome-based feedback}.
\newblock \emph{Preprint}, arXiv:2211.14275.

\bibitem[{Wang et~al.(2024{\natexlab{a}})Wang, Zheng, Chen, Liu, Dou, Huang, Shen, Jin, Zhou, Shi, Gao, Xu, Zhou, Fan, Xi, Zhao, Wang, Ji, Yan, Shen, Chen, Gui, Zhang, Qiu, Huang, Wu, and Jiang}]{wang2024secretsrlhflargelanguage}
Binghai Wang, Rui Zheng, Lu~Chen, Yan Liu, Shihan Dou, Caishuang Huang, Wei Shen, Senjie Jin, Enyu Zhou, Chenyu Shi, Songyang Gao, Nuo Xu, Yuhao Zhou, Xiaoran Fan, Zhiheng Xi, Jun Zhao, Xiao Wang, Tao Ji, Hang Yan, Lixing Shen, Zhan Chen, Tao Gui, Qi~Zhang, Xipeng Qiu, Xuanjing Huang, Zuxuan Wu, and Yu-Gang Jiang. 2024{\natexlab{a}}.
\newblock \href {https://arxiv.org/abs/2401.06080} {Secrets of rlhf in large language models part ii: Reward modeling}.
\newblock \emph{Preprint}, arXiv:2401.06080.

\bibitem[{Wang et~al.(2023{\natexlab{a}})Wang, Min, Deng, Shen, Wu, Zettlemoyer, and Sun}]{wang-etal-2023-towards}
Boshi Wang, Sewon Min, Xiang Deng, Jiaming Shen, You Wu, Luke Zettlemoyer, and Huan Sun. 2023{\natexlab{a}}.
\newblock \href {https://doi.org/10.18653/v1/2023.acl-long.153} {Towards understanding chain-of-thought prompting: An empirical study of what matters}.
\newblock In \emph{Proceedings of the 61st Annual Meeting of the Association for Computational Linguistics (Volume 1: Long Papers)}, pages 2717--2739, Toronto, Canada. Association for Computational Linguistics.

\bibitem[{Wang et~al.(2025{\natexlab{a}})Wang, Gao, Xu, Liu, Hussein, Korsapati, Labban, Iheasirim, Hassan, Anil, Bartlett, and Sun}]{wang2025processsupervisedrewardmodelsverifying}
Hanyin Wang, Chufan Gao, Qiping Xu, Bolun Liu, Guleid Hussein, Hariprasad Korsapati, Mohamad~El Labban, Kingsley Iheasirim, Mohamed Hassan, Gokhan Anil, Brian Bartlett, and Jimeng Sun. 2025{\natexlab{a}}.
\newblock \href {https://arxiv.org/abs/2412.12583} {Process-supervised reward models for verifying clinical note generation: A scalable approach guided by domain expertise}.
\newblock \emph{Preprint}, arXiv:2412.12583.

\bibitem[{Wang et~al.(2024{\natexlab{b}})Wang, Sun, Li, and Gao}]{wang-etal-2024-boosting-language}
Jianing Wang, Qiushi Sun, Xiang Li, and Ming Gao. 2024{\natexlab{b}}.
\newblock \href {https://doi.org/10.18653/v1/2024.acl-long.271} {Boosting language models reasoning with chain-of-knowledge prompting}.
\newblock In \emph{Proceedings of the 62nd Annual Meeting of the Association for Computational Linguistics (Volume 1: Long Papers)}, pages 4958--4981, Bangkok, Thailand. Association for Computational Linguistics.

\bibitem[{Wang et~al.(2024{\natexlab{c}})Wang, Li, Shao, Xu, Dai, Li, Chen, Wu, and Sui}]{wang-etal-2024-math}
Peiyi Wang, Lei Li, Zhihong Shao, Runxin Xu, Damai Dai, Yifei Li, Deli Chen, Yu~Wu, and Zhifang Sui. 2024{\natexlab{c}}.
\newblock \href {https://doi.org/10.18653/v1/2024.acl-long.510} {Math-shepherd: Verify and reinforce {LLM}s step-by-step without human annotations}.
\newblock In \emph{Proceedings of the 62nd Annual Meeting of the Association for Computational Linguistics (Volume 1: Long Papers)}, pages 9426--9439, Bangkok, Thailand. Association for Computational Linguistics.

\bibitem[{Wang et~al.(2023{\natexlab{b}})Wang, Wei, Schuurmans, Le, Chi, Narang, Chowdhery, and Zhou}]{DBLP:conf/iclr/0002WSLCNCZ23}
Xuezhi Wang, Jason Wei, Dale Schuurmans, Quoc~V. Le, Ed~H. Chi, Sharan Narang, Aakanksha Chowdhery, and Denny Zhou. 2023{\natexlab{b}}.
\newblock \href {https://openreview.net/forum?id=1PL1NIMMrw} {Self-consistency improves chain of thought reasoning in language models}.
\newblock In \emph{The Eleventh International Conference on Learning Representations, {ICLR} 2023, Kigali, Rwanda, May 1-5, 2023}. OpenReview.net.

\bibitem[{Wang et~al.(2025{\natexlab{b}})Wang, Yang, Wang, and Wei}]{wang2025examiningfalsepositivesinference}
Yu~Wang, Nan Yang, Liang Wang, and Furu Wei. 2025{\natexlab{b}}.
\newblock \href {https://arxiv.org/abs/2502.06217} {Examining false positives under inference scaling for mathematical reasoning}.
\newblock \emph{Preprint}, arXiv:2502.06217.

\bibitem[{Wang et~al.(2024{\natexlab{d}})Wang, Li, Yang, Liu, Hao, Chen, Chu, and Sui}]{wang-etal-2024-analyzing}
Zecheng Wang, Chunshan Li, Zhao Yang, Qingbin Liu, Yanchao Hao, Xi~Chen, Dianhui Chu, and Dianbo Sui. 2024{\natexlab{d}}.
\newblock \href {https://aclanthology.org/2024.lrec-main.81/} {Analyzing chain-of-thought prompting in black-box large language models via estimated {V}-information}.
\newblock In \emph{Proceedings of the 2024 Joint International Conference on Computational Linguistics, Language Resources and Evaluation (LREC-COLING 2024)}, pages 893--903, Torino, Italia. ELRA and ICCL.

\bibitem[{Wei et~al.(2022)Wei, Wang, Schuurmans, Bosma, ichter, Xia, Chi, Le, and Zhou}]{NEURIPS2022_9d560961}
Jason Wei, Xuezhi Wang, Dale Schuurmans, Maarten Bosma, brian ichter, Fei Xia, Ed~Chi, Quoc~V Le, and Denny Zhou. 2022.
\newblock \href {https://proceedings.neurips.cc/paper_files/paper/2022/file/9d5609613524ecf4f15af0f7b31abca4-Paper-Conference.pdf} {Chain-of-thought prompting elicits reasoning in large language models}.
\newblock In \emph{Advances in Neural Information Processing Systems}, volume~35, pages 24824--24837. Curran Associates, Inc.

\bibitem[{Wei et~al.(2025)Wei, Liu, Wu, and Fang}]{wei2025surveyfeedbackbasedmultistepreasoning}
Ting-Ruen Wei, Haowei Liu, Xuyang Wu, and Yi~Fang. 2025.
\newblock \href {https://arxiv.org/abs/2502.14333} {A survey on feedback-based multi-step reasoning for large language models on mathematics}.
\newblock \emph{Preprint}, arXiv:2502.14333.

\bibitem[{Wu et~al.(2024{\natexlab{a}})Wu, Gu, Yin, Peng, and Chang}]{wu-etal-2024-synchronous}
Di~Wu, Jia-Chen Gu, Fan Yin, Nanyun Peng, and Kai-Wei Chang. 2024{\natexlab{a}}.
\newblock \href {https://doi.org/10.18653/v1/2024.emnlp-main.527} {Synchronous faithfulness monitoring for trustworthy retrieval-augmented generation}.
\newblock In \emph{Proceedings of the 2024 Conference on Empirical Methods in Natural Language Processing}, pages 9390--9406, Miami, Florida, USA. Association for Computational Linguistics.

\bibitem[{Wu et~al.(2024{\natexlab{b}})Wu, Yang, Wang, Okumura, and Zhang}]{wu2024cofcastepwisecounterfactualmultihop}
Jian Wu, Linyi Yang, Zhen Wang, Manabu Okumura, and Yue Zhang. 2024{\natexlab{b}}.
\newblock \href {https://arxiv.org/abs/2402.11924} {Cofca: A step-wise counterfactual multi-hop qa benchmark}.
\newblock \emph{Preprint}, arXiv:2402.11924.

\bibitem[{Wu et~al.(2024{\natexlab{c}})Wu, Li, Wang, Xia, Xiong, Wang, Yu, Chen, Kveton, Yao, Shang, and McAuley}]{wu2024oceanofflinechainofthoughtevaluation}
Junda Wu, Xintong Li, Ruoyu Wang, Yu~Xia, Yuxin Xiong, Jianing Wang, Tong Yu, Xiang Chen, Branislav Kveton, Lina Yao, Jingbo Shang, and Julian McAuley. 2024{\natexlab{c}}.
\newblock \href {https://arxiv.org/abs/2410.23703} {Ocean: Offline chain-of-thought evaluation and alignment in large language models}.
\newblock \emph{Preprint}, arXiv:2410.23703.

\bibitem[{Wu et~al.(2024{\natexlab{d}})Wu, Zhang, and Zhao}]{wu-etal-2024-mitigating}
Yexin Wu, Zhuosheng Zhang, and Hai Zhao. 2024{\natexlab{d}}.
\newblock \href {https://aclanthology.org/2024.lrec-main.990} {Mitigating misleading chain-of-thought reasoning with selective filtering}.
\newblock In \emph{Proceedings of the 2024 Joint International Conference on Computational Linguistics, Language Resources and Evaluation (LREC-COLING 2024)}, pages 11325--11340, Torino, Italia. ELRA and ICCL.

\bibitem[{Xia et~al.(2025)Xia, Li, Liu, Wu, and Liu}]{xia2025evaluatingmathematicalreasoningaccuracy}
Shijie Xia, Xuefeng Li, Yixin Liu, Tongshuang Wu, and Pengfei Liu. 2025.
\newblock \href {https://arxiv.org/abs/2404.05692} {Evaluating mathematical reasoning beyond accuracy}.
\newblock \emph{Preprint}, arXiv:2404.05692.

\bibitem[{Xiao and Wang(2021)}]{xiao-wang-2021-hallucination}
Yijun Xiao and William~Yang Wang. 2021.
\newblock \href {https://doi.org/10.18653/v1/2021.eacl-main.236} {On hallucination and predictive uncertainty in conditional language generation}.
\newblock In \emph{Proceedings of the 16th Conference of the European Chapter of the Association for Computational Linguistics: Main Volume}, pages 2734--2744, Online. Association for Computational Linguistics.

\bibitem[{Xie et~al.(2024)Xie, Goyal, Zheng, Kan, Lillicrap, Kawaguchi, and Shieh}]{xie2024montecarlotreesearch}
Yuxi Xie, Anirudh Goyal, Wenyue Zheng, Min-Yen Kan, Timothy~P. Lillicrap, Kenji Kawaguchi, and Michael Shieh. 2024.
\newblock \href {https://arxiv.org/abs/2405.00451} {Monte carlo tree search boosts reasoning via iterative preference learning}.
\newblock \emph{Preprint}, arXiv:2405.00451.

\bibitem[{Xie et~al.(2020)Xie, Thiem, Martin, Wainwright, Marmorstein, and Jansen}]{xie-etal-2020-worldtree}
Zhengnan Xie, Sebastian Thiem, Jaycie Martin, Elizabeth Wainwright, Steven Marmorstein, and Peter Jansen. 2020.
\newblock \href {https://aclanthology.org/2020.lrec-1.671/} {{W}orld{T}ree v2: A corpus of science-domain structured explanations and inference patterns supporting multi-hop inference}.
\newblock In \emph{Proceedings of the Twelfth Language Resources and Evaluation Conference}, pages 5456--5473, Marseille, France. European Language Resources Association.

\bibitem[{Yang et~al.(2023)Yang, Swope, Gu, Chalamala, Song, Yu, Godil, Prenger, and Anandkumar}]{yang2023leandojo}
Kaiyu Yang, Aidan Swope, Alex Gu, Rahul Chalamala, Peiyang Song, Shixing Yu, Saad Godil, Ryan Prenger, and Anima Anandkumar. 2023.
\newblock {LeanDojo}: Theorem proving with retrieval-augmented language models.
\newblock In \emph{Neural Information Processing Systems (NeurIPS)}.

\bibitem[{Yang et~al.(2018)Yang, Qi, Zhang, Bengio, Cohen, Salakhutdinov, and Manning}]{yang-etal-2018-hotpotqa}
Zhilin Yang, Peng Qi, Saizheng Zhang, Yoshua Bengio, William Cohen, Ruslan Salakhutdinov, and Christopher~D. Manning. 2018.
\newblock \href {https://doi.org/10.18653/v1/D18-1259} {{H}otpot{QA}: A dataset for diverse, explainable multi-hop question answering}.
\newblock In \emph{Proceedings of the 2018 Conference on Empirical Methods in Natural Language Processing}, pages 2369--2380, Brussels, Belgium. Association for Computational Linguistics.

\bibitem[{Yao et~al.(2023)Yao, Yu, Zhao, Shafran, Griffiths, Cao, and Narasimhan}]{NEURIPS2023_271db992}
Shunyu Yao, Dian Yu, Jeffrey Zhao, Izhak Shafran, Tom Griffiths, Yuan Cao, and Karthik Narasimhan. 2023.
\newblock \href {https://proceedings.neurips.cc/paper_files/paper/2023/file/271db9922b8d1f4dd7aaef84ed5ac703-Paper-Conference.pdf} {Tree of thoughts: Deliberate problem solving with large language models}.
\newblock In \emph{Advances in Neural Information Processing Systems}, volume~36, pages 11809--11822. Curran Associates, Inc.

\bibitem[{Yuan et~al.(2024)Yuan, Li, Chen, Cui, Ding, Zhang, Zhou, Liu, and Peng}]{yuan2024implicitprm}
Lifan Yuan, Wendi Li, Huayu Chen, Ganqu Cui, Ning Ding, Kaiyan Zhang, Bowen Zhou, Zhiyuan Liu, and Hao Peng. 2024.
\newblock Free process rewards without process labels.
\newblock \emph{arXiv preprint arXiv:2412.01981}.

\bibitem[{Zeng et~al.(2025)Zeng, Zhang, Wu, Classen, Chae, Ewer, Lee, Kim, Kang, Kunde, Fan, Kim, Koo, Ramchandran, Papailiopoulos, and Lee}]{zeng2025versaprmmultidomainprocessreward}
Thomas Zeng, Shuibai Zhang, Shutong Wu, Christian Classen, Daewon Chae, Ethan Ewer, Minjae Lee, Heeju Kim, Wonjun Kang, Jackson Kunde, Ying Fan, Jungtaek Kim, Hyung~Il Koo, Kannan Ramchandran, Dimitris Papailiopoulos, and Kangwook Lee. 2025.
\newblock \href {https://arxiv.org/abs/2502.06737} {Versaprm: Multi-domain process reward model via synthetic reasoning data}.
\newblock \emph{Preprint}, arXiv:2502.06737.

\bibitem[{Zeng et~al.(2024{\natexlab{a}})Zeng, Chen, Liu, Jiang, and Jia}]{zeng2024mrgsm8kmetareasoningbenchmarklarge}
Zhongshen Zeng, Pengguang Chen, Shu Liu, Haiyun Jiang, and Jiaya Jia. 2024{\natexlab{a}}.
\newblock \href {https://arxiv.org/abs/2312.17080} {Mr-gsm8k: A meta-reasoning benchmark for large language model evaluation}.
\newblock \emph{Preprint}, arXiv:2312.17080.

\bibitem[{Zeng et~al.(2024{\natexlab{b}})Zeng, Liu, Wan, Li, Chen, Dai, Yao, Xu, Qi, Zhao, Shen, Lu, Tan, Chen, Zhang, Shi, Wang, Guo, and Jia}]{NEURIPS2024_d81cb1f4}
Zhongshen Zeng, Yinhong Liu, Yingjia Wan, Jingyao Li, Pengguang Chen, Jianbo Dai, Yuxuan Yao, Rongwu Xu, Zehan Qi, Wanru Zhao, Linling Shen, Jianqiao Lu, Haochen Tan, Yukang Chen, Hao Zhang, Zhan Shi, Bailin Wang, Zhijiang Guo, and Jiaya Jia. 2024{\natexlab{b}}.
\newblock \href {https://proceedings.neurips.cc/paper_files/paper/2024/file/d81cb1f4dc6e13aeb45553f80b3d6837-Paper-Conference.pdf} {Mr-ben: A meta-reasoning benchmark for evaluating system-2 thinking in llms}.
\newblock In \emph{Advances in Neural Information Processing Systems}, volume~37, pages 119466--119546. Curran Associates, Inc.

\bibitem[{Zha et~al.(2023)Zha, Yang, Li, and Hu}]{zha-etal-2023-alignscore}
Yuheng Zha, Yichi Yang, Ruichen Li, and Zhiting Hu. 2023.
\newblock \href {https://doi.org/10.18653/v1/2023.acl-long.634} {{A}lign{S}core: Evaluating factual consistency with a unified alignment function}.
\newblock In \emph{Proceedings of the 61st Annual Meeting of the Association for Computational Linguistics (Volume 1: Long Papers)}, pages 11328--11348, Toronto, Canada. Association for Computational Linguistics.

\bibitem[{Zhang et~al.(2024{\natexlab{a}})Zhang, Zhoubian, Hu, Yue, Dong, and Tang}]{NEURIPS2024_76ec4dc3}
Dan Zhang, Sining Zhoubian, Ziniu Hu, Yisong Yue, Yuxiao Dong, and Jie Tang. 2024{\natexlab{a}}.
\newblock \href {https://proceedings.neurips.cc/paper_files/paper/2024/file/76ec4dc30e9faaf0e4b6093eaa377218-Paper-Conference.pdf} {Rest-mcts* : Llm self-training via process reward guided tree search}.
\newblock In \emph{Advances in Neural Information Processing Systems}, volume~37, pages 64735--64772. Curran Associates, Inc.

\bibitem[{Zhang et~al.(2023{\natexlab{a}})Zhang, Chen, Zhang, Keung, Liu, Zan, Mao, Lou, and Chen}]{zhang-etal-2023-repocoder}
Fengji Zhang, Bei Chen, Yue Zhang, Jacky Keung, Jin Liu, Daoguang Zan, Yi~Mao, Jian-Guang Lou, and Weizhu Chen. 2023{\natexlab{a}}.
\newblock \href {https://doi.org/10.18653/v1/2023.emnlp-main.151} {{R}epo{C}oder: Repository-level code completion through iterative retrieval and generation}.
\newblock In \emph{Proceedings of the 2023 Conference on Empirical Methods in Natural Language Processing}, pages 2471--2484, Singapore. Association for Computational Linguistics.

\bibitem[{Zhang et~al.(2024{\natexlab{b}})Zhang, Li, Zhang, Yin, Liu, and Moshfeghi}]{zhang-etal-2024-geoeval}
Jiaxin Zhang, Zhong-Zhi Li, Ming-Liang Zhang, Fei Yin, Cheng-Lin Liu, and Yashar Moshfeghi. 2024{\natexlab{b}}.
\newblock \href {https://doi.org/10.18653/v1/2024.findings-acl.73} {{G}eo{E}val: Benchmark for evaluating {LLM}s and multi-modal models on geometry problem-solving}.
\newblock In \emph{Findings of the Association for Computational Linguistics: ACL 2024}, pages 1258--1276, Bangkok, Thailand. Association for Computational Linguistics.

\bibitem[{Zhang et~al.(2024{\natexlab{c}})Zhang, Hosseini, Bansal, Kazemi, Kumar, and Agarwal}]{zhang2024generativeverifiersrewardmodeling}
Lunjun Zhang, Arian Hosseini, Hritik Bansal, Mehran Kazemi, Aviral Kumar, and Rishabh Agarwal. 2024{\natexlab{c}}.
\newblock \href {https://arxiv.org/abs/2408.15240} {Generative verifiers: Reward modeling as next-token prediction}.
\newblock \emph{Preprint}, arXiv:2408.15240.

\bibitem[{Zhang et~al.(2023{\natexlab{b}})Zhang, Qiu, Guo, Deng, Zhang, Zhang, Zhou, Wang, and Fu}]{zhang-etal-2023-enhancing-uncertainty}
Tianhang Zhang, Lin Qiu, Qipeng Guo, Cheng Deng, Yue Zhang, Zheng Zhang, Chenghu Zhou, Xinbing Wang, and Luoyi Fu. 2023{\natexlab{b}}.
\newblock \href {https://doi.org/10.18653/v1/2023.emnlp-main.58} {Enhancing uncertainty-based hallucination detection with stronger focus}.
\newblock In \emph{Proceedings of the 2023 Conference on Empirical Methods in Natural Language Processing}, pages 915--932, Singapore. Association for Computational Linguistics.

\bibitem[{Zhang et~al.(2024{\natexlab{d}})Zhang, Du, Pang, Liu, Gao, and Lin}]{zhang2024chainpreferenceoptimizationimproving}
Xuan Zhang, Chao Du, Tianyu Pang, Qian Liu, Wei Gao, and Min Lin. 2024{\natexlab{d}}.
\newblock \href {https://arxiv.org/abs/2406.09136} {Chain of preference optimization: Improving chain-of-thought reasoning in llms}.
\newblock \emph{Preprint}, arXiv:2406.09136.

\bibitem[{Zhang et~al.(2025)Zhang, Zheng, Wu, Zhang, Lin, Yu, Liu, Zhou, and Lin}]{zhang2025lessonsdevelopingprocessreward}
Zhenru Zhang, Chujie Zheng, Yangzhen Wu, Beichen Zhang, Runji Lin, Bowen Yu, Dayiheng Liu, Jingren Zhou, and Junyang Lin. 2025.
\newblock \href {https://arxiv.org/abs/2501.07301} {The lessons of developing process reward models in mathematical reasoning}.
\newblock \emph{Preprint}, arXiv:2501.07301.

\bibitem[{Zheng et~al.(2024{\natexlab{a}})Zheng, Zhang, Zhang, Lin, Lu, Yu, Liu, Zhou, and Lin}]{zheng2024processbenchidentifyingprocesserrors}
Chujie Zheng, Zhenru Zhang, Beichen Zhang, Runji Lin, Keming Lu, Bowen Yu, Dayiheng Liu, Jingren Zhou, and Junyang Lin. 2024{\natexlab{a}}.
\newblock \href {https://arxiv.org/abs/2412.06559} {Processbench: Identifying process errors in mathematical reasoning}.
\newblock \emph{Preprint}, arXiv:2412.06559.

\bibitem[{Zheng et~al.(2023)Zheng, Chiang, Sheng, Zhuang, Wu, Zhuang, Lin, Li, Li, Xing, Zhang, Gonzalez, and Stoica}]{NEURIPS2023_91f18a12}
Lianmin Zheng, Wei-Lin Chiang, Ying Sheng, Siyuan Zhuang, Zhanghao Wu, Yonghao Zhuang, Zi~Lin, Zhuohan Li, Dacheng Li, Eric Xing, Hao Zhang, Joseph~E Gonzalez, and Ion Stoica. 2023.
\newblock \href {https://proceedings.neurips.cc/paper_files/paper/2023/file/91f18a1287b398d378ef22505bf41832-Paper-Datasets_and_Benchmarks.pdf} {Judging llm-as-a-judge with mt-bench and chatbot arena}.
\newblock In \emph{Advances in Neural Information Processing Systems}, volume~36, pages 46595--46623. Curran Associates, Inc.

\bibitem[{Zheng et~al.(2024{\natexlab{b}})Zheng, Lou, Cao, Wen, Ji, Lin, Lu, Han, Zhang, and Sun}]{zheng2024criticcotboostingreasoningabilities}
Xin Zheng, Jie Lou, Boxi Cao, Xueru Wen, Yuqiu Ji, Hongyu Lin, Yaojie Lu, Xianpei Han, Debing Zhang, and Le~Sun. 2024{\natexlab{b}}.
\newblock \href {https://arxiv.org/abs/2408.16326} {Critic-cot: Boosting the reasoning abilities of large language model via chain-of-thoughts critic}.
\newblock \emph{Preprint}, arXiv:2408.16326.

\bibitem[{Zhong et~al.(2021)Zhong, Wang, Tang, Xu, Guo, Wang, Yin, Zhou, and Duan}]{zhong2021arlsatinvestigatinganalyticalreasoning}
Wanjun Zhong, Siyuan Wang, Duyu Tang, Zenan Xu, Daya Guo, Jiahai Wang, Jian Yin, Ming Zhou, and Nan Duan. 2021.
\newblock \href {https://arxiv.org/abs/2104.06598} {Ar-lsat: Investigating analytical reasoning of text}.
\newblock \emph{Preprint}, arXiv:2104.06598.

\bibitem[{Zhou et~al.(2024)Zhou, Tao, Zhu, Luo, Wang, and Han}]{zhou2024languagemodelsperformrobust}
Zhanke Zhou, Rong Tao, Jianing Zhu, Yiwen Luo, Zengmao Wang, and Bo~Han. 2024.
\newblock \href {https://arxiv.org/abs/2410.23856} {Can language models perform robust reasoning in chain-of-thought prompting with noisy rationales?}
\newblock \emph{Preprint}, arXiv:2410.23856.

\bibitem[{Zhu et~al.(2024{\natexlab{a}})Zhu, Hwang, Dugan, and Callison-Burch}]{zhu-etal-2024-fanoutqa}
Andrew Zhu, Alyssa Hwang, Liam Dugan, and Chris Callison-Burch. 2024{\natexlab{a}}.
\newblock \href {https://doi.org/10.18653/v1/2024.acl-short.2} {{F}an{O}ut{QA}: A multi-hop, multi-document question answering benchmark for large language models}.
\newblock In \emph{Proceedings of the 62nd Annual Meeting of the Association for Computational Linguistics (Volume 2: Short Papers)}, pages 18--37, Bangkok, Thailand. Association for Computational Linguistics.

\bibitem[{Zhu et~al.(2024{\natexlab{b}})Zhu, Zhang, Xie, and Su}]{zhu2024deductivebeamsearchdecoding}
Tinghui Zhu, Kai Zhang, Jian Xie, and Yu~Su. 2024{\natexlab{b}}.
\newblock \href {https://arxiv.org/abs/2401.17686} {Deductive beam search: Decoding deducible rationale for chain-of-thought reasoning}.
\newblock \emph{Preprint}, arXiv:2401.17686.

\bibitem[{Zhu et~al.(2025)Zhu, Li, Jiang, Li, Mei, Jin, and Dong}]{zhu2025uncertaintyguidedchainofthoughtcodegeneration}
Yuqi Zhu, Ge~Li, Xue Jiang, Jia Li, Hong Mei, Zhi Jin, and Yihong Dong. 2025.
\newblock \href {https://arxiv.org/abs/2503.15341} {Uncertainty-guided chain-of-thought for code generation with llms}.
\newblock \emph{Preprint}, arXiv:2503.15341.

\bibitem[{Zuo et~al.(2025)Zuo, Qu, Li, Chen, Zhu, Hua, Zhang, Ding, and Zhou}]{zuo2025medxpertqabenchmarkingexpertlevelmedical}
Yuxin Zuo, Shang Qu, Yifei Li, Zhangren Chen, Xuekai Zhu, Ermo Hua, Kaiyan Zhang, Ning Ding, and Bowen Zhou. 2025.
\newblock \href {https://arxiv.org/abs/2501.18362} {Medxpertqa: Benchmarking expert-level medical reasoning and understanding}.
\newblock \emph{Preprint}, arXiv:2501.18362.

\end{thebibliography}

\newpage
\appendix

\section{Tasks}
\label{sec:appendix-task}

This section describes different reasoning tasks and datasets in more detail. While all reasoning tasks fundamentally share the same criteria, literature about a specific task has focused on one criterion over others. For instance, evaluators for factual reasoning tasks often emphasized detecting infactual statements, while evaluators for math reasoning tasks aimed for invalid statements. These discrepancies are one of the fundamental causes of the divergence of the terminologies and definitions in the field.

\subsection{Multi-hop Question Answering}
\label{sec:appendix-task-mhqa}

Multi-hop question answering (MHQA) tasks require taking information from multiple sources to derive the correct answer \citep{yang-etal-2018-hotpotqa}. MHQA is often divided into two subcategories, \textbf{factual reasoning} and \textbf{commonsense reasoning}.

Answering factual MHQAs can be seen as finding the sequence of \textit{bridging entities} that leads to the final answer \citep{yang-etal-2018-hotpotqa, talmor-berant-2018-web, kwiatkowski-etal-2019-natural}. For example, to solve a factual MHQA question \textit{"The Argentine PGA Championship record holder has won how many tournaments worldwide?"}, one must first find who the Argentine PGA championship record holder is (bridging entity) and determine how many tournaments he has won worldwide. As bridging entity identification does not require sophisticated reasoning ability compared to other tasks, reasoning trace evaluation on factual MHQA mostly focuses on the factuality based on semantic alignment between the query (retrieved documents) and the trace \citep{DBLP:conf/iclr/GolovnevaCPCZFC23}.

In contrast, an inference step in commonsense MHQAs \citep{clark2018thinksolvedquestionanswering, mihaylov2018suitarmorconductelectricity, talmor-etal-2019-commonsenseqa, bisk2019piqareasoningphysicalcommonsense, geva-etal-2021-aristotle, trivedi-etal-2022-musique} can require information that is not present in the query. The form of such commonsense knowledge can be diverse, ranging from well-known facts (\textit{Paris is in France.}) to logical rules (\textit{If A was born after B was dead, they have never met each other}). Due to these implicit steps, factuality, validity, and coherence are often hard to separate in evaluating commonsense reasoning traces \citep{jacovi-etal-2024-chain, NEURIPS2024_d81cb1f4}. Furthermore, due to the inherent subjectiveness of validity and coherence in commonsense reasoning, there might be non-negligible inter-annotator disagreement on certain questions \citep{jacovi-etal-2024-chain}

LLMs are known to achieve strong performance in challenging MHQA datasets such as ARC-Challenge and PIQA, sometimes exceeding human performance \citep{openai2024gpt4technicalreport, anil2023palm2technicalreport}. However, multiple studies report that even modern LLMs like GPT-4 \citep{openai2024gpt4technicalreport} are vulnerable to errors, such as failing to correctly adhere to long evidence \citep{zhu-etal-2024-fanoutqa}, leveraging shortcuts \citep{schnitzler2024morehopqa}, or ignoring the temporal relation between events \citep{NEURIPS2024_e560a0b2}. Therefore, identifying and categorizing mistakes made by LLMs in these tasks is still an important goal.

\subsection{Symbolic Reasoning}
\label{sec:appendix-task-symbol}

Since the discovery of Chain-of-thought prompting \citep{NEURIPS2022_9d560961, NEURIPS2022_8bb0d291}, step-by-step reasoning largely expanded LLMs' ability to solve symbolic reasoning tasks\footnote{While symbolic reasoning may strictly refer to \textit{algorithmic reasoning} in some literature \citep{NEURIPS2022_9d560961, suzgun2022challengingbigbenchtaskschainofthought}, we adopt the broader sense including math and logical reasoning that can be readily expressed in symbols (\textit{e.g.}, equation, logic) \citep{sprague2024cotcotchainofthoughthelps}.} such as \textbf{mathematical reasoning}, \textbf{logical reasoning}, and \textbf{algorithmic reasoning}. As the final answer and the reasoning process are highly objective in these tasks, utility and validity are the two most popular criteria for evaluating reasoning traces from symbolic tasks.

\textbf{Arithmetic reasoning}, where the model has to predict the correct answer from arithmetic word problems, is the most renowned variant of math reasoning. Popular benchmarks include MathQA \citep{amini-etal-2019-mathqa} and GSM8k \citep{cobbe2021trainingverifierssolvemath}, which provide long, diverse natural language queries in contrast to relatively synthetic, simple benchmarks \citep{koncel-kedziorski-etal-2016-mawps, miao-etal-2020-diverse}. Game of 24 \citep{NEURIPS2023_271db992} and Mathador \citep{kurtic-etal-2024-mathador} ask to combine given numbers and arithmetic operations to generate the target number, requiring exploration and backtracking in the exponential solution space.

The recent saturation of LLMs in arithmetic word problems facilitated more challenging \textbf{mathematical reasoning} benchmarks from math competitions and university textbooks, covering fields like calculus, probability, statistics, geometry, number theory, and more \citep{he-etal-2024-olympiadbench, gao2024omnimathuniversalolympiadlevel, glazer2024frontiermathbenchmarkevaluatingadvanced, zhang-etal-2024-geoeval}. While these benchmarks were highly challenging to the state-of-the-art LLMs of the time of release, recently emerging large reasoning models \citep{openai2024openaio1card, qwenlmQwQReflect, deepseekai2025deepseekr1incentivizingreasoningcapability} achieve unprecedented performance in these benchmarks by generating long reasoning traces often with self-verification and backtracking.

\textbf{Deductive logical reasoning} \citep{tafjord-etal-2021-proofwriter, tian-etal-2021-diagnosing, PrOntoQA, han-etal-2024-folio} mainly focuses on logical deduction, where one should repeatedly apply the general rules to specific facts as in classical syllogism. \textbf{Constraint-based reasoning} \citep{zhong2021arlsatinvestigatinganalyticalreasoning, tyagi-etal-2024-step} is a variant of deductive reasoning where one must find the solution that satisfies the provided initial constraints (\textit{e.g.}, grid puzzles \citep{zhong2021arlsatinvestigatinganalyticalreasoning}). As these datasets are easy to solve in a symbolic form like logic programming \citep{PrOntoQA, pan-etal-2023-logic, olausson-etal-2023-linc, lee2025symbasymbolicbackwardchaining} but harder in natural language due to the size of the search space \citep{kang2024empiricalcomplexityreasoningplanning}, they have served as a diagnostic benchmark for understanding and analyzing the complex reasoning capability of large language models \citep{sinha-etal-2019-clutrr, PrOntoQA, han-etal-2024-folio}. However, as these datasets are often synthetically generated from their symbolic representations, they might not fully generalize to real-world problems with linguistic diversity and commonsense.

Finally, \textbf{algorithmic reasoning} tasks include manipulating strings and data structures, such as concatenating the last letters of the given words \citep{NEURIPS2022_9d560961} or completing the incomplete Dyck language. BIG-Bench-Hard (BBH; \citet{suzgun2022challengingbigbenchtaskschainofthought}) and NPHardEval \citep{fan2024nphardevaldynamicbenchmarkreasoning} include 11 and 9 algorithmic reasoning tasks, respectively, which are challenging for modern LLMs like GPT-4 and PaLM-540B.

\subsection{Others}

\textbf{Science reasoning} tasks lie between factual/commonsense reasoning tasks and symbolic reasoning tasks, as they often require understanding complicated facts combined with world knowledge and performing precise math/logical reasoning \citep{hendrycks2021measuringmassivemultitasklanguage, rein2024gpqa, he-etal-2024-olympiadbench, lu2025scp116khighqualityproblemsolutiondataset}. The most popular benchmark in this field, GPQA-Diamond \citep{rein2024gpqa}, contains 546 questions from physics, chemistry, and biology, where human experts only get ~65\% of the problems correct.

\textbf{Expert-domain reasoning} includes domain-specific reasoning tasks that often require significant expertise in the field, \textit{e.g.}, biomedical reasoning \citep{suster-daelemans-2018-clicr, savage2024diagnostic, zuo2025medxpertqabenchmarkingexpertlevelmedical}, legal reasoning \citep{holzenberger2020datasetstatutoryreasoningtax, guha2023legalbenchcollaborativelybuiltbenchmark, kimyeeun-etal-2024-developing}, and financial reasoning \citep{chen2022finqadatasetnumericalreasoning, li-etal-2024-alphafin}. These tasks require both domain-specific knowledge and reasoning strategies, posing a significant challenge to modern language models \citep{zuo2025medxpertqabenchmarkingexpertlevelmedical, li-etal-2024-alphafin}. However, due to the high cost of expert annotation, existing methods often oversimplify real-world challenges \citep{holzenberger-van-durme-2021-factoring, guha2023legalbenchcollaborativelybuiltbenchmark}; consequently, the demand for resources that closely reflect real-world expert applications is rising.

\textbf{Programming/coding} is closely related to algorithmic reasoning. Popular benchmarks regarding programming include \textit{competitive coding}, where one has to solve an algorithm problem given in natural language and test codes \citep{chen2021evaluatinglargelanguagemodels, li2022competition}, and \textit{practical coding} that covers tasks of software engineers and developers \citep{zhang-etal-2023-repocoder, jimenez2024swebenchlanguagemodelsresolve, DBLP:journals/corr/abs-2410-07095}. Programming differs from other reasoning tasks in various aspects: (1) codes are strictly constrained by predefined syntax and semantics, and (2) the result is evaluated by the execution result rather than the code itself. These constraints make (1) segmenting the trace (code) into steps and (2) applying metrics that require explicitly stated answers, \textit{i.e.}, $\mathcal{V}$-information, difficult than in natural language reasoning traces. Therefore, most evaluators specialized in code focus on trace-level utility rather than step-wise evaluation, defined as the pass rate of predefined unit tests \citep{dai2025processsupervisionguidedpolicyoptimization}.
\section{Appendix for Meta-evaluation Datasets}
\label{sec:appendix-resources}

This appendix includes discussions on the dataset construction process, with a focus on data annotation and label types. A summary of existing datasets can be found in Table \ref{tab:resources}. 

\subsection{Data collection process}

\subsubsection{Labeling methods}

\paragraph{Human annotation} The most straightforward approach to decide the ground truth label is to use \textit{human evalua tion} \citep{DBLP:conf/iclr/LightmanKBEBLLS24, jacovi-etal-2024-chain, zeng2024mrgsm8kmetareasoningbenchmarklarge, zheng2024processbenchidentifyingprocesserrors}. The largest human annotation experiment was conducted by \citet{DBLP:conf/iclr/LightmanKBEBLLS24}, where crowdsourced annotators labeled the validity of 800k steps (75k reasoning traces). Due to the sheer volume of annotation, an active learning strategy was used; the annotators were requested to annotate \textit{hard} samples (the final answer is incorrect but judged as valid by the reward model), which were added to the training data for the next version of the reward model.

\paragraph{LLM annotation} As a cheap alternative for human evaluation, LLM-as-a-judge is often used to generate labels \citep{gao2024llmcriticshelpcatch, zhang2025lessonsdevelopingprocessreward}. However, LLM-assigned labels are not fully credible, given that state-of-the-art LLMs still make errors in human-annotated datasets \citep{zheng2024processbenchidentifyingprocesserrors, kim2025scalingevaluationtimecomputereasoning}. Therefore, LLM-annotated data is often used to augment the training data rather than for meta-evaluation purposes.

\paragraph{Perturbation} Another method to create positive and negative samples is to insert errors into correct reasoning traces. For instance, \citet{zhu2024deductivebeamsearchdecoding, lu2024stepcontrolleddpoleveragingstepwise} samples traces that reach the correct answer, and prompts an LLM to introduce a predefined form of perturbation to the reasoning trace. This allows easy sampling of diverse erroneous traces that can improve the robustness of evaluators, but using human-defined errors might not correctly reflect the true distribution of LLM-generated errors.

\paragraph{Step-level utility} Some datasets use step-level utility as their labels. The most prominent approach is \textit{Monte Carlo Tree Search} \citep{wang-etal-2024-math}, where the step-level utility is measured by sampling \textit{rollouts} from a step and checking if they reach the correct answer. However, to increase the efficiency of the search for negative labels (low utility), \citet{luo2024improvemathematicalreasoninglanguage, dai2025processsupervisionguidedpolicyoptimization} implements a binary search algorithm to locate the first step with low utility. One notable variant of step-level utility labels is \textit{advantage}, where the evaluators are not trained to predict the expected reward of each node but the \textit{change} in the expected rewards before and after generating the step \citep{setlur2024rewardingprogressscalingautomated}.

\paragraph{Trace-level utility} The coarsest label is the trace-level utility, simply measured by the correctness of the final answer \citep{lambert-etal-2025-rewardbench}.

Both trace-level and step-level utilities do not require human annotation other than the final answer, which is much cheaper to obtain than human annotations \citep{wang-etal-2024-math}. However, they cannot serve as a reliable proxy of factuality/coherence/validity due to \textit{unfaithful reasoning}, where traces that reach the correct answer (high utility) often include factual/logical errors \citep{lanham2023measuringfaithfulnesschainofthoughtreasoning, lyu-etal-2023-faithful, zheng2024processbenchidentifyingprocesserrors, kim2025scalingevaluationtimecomputereasoning}.

\subsubsection{Inter-annotator agreement}

While reasoning trace evaluation is considered more objective than other long-text evaluation tasks (\textit{e.g.}, helpfulness, bias/harmfulness, and language proficiency) \citep{wang2024secretsrlhflargelanguage}, a certain amount of inter-annotator disagreement is inevitable. Here, we report the trend in inter-annotator agreement observed in existing human annotation works.

\paragraph{Incorrect solutions for harder problems lead to higher disagreement} ProcessBench \citep{zheng2024processbenchidentifyingprocesserrors} consolidates the intuitive hypothesis that inter-annotator disagreement grows when the query is difficult and the trace is incorrect in at least one step. Compared to the easiest case (GSM8k queries, correct trace), where three annotators agree in 95.9\% of the cases, the hardest case (OmniMATH, incorrect trace) shows only 47.8\% of three-annotator agreement. 

\paragraph{Inter-annotator disagreement reflects vagueness in natural language} In many cases, the disagreement is significantly affected by the linguistic aspects of the reasoning trace. REVEAL \citep{jacovi-etal-2024-chain} manually classifies steps that aroused disagreement between annotators into 13 distinct categories. Among these, frequent disagreement types like "World knowledge (some world knowledge might not be taken for granted)" and "Unclear reference (one proper noun can refer to multiple real-world entities)" are typical disagreement types observed in simpler \textit{recognizing textual entailment} (natural language inference) tasks \citep{NEURIPS2018_4c7a167b, lee2025entailmentpreservingfirstorderlogicrepresentations}, showing that these vagueness is present even in \textit{minimal} settings.

On the other hand, synthetic, algorithmic reasoning tasks like BIG-Bench-Hard \citep{suzgun2022challengingbigbenchtaskschainofthought} are linguistically uniform. Consequently, BIG-Bench-Mistake that annotate errors in this benchmark \citep{tyen-etal-2024-llms} observes near-perfect inter-annotator agreement (Krippendorf's $\alpha>0.97$), again demonstrating the strong connection between linguistic variation and inter-annotator agreement.

\subsection{Label types}

\paragraph{Sequence classification} The most common label type is sequence classification, where a quality label is assigned to each step/trace. For example, \citet{wang-etal-2024-math} assigns binary labels to steps based on the utility, and \citet{DBLP:conf/iclr/LightmanKBEBLLS24} assigns ternary validity labels (\textit{correct/incorrect/neutral}) obtained by human annotation. The \textit{neutral} label in \citet{DBLP:conf/iclr/LightmanKBEBLLS24} was introduced to absorb ambiguous cases and minimize inter-annotator disagreement; considering it as positive or negative when training the evaluator does not significantly affect the Best-of-N performance (<1.0p).

One caveat of sequence classification is that it is hard to define the labels \textit{after} the first error (propagated error). It is often unclear whether steps that rely on the first erroneous step should be labeled as incorrect (because they rely on incorrect premises) or correct (because the reasoning is correct if assuming the premises are correct) \citep{jacovi-etal-2024-chain, mukherjee2025premiseaugmentedreasoningchainsimprove}. Two different label schemas are used to bypass this ambiguity: annotating the \textit{pairwise preference} and annotating the \textit{index of the first erroneous step}.

\paragraph{Preference (win/lose)} Reasoning trace evaluation can be formulated as a preference problem \citep{lai2024stepdpostepwisepreferenceoptimization, lu2024stepcontrolleddpoleveragingstepwise, lambert-etal-2025-rewardbench}. In this scenario, data points are defined as pairs of reasoning traces, one as the winner and the other as the loser. The pairs are often constructed by sampling two different continuations from a shared prefix or perturbing a correct trace. These data are often used to train the LLM-as-a-value-function models via preference learning algorithms, \textit{e.g.}, DPO \citep{NEURIPS2023_a85b405e}.

\paragraph{Identifying first erroneous index} Another method is to label the index of the first erroneous step \citep{zheng2024processbenchidentifyingprocesserrors, zeng2024mrgsm8kmetareasoningbenchmarklarge}. In this setting, the reasoning trace is given as a list of steps, and the evaluator must predict the index of the first error. If there is no error, the model should predict -1. This setting effectively bypasses the propagated error problem, but converting these labels to binary classification can lead to better performance in sequence classifiers and critic models \citep{kim2025scalingevaluationtimecomputereasoning}.

\section{Comparing criteria definitions}
\label{sec:comparison}

\subsection{Comparison between proposed definitions}
\label{sec:comparison-ours}

\paragraph{Factuality$\leftrightarrow$Validity} Factuality focuses on the relationship between a step and provided/external knowledge, while validity focuses on the relationship between two model-generated steps. For instance, Given an incorrect step \textit{Albert Einstein died in 1965} (he died in 1955), this step is not factual if the query explicitly mentions that \textit{Einstein died in 1955}. Apart from that, if the previous steps provide the premises for reaching 1955, \textit{i.e.} \textit{Einstein was born in 1879, and he died at the age of 76}, the step is invalid.

While the standard practice is to treat factuality and validity separately \citep{prasad-etal-2023-receval, zhu2024deductivebeamsearchdecoding, jacovi-etal-2024-chain}, the boundary between stating facts and making logical inferences is often vague, especially in commonsense reasoning. For example, if the step states \textit{Einstein died between 1960 and 1970} when given the information \textit{Einstein died in 1955}, is this step a factual error or a logical error? The boundary heavily relies on the definition of what can be taken as granted, which is also a key factor in defining coherence. REVEAL \citep{jacovi-etal-2024-chain} delegates the decision to human annotators, and shows that LLMs \citep{anil2023palm2technicalreport, brown2020languagemodelsfewshotlearners} perform poorly (F1<0.65) at classifying the steps between factual statements and logical inference.

\paragraph{Validity$\leftrightarrow$Coherence} Existing works often treat coherence as a subtype of validity \citep{DBLP:conf/iclr/GolovnevaCPCZFC23, zhu2024deductivebeamsearchdecoding, kim2024biggenbenchprincipledbenchmark, jacovi-etal-2024-chain}, as both criteria judge a step based on its previous steps. However, validity and coherence are different by definition, as validity focuses on the logical correctness of a step while coherence focuses on the pragmatic aspect of informativeness. For instance (Figure \ref{fig:criteria}-Coherence), omitting a step (Step 3) from the correct trace will make the subsequent step (Step 3') incoherent, but it is still valid since it can be eventually deduced from the query and previous steps.

\paragraph{Validity$\leftrightarrow$Utility} Previous studies have continuously pointed out that validity does not necessarily lead to utility and vice versa \citep{lyu-etal-2023-faithful, nguyen-etal-2024-direct}. One case is \textit{shortcut reasoning} \citep{schnitzler2024morehopqa, lee2025symbasymbolicbackwardchaining}, where LLM generates invalid Chain-of-thoughts but guesses the correct answer directly from the query. ProcessBench \citep{zheng2024processbenchidentifyingprocesserrors} reports that invalid traces with correct answers can be easily found in challenging problems, reaching 51.8\% in the olympiad-level Omni-MATH \citep{gao2024omnimathuniversalolympiadlevel}.

The distinction between validity and utility has been highlighted by multiple empirical results. Treating these metrics as different yields substantial performance gain when training sequence classifiers \citep{zhang2025lessonsdevelopingprocessreward} and in Best-of-N decoding \citep{sun2024easytohardgeneralizationscalablealignment, kim2025scalingevaluationtimecomputereasoning}. See Section \ref{sec:analysis} for details.

\subsection{Comparison to other definitions}

\begin{figure}[tb]
    \centering
    \includegraphics[width=0.8\linewidth]{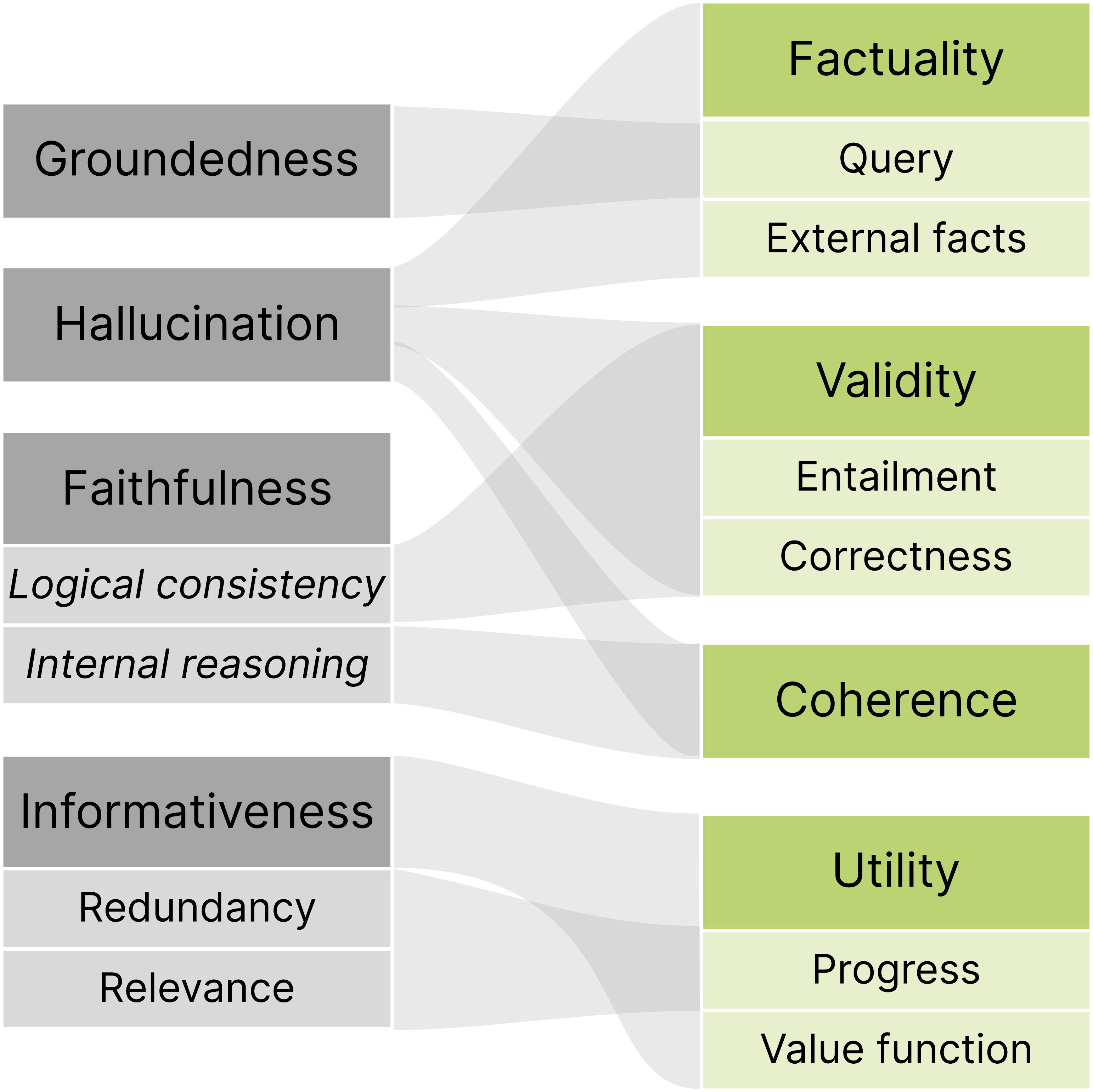}
    \caption{A Sankey diagram displaying the relationship between commonly used terminologies (left) to the proposed taxonomy (right).}
    \label{fig:sankey}
\end{figure}


\textbf{Hallucination} is most commonly defined as \textit{"models either generating (1) nonsensical or (2) unfaithful to the source content"} \citep{ji2023survey, banerjee2024llmshallucinateneedlive, huang2024survey}, which corresponds to (1) validity/coherence and (2) factuality. However, some works restrict the meaning of hallucination to factual errors, \textit{i.e.} \textit{"models generating description tokens that are not supported by the source inputs"} \citep{xiao-wang-2021-hallucination, akbar-etal-2024-hallumeasure}.

\textbf{Faithfulness} is also used with different senses. The most common definition for faithfulness is \textit{"logical consistency between the generated text and the query/previous steps"} \citep{maynez2020faithfulnessfactualityabstractivesummarization, creswell2022faithfulreasoningusinglarge, huang2024survey}, which includes both factuality (query groundedness) and validity (previous step). Instead, faithfulness can be used as \textit{"accurately representing the model's internal reasoning process"} \citep{lyu-etal-2023-faithful, lanham2023measuringfaithfulnesschainofthoughtreasoning}. Under this definition, the final step containing the answer is unfaithful if it is not supported by the previous steps, which falls under the definition of coherence.

\textbf{Informativeness} is defined as "providing new information that is helpful towards deriving the generated answer" \citep{golovneva2023pathfinderguidedsearchmultistep, prasad-etal-2023-receval}. Lack of informativeness is often described as \textbf{redundancy} \textit{"removing the step does not affect the reasoning process"} \citep{chiang-lee-2024-reasoning, song2025prmbenchfinegrainedchallengingbenchmark, zhou2024languagemodelsperformrobust} or \textbf{irrelevance} \textit{"unrelated to the query's topic or task"} \citep{wang-etal-2023-towards, zhou2024languagemodelsperformrobust, jacovi-etal-2024-chain}. Informativeness is synonymous with utility, as it aims to evaluate the contribution of a step to reaching the final answer.




\section{Details for Section 6}
\label{sec:appendix-analysis}

This section provides further details regarding the Section \ref{sec:analysis}, specifically Figure \ref{fig:compute-to-processbench}.

\subsection{Estimating Compute}

To estimate the compute in Figure~\ref{fig:compute-to-processbench}, we follow the approximation equation from \citet{snell2024scalingllmtesttimecompute, kim2025scalingevaluationtimecomputereasoning}. Specifically, the computational cost can be asymptotically approximated as
\[
C \in O(N \times L),
\]
where $C$ is the total computational cost, $N$ is the number of parameters, and $L$ is the number of tokens. Note that since all compared evaluators use the same base model, $N$ remains constant.

Below, we describe how the computation budget for each method is calculated in Figure~\ref{fig:compute-to-processbench}:

\begin{itemize}
    \item Unit relative compute (1) corresponds to a single forward pass for an average-length trace. This applies to \textit{Fine-tuned Sequence Classifiers}, as they take the whole trace as the input.
    
    \item The \textit{Base Model}, \textit{Fine-tuned Critic Models} \citep{she2025rprmreasoningdrivenprocessreward}, and \textit{Fine-tuned LRMs} evaluate each step with a separate forward pass. Thus, the compute is scaled by the number of steps per trace, which is 6.11 on average in ProcessBench (GSM8k + MATH). Note that LRMs like \texttt{DeepSeek-Distill-Qwen-2.5-7B} \citep{deepseekai2025deepseekr1incentivizingreasoningcapability} generate significantly longer traces, with $L$ scaled by $7.58$.
    
    \item \textit{PARC} \citep{mukherjee2025premiseaugmentedreasoningchainsimprove} also uses step-wise critic evaluations, but only using the \textit{Partial context} (average 1.57 premises per step) makes PARC require lower compute by reducing $L$.
    
    \item In \textit{Majority Voting} setting, where 8 step-wise evaluations are sampled per step and aggregated via majority voting, the total computation cost is multiplied by 8.
\end{itemize}

\subsection{Data source}

\begin{itemize}
    \item For \textit{Base model} and \textit{Majority voting} scores, authors conducted experiments with \texttt{Qwen-2.5-7B-Instruct} using the code from \citet{kim2025scalingevaluationtimecomputereasoning}. \textit{Fine-tuned LRM} scores are as reported in the same paper.
    
    \item \textit{Sequence classifiers} scores are obtained from \citet{zhang2025lessonsdevelopingprocessreward}.
    
    \item \textit{Partial context} scores are provided by the authors of PARC \citep{mukherjee2025premiseaugmentedreasoningchainsimprove}, upon requested by the authors of this survey. While the currently available version of the paper does not contain the result, it will appear in the published version.
    
    \item \textit{Fine-tuned Critic Model} scores are from \citet{she2025rprmreasoningdrivenprocessreward}.
    
\end{itemize}

\end{document}